\documentclass[11pt,a4paper]{article}
\usepackage[utf8]{inputenc}
\usepackage[american]{babel}
\usepackage[superscript,nomove]{cite}
\usepackage{amsmath,amssymb,amsfonts}
\usepackage{algorithmic}
\usepackage{graphicx}
\usepackage{textcomp}
\usepackage{booktabs}
\usepackage{numprint}
\usepackage{authblk}
\def\BibTeX{{\rm B\kern-.05em{\sc i\kern-.025em b}\kern-.08em
    T\kern-.1667em\lower.7ex\hbox{E}\kern-.125emX}}
\usepackage{soul}
\usepackage[hyphens]{url}
\usepackage{breakurl}
\usepackage{float}
\usepackage{hyperref}
\usepackage{subfig}
\usepackage{longtable}
\usepackage[table]{xcolor}
\usepackage{multirow}
\usepackage[font={small},labelfont={bf}]{caption}
\usepackage[misc]{ifsym}
\usepackage{sectsty}
\usepackage[most]{tcolorbox}
\usepackage[margin=1in]{geometry}
\usepackage{fontawesome5}

\definecolor{darkblue}{rgb}{0.0, 0.0, 0.6}
\definecolor{darkerblue}{rgb}{0.0, 0.0, 0.4}
\definecolor{darkcyan}{rgb}{0.0, 0.6, 0.6}
\definecolor{darkercyan}{rgb}{0.0, 0.4, 0.4}
\definecolor{darkerbluegray}{RGB}{67, 106, 161}
\definecolor{darkestbluegray}{RGB}{27, 66, 121}
\definecolor{darkred}{rgb}{0.6, 0.0, 0.0}
\definecolor{darkerred}{rgb}{0.4, 0.0, 0.0}
\definecolor{darkgreen}{rgb}{0.0, 0.4, 0.0}
\definecolor{darkergreen}{rgb}{0.0, 0.3, 0.0}
\definecolor{darksalmon}{rgb}{0.71, 0.39, 0.28}
\definecolor{darkersalmon}{rgb}{0.61, 0.29, 0.18}
\definecolor{darkpurple}{rgb}{0.5, 0.0, 0.5}
\definecolor{darkerpurple}{rgb}{0.4, 0.0, 0.4}
\definecolor{darkteal}{rgb}{0.0, 0.4, 0.4}
\definecolor{darkerteal}{rgb}{0.0, 0.3, 0.3}
\definecolor{darklavender}{rgb}{0.5294, 0.3020, 0.4745}
\definecolor{darkviolet}{RGB}{122, 65, 145}
\definecolor{darkyellow}{rgb}{0.5, 0.5, 0.0}
\definecolor{darkpink}{rgb}{0.7, 0.0, 0.5}
\definecolor{darkerpink}{rgb}{0.6, 0.0, 0.4}
\definecolor{black}{rgb}{0.0, 0.0, 0.0}
\definecolor{white}{rgb}{1.0, 1.0, 1.0}
\definecolor{darklime}{RGB}{102, 153, 0}
\definecolor{darkerlime}{RGB}{92, 123, 0}
\definecolor{evendarkerlime}{RGB}{67, 89, 0}
\definecolor{mednext}{HTML}{7d2727}
\definecolor{swinunetr}{HTML}{404d2e}
\definecolor{transunet}{HTML}{3b3f47}

\newcommand{\textbfb}[1]{\textbf{\textcolor{darkblue}{#1}}}
\newcommand{\textbfr}[1]{\textbf{\textcolor{darkred}{#1}}}
\newcommand{\textbfg}[1]{\textbf{\textcolor{darkgreen}{#1}}}

\newcommand{\textbft}[1]{\textbf{\textcolor{darkteal}{#1}}}

\newcommand{\textbfbg}[1]{\textbf{\textcolor{darkerbluegray}{#1}}}
\newcommand{\textbfm}[1]{\textbf{\textcolor{darkpink}{#1}}}
\newcommand{\textbfdbg}[1]{\textbf{\textcolor{darkestbluegray}{#1}}}
\newcommand{\textbfs}[1]{\textbf{\textcolor{darkersalmon}{#1}}}
\newcommand{\textbfli}[1]{\textbf{\textcolor{darkerlime}{#1}}}
\newcommand{\textbftu}[1]{\textbf{\textcolor{transunet}{#1}}}

\begingroup\lccode`~=`_
\lowercase{\endgroup\def~}#1{_{\scriptscriptstyle#1}}

\AtBeginDocument{\mathcode`_="8000 \catcode`_=12 }

\makeatletter
\renewcommand{\@seccntformat}[1]{}
\newcommand{\sectionn}{\@startsection{section}{1}{0mm}%
                                {-\baselineskip}%
                                {0.5\baselineskip}%
                                {\normalfont\large\bfseries\MakeUppercase}}%
\makeatother

\newcommand\rurl[1]{%
  \href{http://#1}{\nolinkurl{#1}}%
}

\sectionfont{\fontsize{18}{18}\selectfont}

\hypersetup{breaklinks,hidelinks=true,colorlinks=true,citecolor=darkerpurple,linkcolor=darkblue,urlcolor=darkerteal}

\title{\bfseries\huge MIRAGE: Multimodal foundation model and benchmark for comprehensive retinal OCT image analysis}

\author[1,2,{\normalfont\faEnvelope[regular]}]{Jos\'e Morano}
\author[1,2]{Botond Fazekas}
\author[3]{Emese S\"ukei}
\author[1,2]{Ronald Fecso}
\author[1,2]{Taha Emre}
\author[3]{Markus Gumpinger}
\author[1,2]{Georg Faustmann}
\author[1,2]{Marzieh Oghbaie}
\author[3]{Ursula Schmidt-Erfurth}
\author[1,2,{\normalfont\faEnvelope[regular]}]{Hrvoje Bogunovi\'c}

\affil[1]{\small Christian Doppler Laboratory for Artificial Intelligence in Retina, Institute of Artificial Intelligence, Center for Medical Data Science, Medical University of Vienna, Vienna, Austria}
\affil[2]{\small Comprehensive Center for AI in Medicine, Medical University of Vienna, Vienna, Austria}
\affil[3]{\small OPTIMA Lab, Department of Ophthalmology, Medical University of Vienna, Vienna, Austria

\vspace{0.4em}

{\textsuperscript{\normalfont\faEnvelope[regular]}~\texttt{jose.moranosanchez@meduniwien.ac.at}, \texttt{hrvoje.bogunovic@meduniwien.ac.at}}
}

\date{Accepted for publication in {\sffamily\itshape\textcolor{darkred}{npj} Digital Medicine}\\
DOI: \texttt{\href{https://doi.org/10.1038/s41746-025-01852-3}{10.1038/s41746-025-01852-3}}
}

\begin{document}

\maketitle



\begin{abstract}
    Artificial intelligence (AI) has become a fundamental tool for assisting clinicians in analyzing ophthalmic images, such as optical coherence tomography (OCT).
    However, developing AI models often requires extensive annotation, and existing models tend to underperform on independent, unseen data.
    Foundation models (FMs), large AI models trained on vast unlabeled datasets, have shown promise in overcoming these challenges.
    Nonetheless, available FMs for ophthalmology lack extensive validation, especially for segmentation tasks, and focus on a single imaging modality.
    In this context, we propose MIRAGE, a novel multimodal FM for the analysis of OCT and scanning laser ophthalmoscopy (SLO) images.
    Additionally, we propose a new evaluation benchmark with OCT/SLO classification and segmentation tasks.
    The comparison with general and specialized FMs and segmentation methods shows the superiority of MIRAGE in both types of tasks, highlighting its suitability as a basis for the development of robust AI systems for retinal OCT image analysis.
    Both MIRAGE and the evaluation benchmark are publicly available at \url{https://github.com/j-morano/MIRAGE}.
\end{abstract}

\section{Introduction}

The prevalence of ocular and systemic diseases affecting the eye represents a significant public health concern, with an estimated 1 billion individuals worldwide affected by visual impairment or blindness~\cite{burton2021lancet}.
Major causes of vision loss include retinal diseases such as age-related macular degeneration (AMD) and glaucoma, and systemic conditions such as hypertension and diabetes, which frequently leads to diabetic retinopathy (DR).
The aging of the population and the increasing prevalence of hypertension and diabetes are expected to further worsen the burden of retinal diseases.
In this context, early detection and accurate diagnosis and characterization are crucial for timely intervention and adequate management to prevent or slow vision loss.

Diagnosis and monitoring of retinal diseases rely on imaging modalities such as color fundus photography (CFP) and optical coherence tomography (OCT) with scanning laser ophthalmoscopy (SLO) (also referred to as near-infrared imaging).
While CFP and SLO provide 2D \textit{en-face} views of the retina, 3D OCT provides cross-sectional images (B-scans) for a detailed 3D analysis of retinal layers and their thickness and integrity~\cite{kanski2011clinical,mohammadpour2020diagnostics}.
As a result, OCT has become the gold standard for the diagnosis of diseases such as AMD, glaucoma and diabetic macular edema (DME) as well as the primary modality in the management of patients undergoing treatment with anti-vascular endothelial growth factor (anti-VEGF) drugs~\cite{SE2016prer,schuman2024optical}.
However, the interpretation of these detailed OCT images is complex, time-consuming, and requires specialized expertise.
Furthermore, manual qualitative analysis is inherently subjective and prone to inter- and intra-observer variability, which affects diagnostic accuracy and reliability~\cite{bemme2021reliability,schmitz2023interreader}.

Several artificial intelligence (AI) methods, especially based on deep learning (DL), have been developed for the automated or semi-automated analysis of retinal OCT images~\cite{defauw2018clinically,schmidt2018artificial,Yim2020,Ting2019}.
AI has proven particularly valuable in detecting and quantifying OCT biomarkers in AMD, the leading cause of vision loss in the developed world, to support patient management~\cite{schmidt2022aibased,reiter2024ai}.
However, their performance often depends on the availability of large, manually annotated datasets, which are crucial for training large DL models.
Creating these datasets is particularly challenging in the medical domain due to the sensitive nature of medical data, the high acquisition costs, and the required expertise.
As a result, datasets tend to be small and homogeneous, limiting the generalizability and robustness of AI models and thus their translation into clinical practice~\cite{paschali2018generalizability,hemelings2023generalizable,yang2024limits}.

\begin{figure}[tbp]
    \centering
    \includegraphics[width=\linewidth]{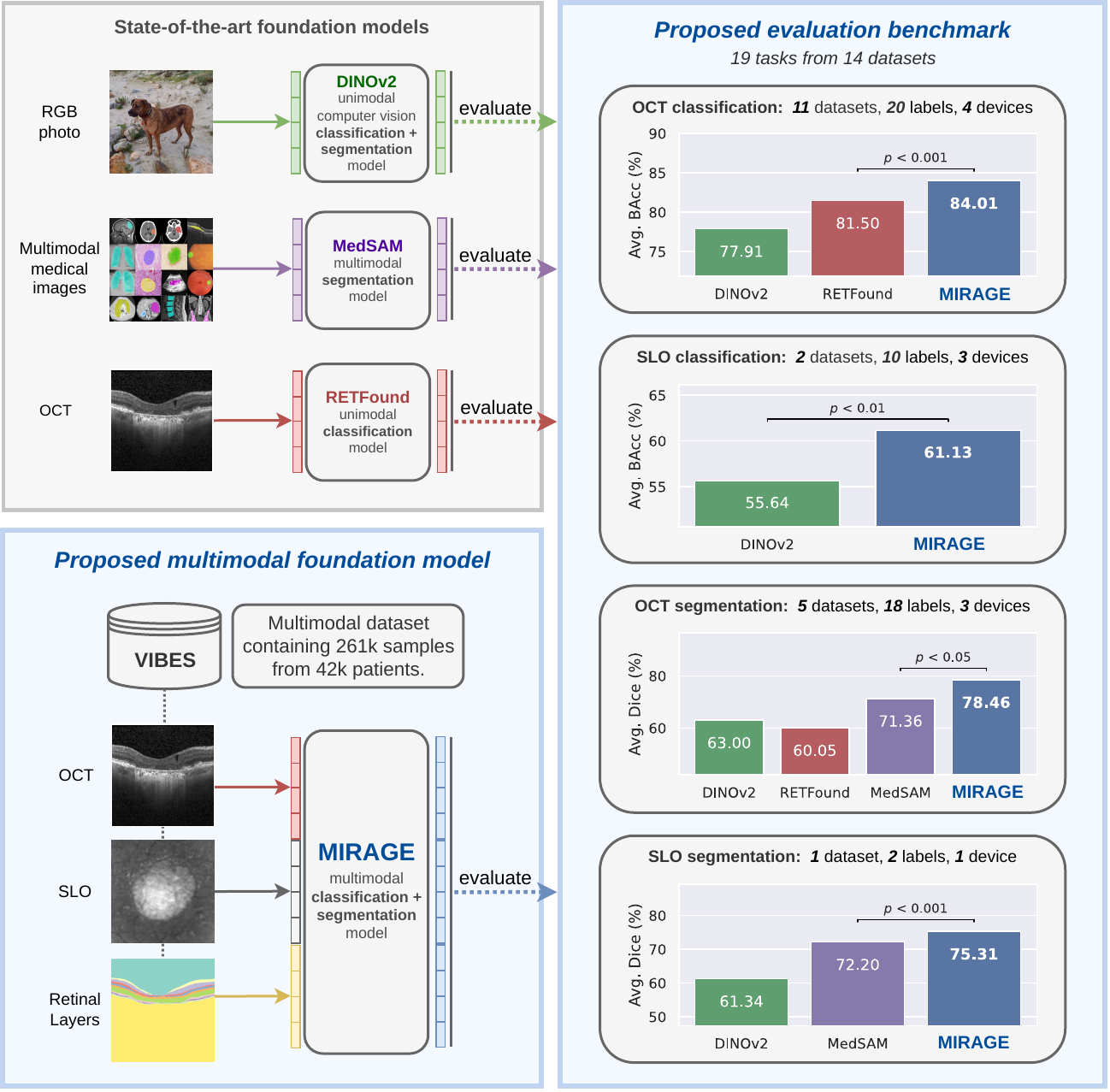}
    \caption{%
        \textbf{Overview of the proposed model (MIRAGE) and other general (DINOv2) and domain-specific (MedSAM, RETFound) foundation models.}
        In contrast to existing unimodal foundation models, our approach utilizes multimodal self-supervised learning to train a Vision Transformer on a large dataset of paired multimodal retinal images, including optical coherence tomography (OCT), scanning laser ophthalmoscopy (SLO), and automatically generated labels for retinal layers.
        We evaluated the model on a comprehensive benchmark consisting of 19 tasks from 14 publicly available datasets and two private datasets, covering both OCT and SLO classification and segmentation tasks.
        Statistical significance was calculated using the Wilcoxon signed-rank test across all datasets.
        Our foundation model, MIRAGE, significantly outperforms state-of-the-art foundation models across all task types.
    }%
    \label{fig:overview}
\end{figure}

Self-supervised learning (SSL)~\cite{ericsson2022self} offers a promising solution to these problems by enabling models to learn meaningful representations from the data without expert annotation.
SSL methods train models on unlabeled datasets using pretext tasks such as masked autoencoding (MAE)~\cite{he2022mae}, contrastive learning~\cite{le2020contrastive,radford2021clip}, and self-distillation~\cite{caron2021emerging}.
While most SSL methods are unimodal, recent research highlights the exciting potential of multimodal approaches to significantly improve the performance of DL models in downstream tasks~\cite{radford2021clip,girdhar2023imagebind,bachmann2022multimae}.
A successful example in computer vision is MultiMAE~\cite{bachmann2022multimae}, which extends the concept of MAE to multimodal image data.
In the context of retinal imaging, multimodal reconstruction~\cite{morano2020multimodal,li2020self,morano2023self}, which aims to learn representations by reconstructing one modality from another, and contrastive learning using multimodal pairs (image--text~\cite{silva2023flair}, image--image~\cite{sukei2024multimodal}) have shown superior performance compared to unimodal approaches.

Advances in SSL and scalable network architectures such as Vision Transformers (ViTs)~\cite{dosovitskiy2020image} have facilitated the development of foundation models (FMs)~\cite{bommasani2021opportunities,tu2023towards}, large DL models that are pretrained on vast datasets and can be applied to diverse tasks with minimal tuning.
In the field of computer vision, the DINOv2 model~\cite{oquab2024dinov2} has established a new benchmark for SSL methods in downstream tasks such as classification, detection, and segmentation.
In ophthalmology, FMs such as RETFound~\cite{zhou2023retfound} (which includes two separate models for OCT and CFP), FLAIR~\cite{silva2023flair} (for CFP), VisionFM~\cite{qiu2024visionfm} (including separate models for five modalities), EyeFound~\cite{shi2024eyefound}, and EyeCLIP~\cite{shi2024eyeclip} (both multimodal, focusing mainly on CFP and fluorescein angiography) have shown superior performance in diagnostic and prognostic tasks compared to models pretrained on ImageNet~\cite{deng2009imagenet} (IN) and other datasets of natural images.
However, most of these models either focus on a single retinal imaging modality~\cite{silva2023flair,zhou2023retfound,qiu2024visionfm} or naively mix multiple unpaired modalities~\cite{shi2024eyefound,cai2024uni4eye++} during training, completely ignoring the relationship between them.
Importantly, although VisionFM was presented as a multimodal FM, it has different (and separately trained) encoders for each image modality.
This strategy, also used in RETFound, results in a ``zoo'' of unimodal models, but not in a truly multimodal FM.
Other models, such as EyeCLIP, although trained on partially paired multimodal data, still mixes arbitrary non-paired modalities for training, and relies on a CLIP-based contrastive learning approach to learn a joint representation space of the fewer paired cases.
This CLIP-based approach strongly focuses on global features, and does not exploit the more fine-grained relations between the modalities, which are crucial for pixel-level tasks such as segmentation~\cite{dong2023maskclip}.
Furthermore, the lack of evaluation on segmentation tasks of models such as RETFound, FLAIR, and EyeCLIP further limits their potential adoption in real-world clinical settings, where segmentation is one of the most useful applications of AI in retina~\cite{schmidt2018artificial}.
On the other hand, FMs for medical image segmentation~\cite{mazurowski2023segment,huang2024segment,ma2024segment,zhu2024medsam2}, most of them based on SAM~\cite{kirillov2023segment}, are trained on large medical image datasets containing heterogeneous, unpaired image modalities (such as X-ray, magnetic resonance imaging, etc.) and some are not even trained on any OCT scans, such as MedSAM-2D~\cite{zhu2024medsam2}.
Thus, they lack the necessary specialization to perform well in a fully automated OCT segmentation setting~\cite{fazekas2023adapting}.

In this context, the main contribution of this work is twofold.
First, we introduce MIRAGE (Figure~\ref{fig:overview}), the first multimodal foundation model for comprehensive analysis of retinal OCT/SLO images, and second, we propose a comprehensive evaluation benchmark for validating foundation models in retinal OCT/SLO imaging, including several classification and segmentation tasks.
MIRAGE was trained on a large dataset using a paired multimodal MAE approach, which allows the model to effectively exploit the complementary information from different modalities and process any of them at inference time.
We thoroughly evaluate MIRAGE on the proposed benchmark and compare it with state-of-the-art (SOTA) SSL methods and foundation models, including DINOv2~\cite{oquab2024dinov2}, RETFound~\cite{zhou2023retfound}, and MedSAM~\cite{ma2024segment}.
Our benchmark results show the superior performance of MIRAGE on both classification and segmentation tasks, showcasing its generalizability and robustness.
These findings show the excellent suitability of MIRAGE as a foundation for the development of AI systems for OCT and SLO image analysis.
Both MIRAGE and the proposed evaluation benchmark, including code and data splits, have been made available to the academic community to facilitate reproducibility and monitoring of research progress in the field.

\section{Results}

\subsection{MIRAGE development data}

MIRAGE was pretrained on a large in-house dataset of multimodal retinal images, the Vienna Imaging Biomarker Eye Study~\cite{gerendas2022validation} (VIBES) registry of the Macula Clinic at the Medical University of Vienna, including OCT and SLO.
For each sample, we additionally generated labels of retinal layers using an automatic method~\cite{garvin2008intraretinal,antony2011automated}.
The model was trained using a multimodal MAE pretext task, which aims to reconstruct all input modalities from masked versions of the same images using a shared ViT encoder.
A total of 261\,184 paired OCT--SLO--Layers samples were used for training.
The complete information about the pretraining dataset can be found in the section~\textit{Methods---Pretraining dataset}.

\subsection{Ocular disease diagnosis}

\begin{figure}[tbp]
    \centering
    \includegraphics[width=\linewidth]{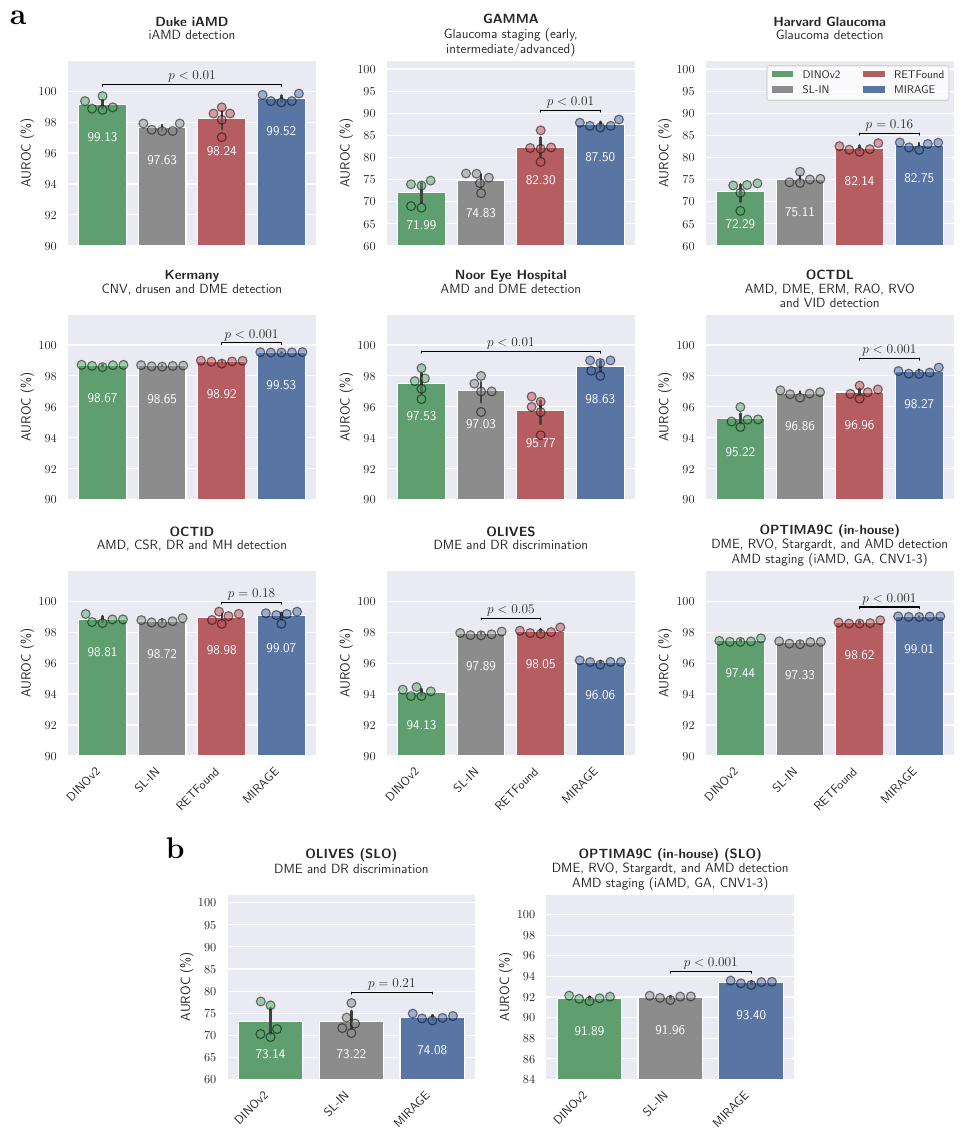}
    \caption{%
        \textbf{Comparison of MIRAGE and other SOTA models for ocular disease diagnosis.}
        \textbf{a} OCT diagnosis.
        \textbf{b} SLO diagnosis.
        For each classification task, we trained the models (all based on ViT-Large) with linear probing using five different random seeds (determining the shuffling of the training data and the augmentation).
        We then evaluated these models on the hold-out test set and obtained five replicas from which the statistics were derived.
        The error bars show the standard deviation.
        In each case, the performance of the two best models was compared to check for statistically significant differences. The \textit{p}-value is calculated using the one-tailed Student's \textit{t}-test and is indicated in the figure.
        MIRAGE outperforms all other models in all but one dataset, with statistically significant differences in 7 out of 11 cases.
    }%
    \label{fig:cls_sota_plots}
\end{figure}

We evaluated MIRAGE on several diagnostic and staging tasks involving different ocular diseases, OCT devices, and imaging modalities from datasets acquired at different institutions worldwide.
In particular, there are 9 tasks based on the same number of datasets (8 public and 1 in-house):
Duke iAMD~\cite{farsiu2014quantitative}, GAMMA~\cite{wu2023gamma}, Harvard Glaucoma~\cite{luo2023harvard}, Kermany~\cite{kermany2018dataset,kermany2018identifying}, Noor Eye Hospital~\cite{rasti2017macular}, OCTDL~\cite{kulyabin2024octdl}, OCTID~\cite{gholami2020octid}, and OLIVES~\cite{prabhushankar2022olives} and OPTIMA9C~\cite{oghbaie2024vlfatrollout} (in-house).
OLIVES and OPTIMA9C are the only datasets that include SLO images.
All tasks are different and feature different diseases or disease stages.
Section~\textit{Methods---Benchmark datasets}
contains further information about these datasets.

For each task, we compared MIRAGE to three other selected SOTA models, all based on ViT: supervised ImageNet pretraining (SL-IN)~\cite{dosovitskiy2020image}, DINOv2~\cite{oquab2024dinov2}, and RETFound~\cite{zhou2023retfound}.
The models were tuned and evaluated on each dataset using \emph{linear probing} (LP), where all the parameters of the model are frozen except for a final linear layer.
In this way, it is possible to determine how discriminative (i.e., how meaningful) the extracted features are for the classification task, and thus how effective the pretraining method was in learning useful data representations.

Figure~\ref{fig:cls_sota_plots} shows the performance of the OCT- and SLO-based models using linear probing in terms of the receiver operating characteristic curve (AUROC) value for each dataset.
The average performances across all OCT and SLO classification datasets are presented in Table~\ref{tab:cls_sota_avg}, including the AUROC, average precision (AP), and balanced accuracy (BAcc) values.
Full quantitative results are presented in Supplementary Tables~\ref{sup:tab:sota_class_oct} and~\ref{sup:tab:sota_class_slo}
for OCT and SLO tasks, respectively (all extended results are presented in Supplementary Note~\ref{sup:sec:results}).

\begin{table}[tbp]
    \caption{\textbf{Average performance of MIRAGE and other SOTA models for ocular disease diagnosis with linear probing.}
    All models are based on the ViT-Large architecture.
    In each case, the best results (in bold) and the second best results (underlined) across all datasets were compared to determine if there were statistically significant differences in average performance.
    To this end, the Wilcoxon signed-rank test was used.
    The results are presented in the table, with the significance level indicated by asterisks (*: $p<0.5$, **: $p<0.01$, ***: $p < 0.001$).
    }
    \label{tab:cls_sota_avg}
    \centering
        \begin{tabular}{lllll}
            \toprule

            \textbf{Modality}
            & \textbf{Model}
            & \textbf{AUROC}
            & \textbf{AP}
            & \textbf{BAcc}
            \\

        \midrule

        \multirow{4}{*}{\textbf{OCT}}
        & DINOv2
        & 91.69 $\pm$ 10.64
        & 87.69 $\pm$ 12.23
        & 77.91 $\pm$ 15.10
        \\

        & SL-IN
        & 92.67 $\pm$ 9.51
        & 88.63 $\pm$ 11.21
        & 80.44 $\pm$ 12.86
        \\

        & RETFound
        & \underline{94.44 $\pm$ 6.66}
        & \underline{91.35 $\pm$ 8.15}
        & \underline{81.50 $\pm$ 11.48}
        \\

        & MIRAGE
        & \textbf{95.59 $\pm$ 5.80}***
        & \textbf{92.99 $\pm$ 6.39}***
        & \textbf{84.01 $\pm$ 10.51}***
        \\

        \midrule

        \multirow{3}{*}{\textbf{SLO}}
        & DINOv2
        & 82.52 $\pm$ 9.68
        & 72.39 $\pm$ 2.70
        & 55.64 $\pm$ 3.48
        \\

        & SL-IN
        & \underline{82.59 $\pm$ 9.52}
        & \underline{74.85 $\pm$ 2.28}
        & \underline{58.79 $\pm$ 6.21}
        \\

        & MIRAGE
        & \textbf{83.74 $\pm$ 9.67}*
        & \textbf{75.39 $\pm$ 2.05}
        & \textbf{61.13 $\pm$ 5.65}**
        \\

            \bottomrule
        \end{tabular}
\end{table}

The results on OCT-based tasks show that MIRAGE outperformed competing models in all but one task, with statistically significant differences in 6 out of 9 cases when evaluated with the one-tailed Student's \textit{t}-test.
The performance of our model was particularly high on datasets involving the diagnosis or staging of AMD and complex multi-class classification tasks.
For example, on the Duke iAMD dataset (for intermediate AMD detection), MIRAGE achieved an AUROC of 99.52\%, outperforming the second-best model, DINOv2, by a significant difference of 0.39 percentage points (pp) ($p < 0.01$).
The same trend was observed for glaucoma staging in the small GAMMA dataset, for which MIRAGE outperformed RETFound by 5.20 pp ($p < 0.01$).
Smaller but still significant differences were also found for Kermany (choroidal neovascularization [CNV], drusen, and DME detection), Noor Eye Hospital (AMD and DME detection), OCTDL (6 disease detection), and OPTIMA9C (4 disease detection and AMD staging).
In Harvard Glaucoma, for glaucoma detection, and OCTID (4 disease detection), MIRAGE also outperformed the other approaches, but no statistically significant differences were found.
On the other hand, MIRAGE showed consistently lower performance than other SOTA approaches in OLIVES, for DME and DR discrimination.
These results can be attributed to the distribution of diseases in the pretraining dataset (see Table~\ref{tab:diagnostic_data} in \textit{Methods}).
While our in-house pretraining dataset contains a high prevalence of AMD, other diseases, such as diabetes (which was highly represented in the RETFound pretraining dataset), are scarcely represented.
In particular, at least 85\% of the OCT scans used to train RETFound were from the Moorfields diabetic image dataset (MEH-MIDAS), while less than 2\% of the scans in our dataset were from diabetic patients.
Further analysis of the results (Supplementary Table~\ref{sup:tab:olives_first_last})
shows that MIRAGE performs equally well on the first scans of the patients, which are treatment-na\"ive.
Performance declines on the later scans, which are more likely to be affected by treatment, which introduces a data shift to which our model is less robust.
Notwithstanding, the overall performance of MIRAGE on OCT-based tasks was still significantly better than that of the other models, with an average AUROC of 95.59\% across all OCT tasks, outperforming the second-best model, RETFound, by a significant margin of 1.15 pp ($p < 0.001$).
Similar differences were observed for the other metrics, with MIRAGE achieving an average AP of 92.99\% (+1.64 pp, $p < 0.001$) and BAcc of 84.01\% (+2.51 pp, $p < 0.001$).

For the SLO tasks, MIRAGE significantly outperformed SL-IN and DINOv2 on the OPTIMA9C dataset, with an AUROC of 93.40\% and more than $1.5$ pp difference ($p < 0.001$).
Our model also outperformed the state of the art on the OLIVES dataset, but the differences were not statistically significant.
On average, MIRAGE achieved an AUROC of 83.74\% (+1.15 pp over the second model, $p < 0.05$), AP of 75.39\% (+0.54 pp), and BAcc of 61.13\% (+2.34 pp, $p < 0.01$).
Consequently, our MIRAGE model demonstrated significantly superior performance in both OCT- and SLO-based diagnosis and staging.

\subsection{Cross-dataset OCT classification performance}

To further assess the generalization capabilities of MIRAGE, we evaluated its performance in cross-dataset scenarios and compared it to the performance of the other SOTA models used in the previous experiments.
The results of this evaluation reflect how well the models can adapt to new, unseen data, which is a key indicator of their robustness and potential applicability in real-world clinical settings.

For classification, we evaluated the models tuned on the Noor Eye Hospital~\cite{rasti2017macular} dataset (with AMD and DME classes) on a combined external dataset consisting of the UMN~\cite{rashno2018fully} (AMD and DME) and Duke Srinivasan~\cite{srinivasan2014fully} (DME and iAMD) datasets.
In addition, we performed the inverse evaluation, tuning the models on the UMN + Duke Srinivasan dataset and evaluating them on the Noor Eye Hospital dataset.
As in the previous experiments, we used linear probing to tune the models for a specific classification task.
The results of this evaluation, shown in Figure~\ref{fig:cross_cls} and detailed in Supplementary Table~\ref{sup:tab:sota_class_external},
demonstrate the superior performance of our model over the other SOTA models, especially DINOv2 and RETFound.
Significant performance improvements were observed for Noor Eye Hospital, with MIRAGE outperforming SL-IN by 4.74 pp ($p < 0.01$) in terms of AUROC.
Smaller improvements were observed for UMN + Duke Srinivasan, with MIRAGE outperforming SL-IN by 2.26 pp.
These results highlight the robustness and adaptability of our model in handling domain shifts and generalizing effectively to new datasets.

\begin{figure}[h]
    \centering
    \includegraphics[width=0.65\linewidth]{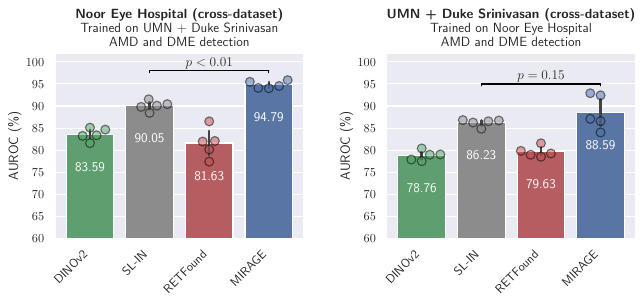}
    \caption{%
        \textbf{Cross-dataset evaluation results for OCT classification tasks.}
        For each task, we trained the models (all based on ViT-Large) with linear probing using five different random seeds.
        We then evaluated these models on the full external dataset and obtained five replicas from which the statistics were derived.
        The error bars show the standard deviation, while the colored bars show the mean AUROC.
        The performance of the two best models was compared using the one-tailed Student's \textit{t}-test, with the resulting \textit{p}-values indicated in the figure.
        MIRAGE outperforms all other models in both cases, with significant differences on Noor Eye Hospital dataset.
    }%
    \label{fig:cross_cls}
\end{figure}

\subsection{Segmentation of retinal lesions and layers}
\label{sec:seg_results}

\begin{figure}[t!]
    \centering
    \includegraphics[width=\linewidth]{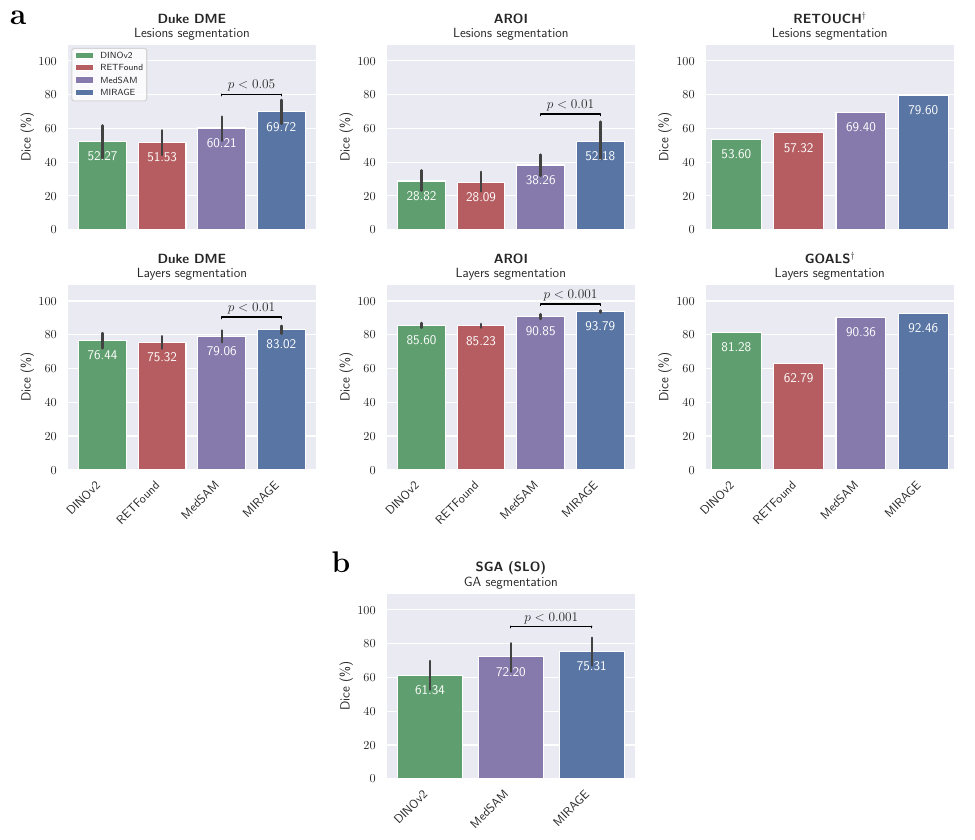}
    \caption{%
        \textbf{Performance comparison of MIRAGE and SOTA FMs at the retinal lesion and layer segmentation tasks.}
        \textbf{a} OCT segmentation.
        \textbf{b} SLO segmentation.
        The error bars show the patient-wise standard deviation, while the colored bars show the mean Dice score.
        The dagger symbol ($\dagger$) indicates that the patient information was not available, so the standard deviation is not shown.
        In each case, the performance of the two best models was compared to see if there were statistically significant differences.
        The \textit{p}-value is calculated using the one-tailed Student's \textit{t}-test and is indicated in the figure.
        MIRAGE outperforms all other FMs on all datasets, with significant differences in all cases for which the standard deviation was available.
    }\label{fig:seg_results}
\end{figure}

We utilized four public and one in-house diverse datasets to validate the adaptability of our model to lesion and layer segmentation tasks in both OCT and SLO images.
Specifically, we used the Duke DME~\cite{chiu2015kernel}, AROI~\cite{melinscak2021aroi}, RETOUCH~\cite{bogunovic2019retouch}, and GOALS~\cite{fang2022dataset} datasets for lesion and layer segmentation in OCT, and the SGA dataset~\cite{bui2022fundus} (in-house) for the segmentation of geographic atrophy (GA) in SLO.
For adapting the model to the segmentation tasks, we used a decoder-only fine-tuning strategy, where a ConvNeXt-based~\cite{liu2022convnext} decoder was trained on top of the pretrained ViT encoder~\cite{bachmann2022multimae}, which is kept frozen during downstream fine-tuning.
The ConvNeXt decoder was chosen based on previous work~\cite{bachmann2022multimae} and preliminary experiments (Supplementary Table~\ref{sup:tab:ablation_decoder})
showing that it is well suited for medical image segmentation.
This efficient strategy allows us to evaluate how well the different pretrained encoders (and thus the corresponding pretraining approaches) are suited for OCT and SLO segmentation tasks while achieving accurate segmentations.
In addition to comparing our model with the DINOv2 and RETFound models previously used for classification, we also compare it with MedSAM~\cite{ma2024segment}, a state-of-the-art ViT-based medical image segmentation FM.
All models were trained simultaneously for layer and lesion segmentation in datasets where both tasks were available (AROI and Duke DME).
However, the results are reported separately for a more detailed analysis.

Figure~\ref{fig:seg_results} shows the performance of the models on the OCT and SLO segmentation tasks in terms of the average Dice score (calculated at the patient level) for each dataset.
The average performance of the models in terms of Dice score, intersection over union (IoU),
and 95\textsuperscript{th} percentile Hausdorff distance (HD95)
across all OCT and SLO datasets is summarized in Table~\ref{tab:average_sota_seg}.
HD95 is a distance-based metric measured in pixels, with lower values indicating better performance.
The detailed results for each dataset are presented in the Supplementary Table~\ref{sup:tab:sota_seg}.
Since the official evaluation server of the RETOUCH challenge was used to obtain the results, and it provides only the average Dice and absolute volume difference (AVD) values, RETOUCH was excluded from the calculation of the average metrics except for the Dice score.

\begin{table}[tbp]
    \caption{%
        \textbf{Average performance of MIRAGE and SOTA FMs on segmentation tasks.}
        Average and standard deviation were calculated across all different datasets.
        No standard deviation was calculated for SLO segmentation tasks because only one dataset was available.
        The performance of the two best models was compared using the Wilcoxon signed-rank test, with the resulting \textit{p}-values indicated in the table by asterisks (*$p<0.05$).
        The best results are in bold, while the second best are underlined.
    }
    \label{tab:average_sota_seg}
    \centering
    \begin{tabular}{llllll}
        \toprule

        \textbf{Modality}
        & \textbf{Model}
        & \textbf{Dice}
        & \textbf{IoU}
        & \textbf{HD95} $\downarrow$
        \\

        \midrule

        \multirow{4}{*}{\textbf{OCT}}
        & DINOv2
        & 63.00 $\pm$ 19.99
        & 52.68 $\pm$ 21.29
        & 42.67 $\pm$ 22.80
        \\

        & RETFound
        & 60.05 $\pm$ 18.15
        & 47.73 $\pm$ 19.58
        & 57.20 $\pm$ 40.57
        \\

        & MedSAM
        & \underline{71.36 $\pm$ 18.37}
        & \underline{60.86 $\pm$ 21.98}
        & \underline{26.47 $\pm$ 20.34}
        \\

        & MIRAGE
        & \textbf{78.46 $\pm$ 14.26}*
        & \textbf{68.24 $\pm$ 18.40}*
        & \textbf{19.61 $\pm$ 16.87}*
        \\

        \midrule

        \multirow{3}{*}{\textbf{SLO}}
        & DINOv2
        & 61.34
        & 49.13
        & 210.02
        \\

        & MedSAM
        & \underline{72.20}
        & \underline{62.22}
        & \underline{182.22}
        \\

        & MIRAGE
        & \textbf{75.31}
        & \textbf{65.81}
        & \textbf{164.42}
        \\

        \bottomrule
    \end{tabular}
\end{table}

In all tasks, our model significantly outperformed the other SOTA FM models, including the specialized medical image segmentation FM MedSAM.
The greatest improvements were observed for the segmentation of retinal lesions in OCT datasets, where MIRAGE achieved Dice scores of 69.72\%, 52.18\%, and 79.60\% on Duke DME, AROI, and RETOUCH, respectively, outperforming the second best FM, MedSAM, by 13.92 pp ($p < 0.001$), 9.51 pp ($p < 0.001$), and 10.20 pp, respectively.
The same trend was observed for the other metrics (see Supplementary Table~\ref{sup:tab:sota_seg}).
No statistical analysis was performed for RETOUCH because the sample-wise results were not available, as the results were obtained using the official evaluation server of the challenge.
The performance of both DINOv2 and RETFound was overall significantly lower than that of MIRAGE and MedSAM, and remained generally low on all datasets.

MIRAGE was also the best performing model for layer segmentation, achieving Dice scores of 83.02\%, 93.79\%, and 92.46\% on Duke DME, AROI, and GOALS, respectively.
In all cases, the differences with respect to the second best model, MedSAM, were greater than 2 pp, with significant differences ($p < 0.001$) in all cases for which the patient-wise results were available (see Figure~\ref{fig:seg_results}a).

Across all OCT segmentation tasks, MIRAGE achieved an average Dice score of 78.46\%, significantly outperforming MedSAM, the second-best model, which achieved a Dice score of 71.36\% ($p < 0.05$).
Significant differences were also observed for the other metrics.
In SLO image segmentation (Figure~\ref{fig:seg_results}b), similar results were observed, where MIRAGE outperformed MedSAM by 3.11 pp ($p < 0.001$), achieving an average Dice score of 75.31\% in GA segmentation.
Our model also outperformed MedSAM and the other FMs by a significant margin in terms of IoU, and HD95.

\subsection{Cross-dataset OCT segmentation performance}
\label{sec:cross_seg}

\begin{figure}[tbp]
    \centering
    \includegraphics[width=0.32\linewidth]{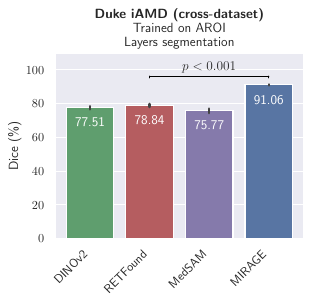}
    \caption{%
        \textbf{Cross-dataset evaluation results for OCT segmentation.}
        The error bars show the patient-wise standard deviation, while the colored bars show the mean Dice score.
        The performance of the two best models was compared to see if there were statistically significant differences.
        The \textit{p}-value is calculated using the one-tailed Student's \textit{t}-test and is indicated in the figure.
        MIRAGE significantly outperforms all other FMs.
    }%
    \label{fig:cross_seg}
\end{figure}

For cross-dataset segmentation evaluation, all models trained on the AROI dataset, which includes segmentation of four retinal layers and three lesions, were tested on the large Duke iAMD dataset~\cite{farsiu2014quantitative}.
The three layer classes in the Duke iAMD dataset are composites of the layers present in AROI.
The results, shown in Figure~\ref{fig:cross_seg} and detailed in Supplementary Table~\ref{sup:tab:sota_seg},
demonstrate the superior performance of our model over the other SOTA models, with significant improvements ($p<0.001$) of 13.55, 12.22, and 15.29 pp in terms of Dice score over DINOv2, RETFound, and MedSAM, respectively.
These results underscore the robustness and generalization capabilities of our model compared to existing general and segmentation-specific FMs, such as MedSAM, which was trained for medical image segmentation tasks.
In fact, while MedSAM was the second best model on the AROI dataset, it achieved the worst performance in the cross-dataset evaluation on Duke iAMD.
These results highlight the limitations of in-dataset evaluation and the importance of cross-dataset evaluation for effective model validation.

\subsection{Segmentation performance comparison with specialist models}

To further assess the relative segmentation capabilities of our model beyond FMs, we compared the performance of MIRAGE with that of specialist models on the same datasets.
Specifically, we evaluated SwinUNETR-V2~\cite{he2023swinunetrv2}, MedNeXt~\cite{roy2023mednext}, TransUNet~\cite{chen2024transunet}, and nnUNet version 2~\cite{isensee2021nnunet,isensee2024nnunet2} (from now on referred to as nnUNet).
SwinUNETR-V2, MedNeXt, and TransUNet are state-of-the-art models for medical image segmentation based on the Swin Transformer~\cite{liu2021swin}, ConvNeXt~\cite{liu2022convnext}, and Transformer~\cite{vaswani2017attention} architectures, respectively.
On the other hand, nnUNet is a fully automated deep learning framework, winner of many medical image segmentation challenges~\cite{isensee2021nnunet}, based on the successful U-Net~\cite{ronneberger2015unet} convolutional neural network (CNN) architecture~\cite{he2021structured,fazekas2023segmentation,morano2024rrwnet,azad2024medical}.
For a more fair comparison, we also trained MIRAGE on the segmentation tasks using full fine-tuning (FFT), the same strategy used by the competing models.
FFT involves training the entire model (both the ViT encoder and the ConvNeXt decoder) on the target dataset.
While this results in a more computationally expensive training process, it usually leads to better performance, especially when the target data differs substantially from the pretraining data.

\begin{figure}[tb]
    \centering
    \includegraphics[width=\linewidth]{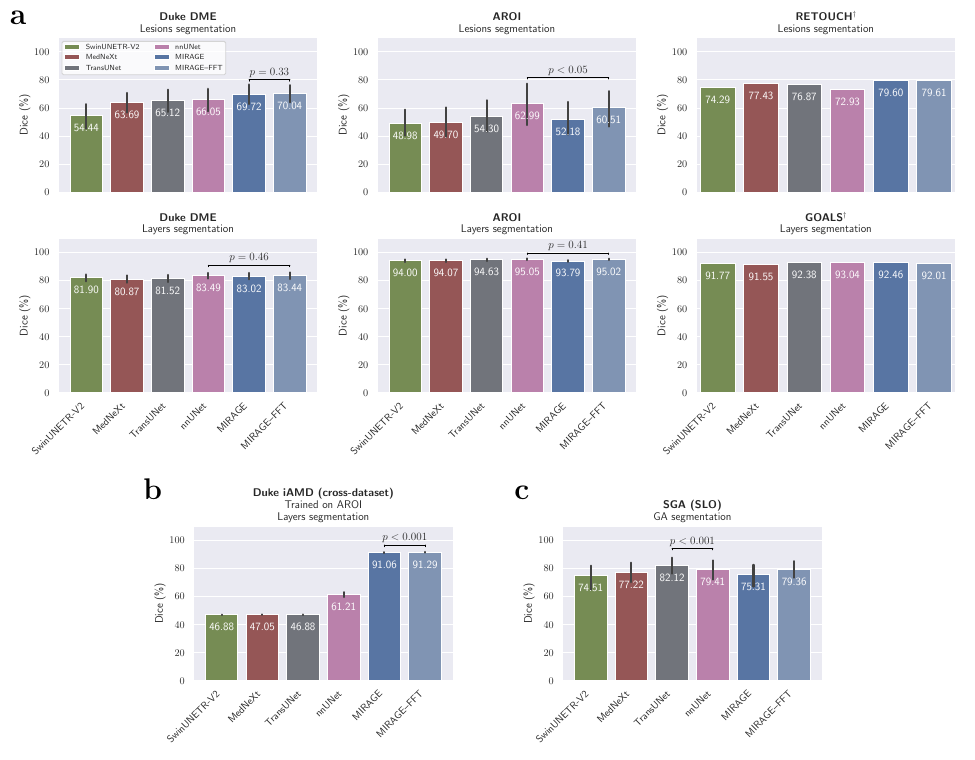}
    \caption{%
        \textbf{Performance comparison of MIRAGE and specialist models for retinal lesion and layer segmentation.}
        \textbf{a} OCT segmentation.
        \textbf{b} Cross-dataset OCT segmentation.
        \textbf{c} SLO segmentation.
        MIRAGE was evaluated using both linear probing and full fine tuning (FFT).
        The error bars show the patient-wise standard deviation, when available, while the colored bars show the mean Dice score.
        The dagger symbol ($\dagger$) indicates that the patient information was not available, so the standard deviation is not shown.
        In each case, the performance of the two best models was compared to see if there were statistically significant differences.
        The \textit{p}-value is calculated using the one-tailed Student's \textit{t}-test and is indicated in the figure.
        MIRAGE performs similarly to the specialist models on most in-domain tasks, while significantly outperforming them on the cross-dataset task.
    }\label{fig:seg_results_specialist}
\end{figure}

The results of this comparison are shown in Figure~\ref{fig:seg_results_specialist}.
In particular, Figure~\ref{fig:seg_results_specialist}a shows the performance of the models on the OCT segmentation tasks; Figure~\ref{fig:seg_results_specialist}b, the cross-dataset evaluation results for OCT segmentation; and Figure~\ref{fig:seg_results_specialist}c, the performance on the SLO segmentation tasks.
The average performance of the models in terms of the Dice score, IoU, and HD95
across all OCT and SLO datasets is shown in Table~\ref{tab:average_sota_seg_specialist}.
Detailed results for each dataset are presented in Supplementary Table~\ref{sup:tab:specialist_seg}.

\begin{table}[tbp]
    \caption{%
        \textbf{Average performance of MIRAGE and specialist models on segmentation tasks.}
        The standard deviation was calculated across all different datasets.
        No standard deviation was calculated for SLO segmentation tasks because only one dataset was available.
        The best results are in bold, while the second best are underlined.
    }
    \label{tab:average_sota_seg_specialist}
    \centering
    \begin{tabular}{llllll}
        \toprule

        \textbf{Modality}
        & \textbf{Model}
        & \textbf{Dice}
        & \textbf{IoU}
        & \textbf{HD95} $\downarrow$
        \\

        \midrule

    \multirow{6}{*}{\textbf{OCT}}
    & SwinUNETR-V2
    & 74.23 $\pm$ 17.26
    & 64.02 $\pm$ 22.09
    & 36.04 $\pm$ 40.01
    \\

    & MedNeXt
    & 76.22 $\pm$ 15.48
    & 65.73 $\pm$ 19.78
    & 27.73 $\pm$ 27.80
    \\

    & TransUNet
    & 77.47 $\pm$ 14.28
    & 67.58 $\pm$ 18.84
    & 21.31 $\pm$ 18.88
    \\

    & nnUNet
    & \underline{78.92 $\pm$ 12.49}
    & \textbf{70.39 $\pm$ 17.06}
    & 20.71 $\pm$ 19.41
    \\

    & MIRAGE
    & 78.46 $\pm$ 14.26
    & 68.24 $\pm$ 18.40
    & \underline{19.61 $\pm$ 16.87}
    \\

    & MIRAGE--FFT
    & \textbf{80.10 $\pm$ 11.98}
    & \underline{70.24 $\pm$ 16.52}
    & \textbf{19.26 $\pm$ 17.45}
    \\

    \midrule

    & SwinUNETR-V2
    & 46.88 $\pm$ 1.72
    & 43.90 $\pm$ 2.70
    & 201.81 $\pm$ 24.71
    \\

    & MedNeXt
    & 47.05 $\pm$ 1.47
    & 44.05 $\pm$ 2.34
    & 203.56 $\pm$ 29.73
    \\

    \multirow{1}{*}{\textbf{OCT}}
    & TransUNet
    & 46.88 $\pm$ 1.59
    & 43.76 $\pm$ 2.61
    & 194.12 $\pm$ 36.45
    \\

    {\small cross-dataset}
    & nnUNet
    & 61.21 $\pm$ 15.33
    & 55.35 $\pm$ 12.98
    & 97.95 $\pm$ 60.82
    \\

    & MIRAGE
    & \underline{91.06 $\pm$ 2.43}
    & \underline{84.62 $\pm$ 3.38}
    & \underline{4.62 $\pm$ 4.52}
    \\

    & MIRAGE--FFT
    & \textbf{91.29 $\pm$ 2.14}
    & \textbf{84.92 $\pm$ 3.15}
    & \textbf{3.94 $\pm$ 2.27}
    \\

    \midrule

    \multirow{6}{*}{\textbf{SLO}}
    & SwinUNETR-V2
    & 74.51
    & 65.09
    & 169.03
    \\

    & MedNeXt
    & 77.22
    & 67.87
    & 157.15
    \\

    & TransUNet
    & \textbf{82.12}
    & \textbf{73.69}
    & \textbf{127.19}
    \\

    & nnUNet
    & \underline{79.41}
    & \underline{71.33}
    & 136.56
    \\

    & MIRAGE
    & 75.31
    & 65.81
    & 164.42
    \\

    & MIRAGE--FFT
    & 79.36
    & 70.03
    & \underline{136.16}
    \\

        \bottomrule
    \end{tabular}
\end{table}

In general, MIRAGE performed similarly to specialist models on the OCT segmentation tasks, with the only exceptions being AROI (lesions) and the RETOUCH datasets (see Figure~\ref{fig:seg_results_specialist}a).
In the former, nnUNet outperformed MIRAGE by 2.48 pp ($p < 0.05$) and MIRAGE--FFT by 10.81 pp ($p < 0.01$) in terms of Dice score, while in the latter, MIRAGE outperformed nnUNet by 6.67 pp.
On the other hand, MIRAGE significantly outperformed all specialist models on the cross-dataset OCT segmentation task (Figure~\ref{fig:seg_results_specialist}b), with Dice scores of 91.06\% and 91.29\% for MIRAGE and MIRAGE--FFT, respectively, compared to 61.21\% for nnUNet and around 47\% for the other specialist models.
This corresponds to a difference of about 30 pp between MIRAGE and the best specialist model.
These results demonstrate the very limited generalizability of nnUNet and other specialist models to new datasets, and the superior generalization capabilities of our model.

Small improvements were observed when using the FFT strategy, with the largest differences seen in AROI (lesions), where the Dice score increased by 8.33 pp, from 52.18\% to 60.51\%.
This improvement, combined with small increases in other datasets such as Duke DME (layers) and AROI (layers), resulted in an average Dice score of 80.10\% across all OCT segmentation tasks, surpassing the second best model, nnUNet, by 1.18 pp, and MIRAGE (with decoder-only tuning) by 1.64 pp.
Improvements were also observed in terms of IoU and HD95.

In the segmentation of GA in SLO (Figure~\ref{fig:seg_results_specialist}b),
TransUNet achieved the best performance, with a Dice score of 82.12\%, followed by nnUNet with 79.41\% and MIRAGE--FFT with 79.36\%.
It also outperformed the other models in terms of IoU and HD95.
In this case, it is important to note that only one binary segmentation task was available for evaluation, limiting the scope of the comparison and the conclusions that can be drawn.
In addition, this model performed consistently worse than MIRAGE, MIRAGE--FFT, and nnUNet on the OCT datasets.

The results presented in this section show that our model is not only superior to all other SOTA foundation models
in terms of segmentation performance (see Table~\ref{tab:average_sota_seg}), but also competitive with specialized models on the same segmentation tasks.
In addition, the results of the cross-dataset evaluation show that our model significantly outperforms the specialist models under domain shifts, highlighting its greater generalization capabilities and potential for real-world applications.

\subsection{Qualitative analysis of segmentation results}

\begin{figure}[tbp]
    \centering
    \includegraphics[width=\linewidth]{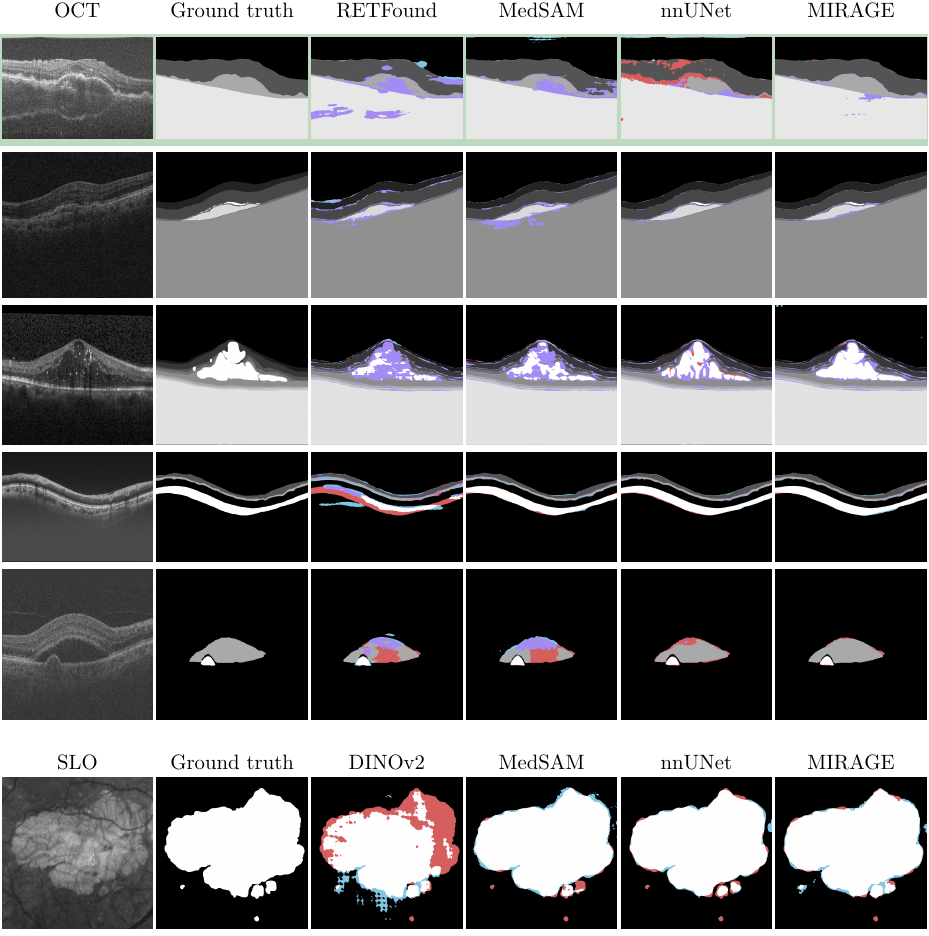}
    \caption{\textbf{Examples of segmentations from different models.}
    The examples belong, from top to bottom, to the following datasets: Duke iAMD (cross-dataset evaluation, highlighted in \textcolor{darkgreen}{\textbf{green}}), AROI, Duke DME, GOALS, RETOUCH, and SGA.
    True positives are depicted in the \textbf{grayscale} value of the class; false background pixels, in \textcolor{darkred}{\textbf{red}}; false lesion or layer pixels, in \textcolor{darkercyan}{\textbf{cyan}}; and true but wrongly classified lesion or layer pixels, in \textcolor{darkviolet}{\textbf{violet}}.
    The results show that our model produces precise and accurate segmentations, rarely missing or misclassifying lesions and layers.
    The quality of our segmentations is appreciably higher than that of nnUNet in the cross-dataset evaluation (first row), and similar in the other datasets.
    Compared to MedSAM and RETFound, the differences are more pronounced, as these models often misclassify pathological regions.
    }%
    \label{fig:qualitative_seg}
\end{figure}

In addition to the quantitative evaluations presented in the previous sections, we also examined the qualitative performance of the models on the segmentation tasks by visualizing the segmentations produced by each model.
Figure~\ref{fig:qualitative_seg} shows examples of segmentations produced by MIRAGE and the SOTA models RETFound (for OCT tasks), DINOv2 (for SLO tasks), MedSAM, and nnUNet on the different OCT and SLO segmentation datasets.

The qualitative results are consistent with the quantitative results, showing that our model produces precise and accurate segmentations, equal to or better than nnUNet, and significantly better than the other SOTA foundation models, here represented by RETFound, DINOv2 and MedSAM.
The higher quality of the segmentations produced by MIRAGE is particularly evident in the cross-dataset setting (Figure~\ref{fig:qualitative_seg}, first row highlighted in green), where the models were evaluated on Duke iAMD but trained on AROI.
In this cross-dataset scenario, MIRAGE demonstrates superior generalization capabilities, producing segmentations that are more accurate than those produced by the other models.
In particular, nnUNet produces an incomplete segmentation of the layers, misclassifying some of the pixels (in red in Figure~\ref{fig:qualitative_seg}) as background, while RETFound and MedSAM, while correctly distinguishing the layers from the background in most cases, assign large areas of the image (in violet) to the wrong layer class.
For the other datasets, the differences between MIRAGE and nnUNet are subtle, with both models producing generally accurate segmentations.
On the other hand, RETFound shows poor overall performance, with many layers or lesions incorrectly classified as background and assigned to the wrong layer or lesion type, especially in regions with pathological signs.
MedSAM, while generally more accurate than RETFound, shares some of the same problems, with the exception of the GOALS and SGA datasets, where the segmentations are nearly as accurate as those produced by MIRAGE and nnUNet.

\subsection{Effect of pseudo-labeling}

We evaluated the impact of using retinal layer pseudo-labels during pretraining by comparing the downstream performance of a ViT model pretrained using only OCT images with that of a model trained using OCT images and pseudo-labels of retinal layers (OCT+Layers).
The models were pretrained using the MultiMAE approach~\cite{bachmann2022multimae} and evaluated on the OCT classification and segmentation tasks using linear probing (LP).
While this strategy is suboptimal for segmentation~{\cite{oquab2024dinov2}}, it was chosen in this case to minimize the impact of the decoder on the results and focus on the pseudo-labeling effect.
In particular, we follow the approach proposed in DINOv2~{\cite{oquab2024dinov2}}, which consists of training a linear layer to classify each patch token and then upsampling the resulting map to full resolution.
The results of this analysis are presented in Table~\ref{tab:pseudo_labeling}, while the detailed results are presented in the Supplementary Tables~\ref{sup:tab:pseudo_labeling_class} and \ref{sup:tab:pseudo_labeling_seg}
for classification and segmentation tasks, respectively.
As shown in the table, the model pretrained with OCT images and pseudo-labels for the retinal layers (OCT+Layers) significantly outperformed the model pretrained with OCT images alone on the OCT classification and segmentation tasks.
These results demonstrate the positive impact of using pseudo-labels during pretraining.

\begin{table}[h]
    \centering
    \caption{%
        \textbf{Effect of pseudo-labeling.}
        Comparison of the performance of a ViT-Base model trained on OCT alone, and OCT with pseudo-labels for the retinal layers (OCT+Layers) using the MultiMAE approach~\cite{bachmann2022multimae} on OCT classification and segmentation tasks with linear probing.
        The best results are in bold.
        In each case, the performance of the two models across all datasets for every metric was compared using the Wilcoxon signed-rank test (*: $p<0.05$, ***: $p < 0.001$).
    }%
    \label{tab:pseudo_labeling}
    \resizebox{\textwidth}{!}{%
    \begin{tabular}{llllll}
        \toprule

        \textbf{Modalities}
            & \multicolumn{2}{c}{\textbf{Classification}}
            & \multicolumn{2}{c}{\textbf{Segmentation (LP)}}
        \\
        \cmidrule(lr){2-3}
        \cmidrule(lr){4-5}

        & \multicolumn{1}{c}{\textbf{AUROC}}
        & \multicolumn{1}{c}{\textbf{BAcc}}
        & \multicolumn{1}{c}{\textbf{Dice}}
        & \multicolumn{1}{c}{\textbf{HD95} $\downarrow$}
        \\

        \midrule

    OCT
    & 93.75 $\pm$ 7.99
    & 79.77 $\pm$ 12.76
    & 66.37 $\pm$ 13.68
    & 40.83 $\pm$ 23.58
    \\

    OCT+Layers
    & \textbf{95.44 $\pm$ 6.00}***
    & \textbf{82.81 $\pm$ 12.62}***
    & \textbf{69.07 $\pm$ 14.21}*
    & \textbf{34.33 $\pm$ 17.73}
    \\

        \bottomrule
    \end{tabular}
    }
\end{table}

\subsection{Effectiveness of domain-specific multimodal pretraining}

We evaluated the effect of domain-specific and multimodal pretraining by comparing the performance of MIRAGE, pretrained on our multimodal VIBES dataset, with that of domain-specific unimodal models pretrained on unimodal retinal imaging data (VIBES-OCT and VIBES-SLO) and general multimodal models pretrained on multimodal data from ImageNet (Multi-IN).
The comparison was made for ViT models pretrained with MAE~{\cite{he2022mae}} and MultiMAE~{\cite{bachmann2022multimae}} on unimodal and multimodal pretraining setups, respectively, and evaluated on the OCT and SLO classification and segmentation tasks using linear probing (LP).
Similar to the previous experiment, we use this strategy to minimize the impact of the decoder on the results and focus on the pretraining effect.
Since the only available MultiMAE-pretrained model is the ViT-Base, and for the sake of computational efficiency, we used the ViT-Base architecture for this analysis.
By comparing the results of the models pretrained with a single modality with those of the multimodal MIRAGE, we were able to effectively measure the importance of integrating the different imaging modalities during pretraining.
The average results of this analysis are presented in Table~\ref{tab:multimodal_pretraining}, while the detailed results are presented in the Supplementary Tables~\ref{sup:tab:impact_pretraining} and \ref{sup:tab:impact_pretraining_seg}
for classification and segmentation tasks, respectively.

\begin{table}[tbp]
    \centering
    \caption{
        \textbf{Effect of domain-specific and multimodal pretraining.}
        Comparison of the impact of domain-specific and multimodal pretraining on the performance of a ViT-Base model in classification and segmentation tasks with linear probing.
        The best results are in bold, and the second best are underlined.
        In each case, the performance of the two best models across all datasets was compared to determine if there were statistically significant differences in average performance.
        For this, the Wilcoxon signed-rank test was used.
        The results are presented in the table, with the significance level indicated by asterisks (*: $p<0.05$, **: $p<0.01$, \mbox{***: $p < 0.001$}).
    }%
    \label{tab:multimodal_pretraining}
    \resizebox{\textwidth}{!}{%
        \begin{tabular}{lllllll}
            \toprule

            \textbf{Tuning}
            & \textbf{Model}
            & \textbf{Pretraining}
            & \multicolumn{2}{c}{\textbf{Classification}}
            & \multicolumn{2}{c}{\textbf{Segmentation (LP)}}
            \\

            \cmidrule(lr){4-5}
            \cmidrule(lr){6-7}

            \textbf{modality}
            &
            & \textbf{dataset}
            & \multicolumn{1}{c}{\textbf{AUROC}}
            & \multicolumn{1}{c}{\textbf{BAcc}}
            & \multicolumn{1}{c}{\textbf{Dice}}
            & \multicolumn{1}{c}{\textbf{HD95} $\downarrow$}
            \\

            \midrule

            \multirow{3}{*}{\textbf{OCT}}
            & MultiMAE
            & Multi-IN
            & 91.01 $\pm$ 9.60
            & 75.73 $\pm$ 13.06
            & 41.09 $\pm$ 26.13
            & 80.46 $\pm$ 47.68
            \\

            & MAE-OCT
            & VIBES-OCT
            & \underline{93.75 $\pm$ 7.99}
            & \underline{79.77 $\pm$ 12.76}
            & \underline{66.37 $\pm$ 13.68}
            & \underline{40.83 $\pm$ 23.58}
            \\

            & MIRAGE
            & VIBES
            & \textbf{94.52 $\pm$ 6.97}***
            & \textbf{81.84 $\pm$ 11.24}***
            & \textbf{69.63 $\pm$ 14.60}*
            & \textbf{34.33 $\pm$ 21.42}
            \\

            \midrule

            \multirow{3}{*}{\textbf{SLO}}
            & MultiMAE
            & Multi-IN
            & \underline{83.52 $\pm$ 6.44}
            & \underline{56.13 $\pm$ 9.72}
            & 68.69
            & 187.07
            \\

            & MAE-SLO
            & VIBES-SLO
            & 74.67 $\pm$ 10.11
            & 51.61 $\pm$ 15.42
            & \underline{70.63} 
            & \underline{174.96}
            \\

            & MIRAGE
            & VIBES
            & \textbf{85.66 $\pm$ 7.19}**
            & \textbf{61.55 $\pm$ 10.28}**
            & \textbf{72.24} 
            & \textbf{166.25}
            \\

            \bottomrule
        \end{tabular}
    }
\end{table}

As shown in the table, the models pretrained with domain-specific data (OCT, SLO, and layers) significantly outperformed those pretrained with a single modality (OCT or SLO) or multimodal data from ImageNet.
Specifically, our model achieved the highest scores across all metrics for OCT-based models, with an average AUROC of 94.52\%, BAcc of 81.84\%, and segmentation Dice of 69.63\% and HD95 of 34.33 pixels, with significant improvements over MultiMAE and MAE-OCT.
Similarly, our model also showed significantly superior performance on SLO classification tasks, achieving an average AUROC of 85.66\%, BAcc of 61.55\%, and consistently higher segmentation performance, with Dice of 72.24\% and HD95 of 166.25 pixels.
These results highlight the substantial benefit of pretraining with domain-specific multimodal data on downstream performance regardless of the modality (OCT/SLO) and the type of task (classification/segmentation).

\subsection{Impact of model capacity}

We explored the impact of model capacity on the performance of MIRAGE by comparing the ViT-Base ($\approx86$M parameters) and ViT-Large ($\approx307$M parameters) architectures across all tasks.
The average results of this analysis are presented in Table~\ref{tab:large_vs_base}, along with the results of the best SOTA FMs in each case: RETFound for OCT classification, SL-IN for SLO classification, and MedSAM for OCT and SLO segmentation.
The detailed results of the ViT-Base version of MIRAGE for each dataset are presented in the Supplementary Tables~\ref{sup:tab:large_vs_base_class} and \ref{sup:tab:large_vs_base_seg}
for classification and segmentation, respectively.

In classification tasks, the performance of the ViT-Large model was significantly superior to that of the ViT-Base model in OCT, while ViT-Base performed better in SLO.
For instance, the ViT-Large model achieved an AUROC of 95.59\% for OCT classification, significantly higher than the 94.52\% achieved by the ViT-Base model.
For SLO classification, the differences were not significant.
In segmentation tasks, the ViT-Large and ViT-Base models achieved similar performance, although the ViT-Large model performed slightly better in most cases in terms of Dice score.
In all cases, MIRAGE-Base outperformed the SOTA models, namely RETFound, SL-IN (both based on ViT-Large) and MedSAM (based on ViT-Base), for classification and segmentation, respectively.

These results demonstrate that the effectiveness of MIRAGE is not dependent on model capacity, as both the ViT-Base and ViT-Large models achieved satisfactory performance in most tasks.
Furthermore, they show that, although the ViT-Large model generally outperforms the ViT-Base model (especially in OCT classification tasks), the performance difference is usually small.
Thus, our MIRAGE-Base model may be an appropriate choice in scenarios where computational resources are limited.

\begin{table}[h]
    \caption{\textbf{Effect of model capacity on downstream performance.}
    Average performance of MIRAGE based on ViT-Base or ViT-Large across all OCT and SLO classification and segmentation tasks.
    The best SOTA results are also shown in gray for reference: RETFound for OCT classification, SL-IN for SLO classification, and MedSAM for OCT and SLO segmentation.
    The best results are in bold, and the second best are underlined.
    The Wilcoxon signed-rank test was used to compare the performance of the two models in each task, with the resulting \textit{p}-values shown in the table (***: $p < 0.001$).
    }
    \label{tab:large_vs_base}
    \centering
    \resizebox{\textwidth}{!}{%
        \begin{tabular}{llllll}
            \toprule

            \textbf{Modality}
            & \textbf{Model}
            & \multicolumn{2}{c}{\textbf{Classification}}
            & \multicolumn{2}{c}{\textbf{Segmentation}}
            \\

            \cmidrule(lr){3-4}
            \cmidrule(lr){5-6}

            &
            & \multicolumn{1}{c}{\textbf{AUROC}}
            & \multicolumn{1}{c}{\textbf{BAcc}}
            & \multicolumn{1}{c}{\textbf{Dice}}
            & \multicolumn{1}{c}{\textbf{HD95} $\downarrow$}
            \\

            \midrule

            \multirow{3}{*}{\textbf{OCT}}
            & SOTA
            & 94.44 $\pm$ 6.66
            & 81.50 $\pm$ 11.48
            & 71.36 $\pm$ 18.37
            & 26.47 $\pm$ 20.34
            \\

            & MIRAGE-Base
            & \underline{94.52 $\pm$ 6.97}
            & \underline{81.84 $\pm$ 11.24}
            & \underline{77.30 $\pm$ 16.84}
            & \underline{20.96 $\pm$ 19.17}
            \\

            & MIRAGE-Large
            & \textbf{95.59 $\pm$ 5.80}***
            & \textbf{84.01 $\pm$ 10.51}***
            & \textbf{78.46 $\pm$ 14.26}
            & \textbf{19.61 $\pm$ 16.87}
            \\

            \midrule

            \multirow{3}{*}{\textbf{SLO}}
            & SOTA
            & 82.59 $\pm$ 9.52
            & 58.79 $\pm$ 6.21
            & 72.20
            & 182.22
            \\

            & MIRAGE-Base
            & \textbf{85.66 $\pm$ 7.19}
            & \textbf{61.55 $\pm$ 10.28}
            & \underline{74.47}
            & \textbf{161.13}
            \\

            & MIRAGE-Large
            & \underline{83.74 $\pm$ 9.67}
            & \underline{61.13 $\pm$ 5.65}
            & \textbf{75.31}
            & \underline{164.42}
            \\

            \bottomrule
        \end{tabular}
    }
\end{table}

\section{Discussion}

This work introduces MIRAGE, a robust multimodal foundation model (FM) for comprehensive retinal image analysis, and extensively evaluates its generalization capabilities across a range of diagnostic, staging, and segmentation tasks.
MIRAGE is based on the ViT architecture~\cite{dosovitskiy2020image} and is trained using SSL~\cite{morano2023self,bachmann2022multimae} on a large dataset of paired OCT and SLO images, along with automatically generated pseudo-labels of retinal layers~\cite{garvin2008intraretinal,antony2011automated}.
The model can be adapted, with minimal tuning, to various retinal imaging tasks, including both classification and segmentation in OCT and SLO images.
Our extensive evaluations demonstrate the superiority of MIRAGE over existing FMs and its potential for research and clinical applications.

Unlike existing FMs, which are either strictly unimodal~\cite{zhou2023retfound,qiu2024visionfm}, naively mix multiple modalities in the same batch~\cite{ma2024segment,cai2024uni4eye++,shi2024eyefound} (ignoring multimodal complementarity), or focus on image-level contrastive learning on partially paired datasets~\cite{shi2024eyeclip}, MIRAGE was pretrained using a \textit{fully-paired} dataset of multimodal retinal images in a multimodal MAE setting~\cite{bachmann2022multimae,morano2023self}.
This pretext task involves reconstructing all input multimodal images from their highly masked versions, requiring the model to infer masked information from the limited visible image patches of the same image and the other modalities.
To allow the model to process any subset of the input modalities in inference, we use a sample strategy that randomly and non-uniformly selects the patches from the different modalities.
Thus, during training, the patches received by the model for a given sample can all be from the same modality (and none from the others) or a mix from different modalities.
This pretraining strategy has several advantages.
First, it allows MIRAGE to be used for both OCT and SLO image analysis, since the model is pretrained with both modalities simultaneously.
This contrasts with existing FMs for ophthalmology, such as RETFound~\cite{zhou2023retfound} and VisionFM~\cite{qiu2024visionfm}, which feature different models for each modality.
Second, it makes the model more robust to domain shifts, as it learns more general features that are not specific to a single modality, but common to all input modalities.
Finally, through the multimodal reconstruction task, the different modalities provide fine-grained, complementary supervisory signals, as lesions or other pathological characteristics are often visible in both modalities, but in different manners.
In this way, the model can learn to associate the appearance of pathological signs in one modality with the abnormal patterns in the other modalities, ultimately improving detection.
All of these factors make MIRAGE the first truly multimodal FM for retinal image analysis and contribute to the learning of robust multimodal representations that are generalizable to a wide range of tasks.

To evaluate the generalization capabilities and the quality of the learned representations of MIRAGE, we conducted extensive evaluations on a number of clinically relevant tasks.
These tasks include diagnosis and staging from OCT and SLO images and, unlike previous works~\cite{zhou2023retfound,silva2023flair}, retinal lesion and layer segmentation.
For the classification of diseases and stages in OCT and SLO images, MIRAGE was compared to other SOTA FMs, including RETFound~\cite{zhou2023retfound} and DINOv2~\cite{caron2021emerging}, as well as models trained using supervised learning on the ImageNet-21k dataset of natural images (SL-IN)~\cite{dosovitskiy2020image}.
RETFound is a unimodal FM trained with MAE~\cite{he2022mae} on a private dataset of 700k OCT images.
DINOv2 was pretrained on a private dataset of 142M natural images using an SSL approach based on self-distillation.
On the other hand, the ImageNet-21k dataset contains 14M images with 21k categories of natural objects.
Similar to previous work on FMs~\cite{singh2022flava,silva2023flair,kirillov2023segment,ma2024segment,oquab2024dinov2,chen2024towards}, all models were evaluated using linear probing on the target datasets.
This evaluation strategy involves optimizing a single linear classifier (only about 2k parameters) on top of the frozen pretrained encoder.
This allows a fair comparison of the quality of the representations learned by the models during pretraining without the need for extensive fine-tuning.
The results on 11 different datasets (10 of which are public) (Figure~\ref{fig:cls_sota_plots} and Table~\ref{tab:cls_sota_avg}) show that MIRAGE significantly outperforms the other models in both OCT and SLO diagnostic and staging tasks.
Among the different tasks, MIRAGE was particularly effective in diagnosing and staging AMD (see, for example, the results on Duke iAMD or 9C from Figure~\ref{fig:cls_sota_plots}), the most prevalent retinal disease worldwide (affecting 8.7\% of the global population)~\cite{guymer2023age}.
On the other hand, the lowest relative performance of MIRAGE was observed in the discrimination of diabetes-related lesions (see the results on the OLIVES dataset from Figure~\ref{fig:cls_sota_plots}).
These results are consistent with the distribution of the diseases in the pretraining dataset, which contains a larger proportion of AMD samples ($>50\%$) compared to other diseases such as diabetes ($<2\%$).
In contrast, at least 85\% of the OCT scans used to train RETFound were from the Moorfields diabetic image dataset (MEH-MIDAS).
Further analysis of the results (Supplementary Table~\ref{sup:tab:olives_first_last})
showed that, while MIRAGE was less robust than RETFound to the data shift introduced by patient treatment in OLIVES, it achieved similar performance on treatment-na\"ive baseline scans of the patients, underscoring the importance of pretraining data in the generalization capabilities of models to very specialized tasks.
However, an additional cross-dataset evaluation (Figure~\ref{fig:cross_cls}) shows that MIRAGE is more robust to general dataset shifts than the other models.

In contrast to previous work~\cite{zhou2023retfound,silva2023flair}, we also evaluated the performance of MIRAGE in retinal lesion and layer segmentation.
We believe that evaluating FMs in these tasks is essential, as they represent the most common use cases of AI in clinical practice, due to a high demand for OCT biomarker detection and quantification.
In addition, such evaluations can provide meaningful insights into the capabilities of FMs in capturing the internal structure of the data and fine-grained but relevant details.
Our model was again compared to DINOv2 and RETFound, but even more importantly to MedSAM~\cite{ma2024segment}, a FM for medical image segmentation.
Similarly to the classification tasks, the encoder of each model was frozen, and only the initial linear projection layers and a convolutional decoder were fine-tuned on the target datasets.
This adds up to about 12M parameters, which is less than half the total parameters of a U-Net model ($\approx31$M parameters)~\cite{ronneberger2015unet}, the most commonly used model for medical image segmentation~\cite{isensee2021nnunet,azad2024medical}.
The results on five public and one private dataset (Figure~\ref{fig:seg_results} and Table~\ref{tab:average_sota_seg}) show that MIRAGE significantly outperforms the other FMs in both lesion and layer segmentation tasks in OCT and SLO images.
The differences were especially pronounced in the cross-dataset evaluation (Figure~\ref{fig:cross_seg}).
To further contextualize these results, we also compared MIRAGE to the specialized segmentation models nnUNet~\cite{isensee2021nnunet} (version 2), SwinUNETR~\cite{he2023swinunetrv2}, MedNeXt~\cite{roy2023mednext}, and TransUNet~\cite{chen2024transunet}.
As shown in Figure~\ref{fig:seg_results_specialist} and Table~\ref{tab:average_sota_seg_specialist}, MIRAGE performs on par with the specialist models on the within-dataset evaluation, and significantly outperforms them in the cross-dataset evaluation.
The nnUNet framework was specifically designed to maximize performance on a single dataset by tuning both the architecture and the hyperparameters, as well as the data augmentation strategies and other training settings, to that specific dataset.
While this approach can lead to high performance on the target dataset, it often comes at the cost of generalization to new datasets, as demonstrated by the results of this evaluation.
While the other specialist models are not as tuned in detail as nnUNet for the specific tasks, they are still designed to maximize performance on existing medical image segmentation benchmarks, which are usually well-curated and homogeneous, with similar training and testing distributions.
Our results show that these models are not well suited for real-world clinical applications, where the ability to generalize to new, unseen data is essential.
In contrast, our model, pretrained on a diverse set of multimodal data, demonstrates superior generalization capabilities, highlighting its potential for real-world clinical applications.
The qualitative analysis of the segmentations (Figure~\ref{fig:qualitative_seg}) is consistent with the previous quantitative results.
While nnUNet, the most performant and generalizable specialist model, generally performs well, it fails more often in the presence of pathological signs, such as large fluid pockets.
MIRAGE, on the other hand, is more robust to these changes, resulting in more accurate segmentations.
The differences with the other FMs are more pronounced, and the higher quality of the segmentations produced by MIRAGE is evident in the visualizations.
These results demonstrate the greater robustness of MIRAGE over both existing FMs and highly specialized models in retinal lesion and layer segmentation tasks, and therefore its greater potential for clinical applications.

To further investigate the benefits of using retinal layer pseudo-labels during pretraining, we compared the performance of a model pretrained with OCT images alone to that of a model pretrained with OCT images and pseudo-labels for the retinal layers.
The results of this analysis (Table~\ref{tab:pseudo_labeling}) show that the model pretrained with OCT images and pseudo-labels for the retinal layers significantly outperformed the model pretrained with OCT images alone in both OCT classification and segmentation tasks.
These results demonstrate the positive impact of using pseudo-labels during pretraining.

Similarly, to further verify the benefits of domain-specific multimodal pretraining, we compared the downstream performance of MIRAGE to equivalent models pretrained on only one modality (i.e., OCT or SLO) and to models pretrained on multimodal natural images.
The results (Table~\ref{tab:multimodal_pretraining}) show that MIRAGE significantly outperforms the other models in most tasks, with significant improvements on average.
This demonstrates the potential of multimodal pretraining on paired imaging data for the development of robust and generalizable medical FMs.

Previous work has shown that the performance of FMs can be significantly improved by increasing the model capacity~\cite{dosovitskiy2020image,caron2021emerging,oquab2024dinov2,kirillov2023segment}.
For this reason, most FMs are based on high-capacity models such as the ViT-Large architecture~\cite{devlin2018bert,zhou2023retfound,chen2024towards}.
However, large models such as ViT-Large are computationally expensive, making them impractical or unusable for low-resource settings or real-time applications.
To mitigate this problem, we have also developed a smaller version of MIRAGE based on the ViT-Base architecture, which has
about 72\% fewer parameters than the ViT-Large model.
Our results (Table~\ref{tab:large_vs_base}) show that, while the performance of the ViT-Large version of MIRAGE is generally better than the ViT-Base version (with significant improvements in OCT classification), the differences are not substantial, and the ViT-Base version still outperforms the other FMs by most metrics.
This suggests that the ViT-Base version of MIRAGE may be a more practical choice for low-resource settings, while still providing state-of-the-art performance in retinal image analysis.

In recent years, the development of FMs has become a central topic in AI research in general~\cite{brown2020language,caron2021emerging,oquab2024dinov2,kirillov2023segment}, and in medicine in particular~\cite{zhou2023retfound,chen2024towards,silva2023flair,moor2023foundation,zhang2024challenges}.
Pivotal works such as RETFound~\cite{zhou2023retfound} and GMAI~\cite{moor2023foundation} have demonstrated the potential of FMs for medical image analysis and for democratizing access to medical AI.
The development of MIRAGE further extends this line of research by demonstrating the benefits of multimodal pretraining on paired medical image data for the development of more robust and generalizable FMs.
Our MIRAGE model can be easily adapted to various retinal imaging tasks, including classification and segmentation in both OCT and SLO images, with minimal tuning.
Through the use of efficient tuning strategies and our proposed evaluation benchmark, we showed the superior performance of MIRAGE in a wide spectrum of downstream retinal image analysis tasks.
However, the applications of MIRAGE are not limited to these tasks.
Similar to other FMs, MIRAGE could be effectively used as a backbone in any method that requires the extraction of high-level features from OCT or SLO images.
More than a performant classification or segmentation method, MIRAGE is a general tool for retinal image analysis with a wide range of applications.
In this sense, we believe that integrating domain-specific FMs like MIRAGE into AI-based systems could significantly improve their accuracy and generalizability, ultimately leading to higher-quality AI models for healthcare.
This contrasts with the typical development of performant but non-generalizable models, which leads to skepticism about the benefits of AI and limits its adoption~\cite{zhou2023retfound}.
In line with previous work~\cite{zhou2023retfound,ma2024segment}, we have made MIRAGE publicly available for research, with the goal of improving reproducibility and accelerating the progress of AI in retinal image analysis.

While this work has comprehensively validated MIRAGE in several tasks, there are still limitations and challenges that need to be explored in future work.
First, all the data used to develop MIRAGE is from a single center (Medical University of Vienna, Austria), and most of the samples are from patients with AMD, with smaller number of samples from other diseases, such as diabetes.
Therefore, it is important to pursue a larger dataset by including retinal images across the world with a more diverse and realistic data distribution.
Second, the pretraining of MIRAGE and the evaluation of its OCT diagnostic performance were limited to the central 2D B-scans of the OCT volumes.
This was done, in line with previous work~\cite{zhou2023retfound}, to avoid the high computational cost of processing the entire 3D OCT derived from the use of ViT models.
However, this is a limitation, as the use of 3D information is essential in clinical practice, and previous work~\cite{george2020attention,lin2021assessing,oghbaie2024vlfatrollout,liu2024simultaneous} has shown that it can significantly improve the performance of models in retinal image analysis tasks.
Thus, research on developing efficient ways to incorporate 3D information into the pretraining and evaluation of FMs is an important direction for future work, that could ultimately lead to the development of more accurate FMs.
Third, retinal layer segmentations for pretraining were generated using a graph-theoretic approach and image processing techniques\cite{garvin2008intraretinal,antony2011automated}.
While this algorithm is OCT device-agnostic, and guarantees the topological correctness of the segmentations, its accurate performance is limited to healthy images.
For this reason, layer segmentations are considered as \textit{weak labels} or \textit{pseduo-labels}.
It is likely that the segmentation performance on pathological images could be improved by using more sophisticated segmentation algorithms, which could even incorporate the segmentation of pathological structures such as fluid pockets.
Such segmentations would likely further improve the representation learning of MIRAGE and its downstream performance.
In particular, we plan to investigate the use of topology-aware deep learning algorithms such as SD-LayerNet~\cite{fazekas2025sdlayernet}.
Fourth, while MIRAGE has shown strong performance in segmentation tasks, especially in the cross-dataset evaluation, it is slightly less performant than some task-specific SOTA models~\cite{fazekas2025sdlayernet,li2021multiscale}.
This is likely due to the fact that these methods are specifically designed for the segmentation task at hand, while MIRAGE was tuned using a general and straightforward semantic segmentation setup.
This was done to minimize the impact of modules other than the pretrained encoder on the results, and to focus primarily on the evaluation of the pretraining strategy.
Nevertheless, MIRAGE is not inherently incompatible with more sophisticated segmentation methods and architectures, and it is likely that its performance could be improved by using any of these innovations.
Exploring this question represents an interesting area for future work.
Fifth, while the linear probing and decoder-only fine-tuning strategies allowed us to effectively evaluate MIRAGE and SOTA FMs and achieve strong performance in most tasks, more sophisticated adapters or tuning strategies could improve downstream performance.
A common example is Low-Rank Adapters (LoRA)~\cite{hu2021lora}, which enables the adaptation of large models to downstream tasks by tuning a small set of modules interleaved with the original architecture.
This approach has been shown to improve the performance of large models in OCT medical image analysis~\cite{fazekas2023adapting}.
In addition, it is very likely that fully fine-tuning the models could improve downstream classification performance if enough data is available~\cite{steiner2022howto,zhou2023retfound}, as is the case for segmentation.
Unfortunately, there are not as many large-scale datasets available for OCT classification as there are for other imaging modalities, such as natural images, and, despite successful advances in the field~\cite{}, it is not guaranteed that full fine-tuning would not lead to overfitting on small datasets.
Investigating how to adapt MIRAGE to maximize performance while maintaining generalizability represents an interesting area for future work.
Finally, the current version of MIRAGE only uses image information from the OCT and SLO modalities, and pseudo-labels of retinal layers.
The inclusion of color fundus photography (CFP), which is considered to be more informative and much more widely used than SLO (especially in developing countries), would make MIRAGE even more useful.
In this study, it was not possible to include CFP since, to the best of our knowledge, no large dataset of \textit{paired} OCT--CFP data is available (either publicly or in our clinic), and this pairing is essential for multimodal pretraining.
This is because, unlike OCT and SLO images (which are commonly acquired together by modern OCT devices), CFP images are either acquired using a separate imaging device or not acquired at all, when fundus is examined primarily with a slit lamp.
This makes it difficult to collect paired OCT--CFP data.
In addition, incorporating data such as clinical notes or demographic information during pretraining could provide further meaningful feedback to the model, ultimately improving representations.
In this sense, we believe that MIRAGE could open the path towards more effective vision language models (VLMs) for OCT.
Furthermore, while generalist large language models (LLMs) have demonstrated strong language capabilities for ophthalmology, they have also demonstrated limited vision capabilities for feature detection~{\cite{jeong2024medical,antaki2024vision}}.
Therefore, incorporating a robust and generalizable vision encoder such as MIRAGE could improve the feature detection and thus the overall performance.
These limitations and future directions represent interesting research opportunities in the field and areas for improvement of the proposed FM.

In conclusion, we proposed MIRAGE, the first multimodal vision foundation model for OCT and SLO image analysis.
To demonstrate its effectiveness and efficiency in adapting to various healthcare applications, we evaluated MIRAGE and other FMs on a newly proposed benchmark.
Unlike previous evaluation approaches, our benchmark includes segmentation tasks in addition to diagnosis and staging, and is composed of a heterogeneous set of 19 tasks from 16 datasets.
The results on our benchmark show significant performance improvements of MIRAGE over existing FMs in detecting and staging ocular diseases and segmenting retinal lesions and layers in OCT and SLO images.
In light of these results, we believe that MIRAGE can serve as an effective multimodal foundation model for retinal image analysis, with potential applications in research and clinical practice.
In addition, we hope that our benchmark will help to better assess the capability of FMs for OCT, facilitating future comparisons and tracking progress in the field.

\section{Methods}\label{sec:methods}

\subsection{Pretraining dataset}

The pretraining dataset, VIBES~\cite{gerendas2022validation}, described in detail in Tables \ref{tab:pt_dataset} and \ref{tab:diagnostic_data}, consists of 261\,184 samples from 42\,082 patients and 75\,653 unique eyes.
This was obtained after filtering out samples with very poor SLO quality from the original dataset of 350\,005 samples.
Each sample consists of a triplet of paired OCT and SLO images and pseudo-labels of retinal layers.
Pseudo-labels of retinal layers were generated specifically for the purpose of this study using an automated segmentation algorithm~\cite{garvin2008intraretinal,antony2011automated}.
All scans were acquired between April 2007 and April 2021 at the Macula Clinic, Department of Ophthalmology and Optometry, Medical University of Vienna (MedUni Wien).
The MedUni Wien Ethics Committee approved the \textit{post hoc} analysis of the dataset (EK-Nr: 2095/2018), and the requirement for informed consent was waived due to the retrospective nature of the study and the de-identification of data, which was performed in accordance with institutional policies. The work adhered to the tenets of the Declaration of Helsinki and MedUni Wien standards of good scientific practice.

Images were acquired with Cirrus (Carl Zeiss Meditec, Dublin, CA, USA) and Spectralis devices (Heidelberg Engineering, Heidelberg, Germany).
The image resolution for Cirrus B-scans is always $512 \times 1024$ pixels (height $\times$ width).
For Spectralis, the B-scan height is always 496 pixels, but the width varies between 512, 768, and 1024 pixels.
Following previous work~\cite{zhou2023retfound}, only the central B-scans of the 3D OCT volumes were used for the analysis.
The dataset includes a wide range of retinal diseases, lesions, and conditions (see Table~\ref{tab:diagnostic_data}), with the most common being cataract, choroidal neovascularization (CNV), age-related macular degeneration (AMD), retinal vascular occlusion (RVO), and glaucoma.

\begin{table}[t]
    \caption{\textbf{Details of the pretraining dataset.}
    Our VIBES~\cite{gerendas2022validation} pretraining dataset consists of 261\,184 samples (OCT and SLO scan pairs) from 42\,082 patients and 75\,653 unique eyes.
    }%
    \label{tab:pt_dataset}
    \centering
    \resizebox{\textwidth}{!}{%
        
\begin{tabular}{rr|rr|rr|rr|rr}
    \toprule

    \multicolumn{2}{c|}{Age at scan*}
    & \multicolumn{2}{c|}{Gender*}
    & \multicolumn{2}{c|}{Vendor}
    & \multicolumn{2}{c|}{\# B-scans}
    & \multicolumn{2}{c}{Field of view (mm\textsuperscript{2})} \\

    \midrule

    0-20 & (0.8\%)
    & Male & (41.2\%)
    & Cirrus & (54.4\%)
    & 128 & (49.8\%)
    & $6\times6$ & (83.8\%) \\

    20-40 & (5.6\%)
    & Female & (58.8\%)
    & Spectralis & (45.6\%)
    & 25 & (15.0\%)
    & $6\times5$ & (5.1\%) \\

    40-60 & (16.0\%)
    & & & &
    & 49 & (14.5\%)
    & $6\times4$ & (1.7\%) \\

    60-80 & (52.8\%)
    & & & &
    & 19 & (5.1\%)
    & $4\times4$ & (1.1\%) \\

    80-103 & (24.8\%)
    & & & &
    & 200 & (4.6\%)
    & $9\times8$ & (1.0\%) \\

    & & & & &
    & 97 & (4.0\%)
    & $6\times7$ & (1.0\%) \\

    & & & & &
    & Other & (7.0\%)
    & Other & (6.3\%) \\

    \bottomrule

    \multicolumn{10}{l}{*Estimate based on the information available for $54.8\%$ of the samples.}

\end{tabular}%

    }
\end{table}

\begin{table}[tbp]
    \caption{%
        \textbf{Distribution of diseases, lesions, and conditions in the pretraining dataset.}
        Estimate based on diagnostic labels available for 30.8\% of the samples.
    }%
    \label{tab:diagnostic_data}
    \centering
    \resizebox{0.67\textwidth}{!}{%
        
\begin{tabular}{lr}
    \toprule

    \textbf{Diagnostic label}
    & \textbf{\%}
    \\

    \midrule

Cataract(a)/lens(opacity) & 57.65 \\
\midrule
Choroidal neovascularization & 35.58 \\
\midrule
Age-related macular degeneration & 33.69 \\
\midrule
Other & 19.67 \\
\midrule
Various (nevus, melanoma, endophthalmitis, retinitis, etc.) & 13.56 \\
\midrule
Anti-VEGF therapy & 8.47 \\
\midrule
Unhealthy retina & 8.09 \\
\midrule
Healthy retina restricted & 7.89 \\
\midrule
Healthy retina & 7.75 \\
\midrule
Retinal vascular occlusion & 7.44 \\
\midrule
Glaucoma & 6.56 \\
\midrule
Chorioretinopathia centralis serosa & 4.61 \\
\midrule
Epiretinal membrane & 4.48 \\
\midrule
Fibrosis, scar & 4.17 \\
\midrule
Refractive anomalies & 4.01 \\
\midrule
Vitreous body & 3.50 \\
\midrule
Geographic atrophy & 3.33 \\
\midrule
Retinal surgery & 3.32 \\
\midrule
Laser intervention & 3.03 \\
\midrule
Fundus hypertonicus & 2.93 \\
\midrule
Cystic macular edema & 2.62 \\
\midrule
Macular hemorrhage & 2.50 \\
\midrule
Disciform macula degeneration (Junius-Khunt disease) & 1.99 \\
\midrule
Myopic CNV & 1.74 \\
\midrule
Pigment epithelial detachment & 1.71 \\
\midrule
Diabetes & 1.69 \\
\midrule
Macular telangiectasia & 1.39 \\
\midrule
Pattern dystrophy & 1.34 \\
\midrule
Macular hole, lamellar hole, NH defect & 1.12 \\
\midrule
Cortisone treatment & 1.09 \\
\midrule
Irvine Gass syndrome & 1.01 \\
\midrule
RPE tear & 0.99 \\
\midrule
Drusen papilla & 0.33 \\
\midrule
Coat, Best, and Stargardt diseases, Terson syndrome  & 0.33 \\
\midrule
Subretinal fluid & 0.16 \\

    \bottomrule

\end{tabular}

    }
\end{table}

\subsection{Benchmark datasets}

The evaluation benchmark consists of 14 publicly available datasets, including 9 for classification, 4 for segmentation, and one for both classification and segmentation.
To make the multimodal evaluation even more comprehensive, we also incorporated a private dataset for classification (including both OCT and SLO images) and a private dataset for lesion segmentation on SLO images.
This adds up to a total of 16 datasets and 19 different tasks.

The selected retinal disease classification datasets, listed in Table~\ref{tab:cls_datasets}, contain images from both normal eyes and from patients exhibiting signs of various retinal diseases and conditions, including glaucoma, diabetic macular edema (DME), early, intermediate, and advanced AMD ([e,i,a]AMD), retinal vein occlusion (RVO), geographic atrophy (GA), macular hole (MH), central serous retinopathy (CSR), epiretinal membrane (ERM), retinal artery occlusion (RAO), vitreomacular interface disease (VID), Stargardt disease, and choroidal neovascularization (CNV).
The scans were acquired using a diverse set of OCT devices: Spectralis, Cirrus, Triton (Topcon, Tokyo, Japan), RTVue XR (Optovue, Fremont, USA) or Bioptigen (Leica Microsystems, Wetzlar, Germany).
In addition, the datasets come from clinics in 6 different countries: USA, China, Austria, Iran, Russia, and India.
As was done for the pretraining dataset and in previous work~\cite{zhou2023retfound}, we restricted the analysis to the central B-scans of the OCT volumes.

\begin{table}[tb]
    \caption{\textbf{Details of the downstream classification datasets.}}%
    \label{tab:cls_datasets}
    \centering
    \resizebox{\textwidth}{!}{%
        
\begin{tabular}{l|l|rr|l|l}
    \toprule



    \textbf{Dataset} & \textbf{Modality} & \textbf{\# Samples} & \textbf{\# Patients} & \textbf{Classes (\# Samples)} & \textbf{Acquisition device} \\
    \midrule

    Duke iAMD~\cite{farsiu2014quantitative} & OCT
    & 383 & 383 & Control (115), iAMD (268) & Cirrus \\
    \midrule

    Duke Srinivasan~\cite{srinivasan2014fully} & OCT
    & 45 & 45 & Control (15), DME (15), iAMD (15) & Bioptigen \\ 
    \midrule

    GAMMA~\cite{wu2023gamma} & OCT
    & 100 & 100 & Control (50), Early glaucoma (26), & Triton \\ 
    & & & & Intermediate/advanced glaucoma (24) & \\
    \midrule

    Harvard Glaucoma~\cite{luo2023harvard} & OCT
    & 1\,000 & 1\,000 & Control (557), Glaucoma (443) & Cirrus \\
    \midrule

    Kermany~\cite{kermany2018dataset,kermany2018identifying} & OCT
    & 109\,309
    & 4,686
    & Control (51\,390), DME (11\,598), & Spectralis \\
    & & & & CNV (37\,455), Drusen (8\,866) & \\
    \midrule

    Noor Eye Hospital~\cite{rasti2017macular} & OCT  
    & 148 & 148 & Control (50), DME (50), AMD (48) & Spectralis \\
    \midrule

    OCTDL~\cite{kulyabin2024octdl} & OCT
    & 2\,064 & 821 & Control (332), AMD (1\,231), DME (147), & RTVue XR \\
    & & & & ERM (155), RAO (22), RVO (101), & \\
    & & & & VID (76) & \\
    \midrule

    OCTID~\cite{gholami2020octid} & OCT
    & 572 & - & Control (206), MH (102), AMD (55), & Cirrus \\
    & & & & CSR (102), DR (107) & \\
    \midrule

    OLIVES~\cite{prabhushankar2022olives} & OCT, SLO
    & 1590 & 96 & DME (931), DR (659) & Spectralis \\
    \midrule

    UMN~\cite{rashno2018fully} & OCT
    & 54 & 54 & DME (30), AMD (24) & Spectralis \\
    \midrule

    OPTIMA9C~\cite{oghbaie2024vlfatrollout} (in-house) & OCT, SLO
    & 4205 & 3652
    & Control (183), RVO (763), Stargardt (130),
    & Spectralis (61\%) \\
    & & & & DME (1\,091), iAMD (1\,128), GA (452), & Cirrus (38\%) \\
    & & & & CNV1 (99), CNV2 (83), CNV3 (276) & Triton (1\%) \\

    \bottomrule

\end{tabular}%

    }
    \raggedright
\end{table}

\begin{table}[tb]
    \caption{\textbf{Details of the downstream segmentation datasets.}}%
    \label{tab:seg_datasets}
    \centering
    \resizebox{\textwidth}{!}{%
        
\begin{tabular}{l|l|rr|l|l}
    \toprule




    \textbf{Dataset}
    & \textbf{Modality}
    &  \multicolumn{1}{r}{\textbf{\# Samples}}
    & \textbf{\# Patients}
    & \textbf{Classes} (excluding background)
    & \textbf{Devices} \\
    & & \multicolumn{1}{l}{\textbf{(\# B-scans/sample)}} & & & \\
    \midrule

    AROI~\cite{melinscak2021aroi} & OCT
    & 24 (128*) & 24 & Layers: ILM--IPL/INL, IPL/INL--RPE, & Cirrus \\
    & & & &  ~~~~~~~~~~~RPE--BM, Below BM& \\
    & & & & Lesions: Cyst, PED, SRF & \\
    \midrule

    Duke DME~\cite{chiu2015kernel} & OCT
    & 10 (11) & 10 & Layers: ILM, RNFL, GCL, IPL, INL, & Spectralis\\
    & & & & ~~~~~~~~~~~OPL, ONL, ISM, OS, RPE & \\
    & & & & Lesions: Fluid & \\
    \midrule

    Duke iAMD~\cite{farsiu2014quantitative} & OCT
    & 384 (100) & 384 & Layers: ILM--Inner RPEDC, & Cirrus \\
    & & & & ~~~~~~~~~~~Inner RPEDC--Outer BM, & \\
    & & & & ~~~~~~~~~~~Below BM & \\
    \midrule

    GOALS~\cite{fang2022dataset} & OCT
    & 100 & - & Layers: RNFL, GCIPL, Choroid & Triton
    \\
    \midrule

    RETOUCH~\cite{bogunovic2019retouch} & OCT
    & 112 (128) & 112 & Lesions: IRF, SRF, PED & Spectralis \\
    & & & & & Cirrus \\
    & & & & & Triton \\
    \midrule

    SGA~\cite{bui2022fundus} (in-house) & SLO
    & 965 & 100 & Lesions: GA & Spectralis
    \\

    \bottomrule

\end{tabular}%

    }
    \raggedright
    \footnotesize
    *Only an average of 47.3 B-scans were annotated.
\end{table}

We also compiled five retinal layer and lesion segmentation datasets to benchmark our model on retinal lesion and layer segmentation tasks (see Table~\ref{tab:seg_datasets}).
In addition to the public datasets, we included a private dataset with SLO images for the segmentation of GA.
These datasets cover a wide range of retinal lesions and layers, some of which are subsets or combinations of others.
In particular, the following lesions are included: cystoid edema (cyst), pigment epithelial detachment (PED), subretinal fluid (SRF), intraretinal fluid (IRF), and GA.
The datasets also include the following retinal layers:
inner limiting membrane (ILM),
retinal nerve fiber layer (RNFL),
retinal pigment epithelium (RPE),
retinal pigment epithelium-drusen complex (RPEDC),
ganglion cell layer (GCL),
inner plexiform layer (IPL),
inner nuclear layer (INL),
outer plexiform layer (OPL),
outer nuclear layer (ONL),
external limiting membrane (ELM),
inner segment myoid (ISM),
outer segment (OS),
ganglion cell-inner plexiform layer complex (GCIPL),
Bruch's membrane (BM),
choroid,
and choroid-sclera interface (below BM).
The scans were acquired with Cirrus, Spectralis, and Triton scanners at clinics in 5 different countries: USA, China, Austria, Croatia, and the Netherlands.
In contrast to the classification datasets, all the B-scans in the segmentation datasets are used for training and evaluating the models.
This was done following standard practices in the literature, where layer and lesion segmentation tasks are usually performed B-scan-wise, and not on the full volumes~\cite{roy2017relaynet,he2021structured,fazekas2023segmentation}.

More details on all these datasets are listed in the Supplementary Note~\ref{sup:sec:data}, with illustrative examples in Supplementary Figures~\ref{sup:fig:datasets} and \ref{sup:fig:datasets_seg}.
When training the models on the tasks listed above, we followed the standard train-test splits for the public datasets when it was available.
In cases where no standard splits were provided, we performed a random split into training, validation, and test sets using patient and label stratification to ensure balanced representation in each set.

\subsection{MIRAGE approach}

\begin{figure}[b!]
    \centering
    \includegraphics[width=\linewidth]{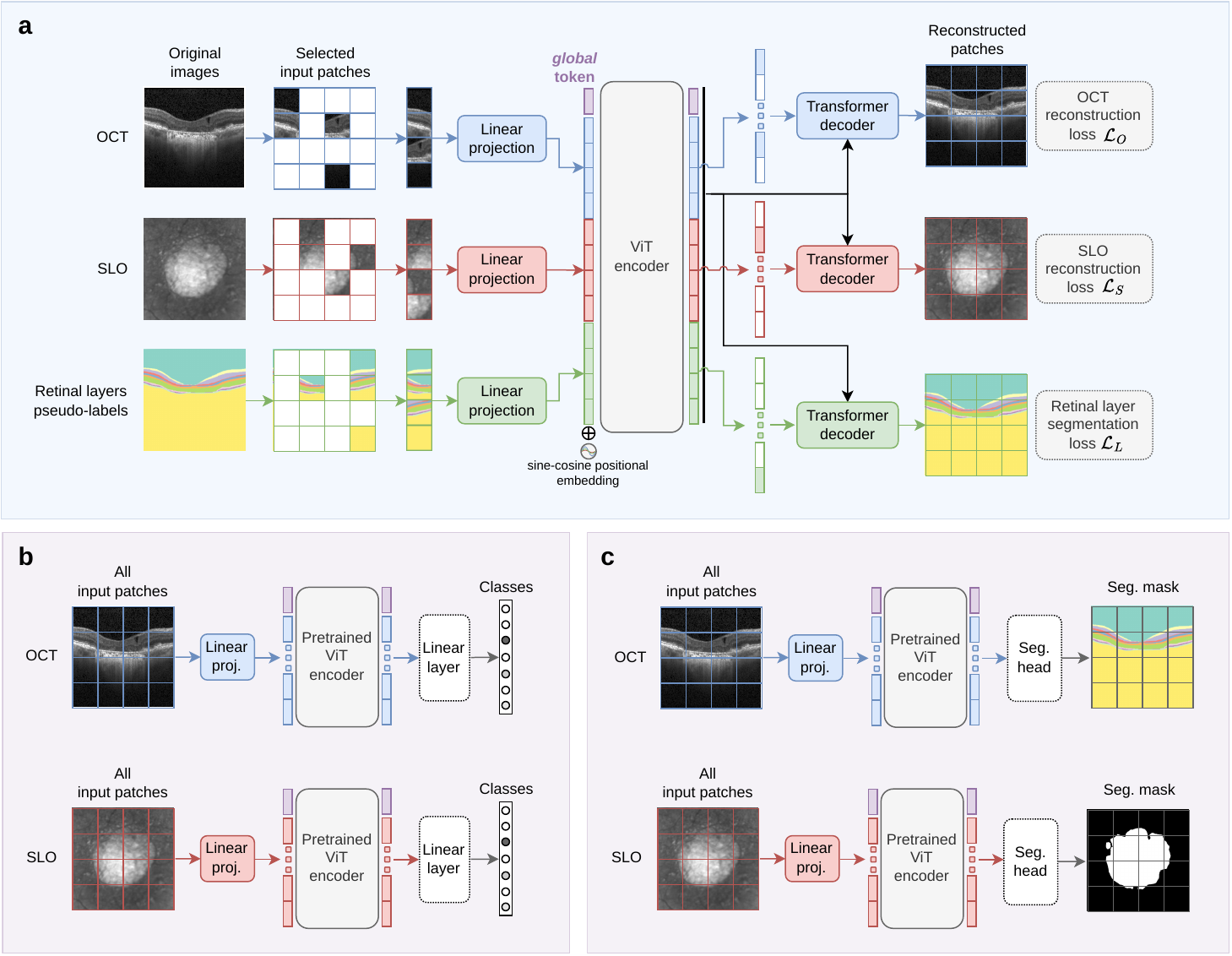}
    \caption{%
        \textbf{Overview of the approach used for training and tuning our multimodal foundation model.}
        \textbf{a} Multimodal pretraining.
        \textbf{b} Classification tuning.
        \textbf{c} Segmentation tuning.
        The approach consists of training a Vision Transformer (ViT) model on a large dataset of multimodal retinal images (including OCT, SLO, and pseudo-labels of retinal layers) by reconstructing the input images from a masked or partial view of them.
        Black arrows represent cross-attention operations between all encoded tokens and the modality tokens.
        The reconstruction loss $\mathcal{L} = \mathcal{L}_O + \mathcal{L}_N + \mathcal{L}_L$ on the masked tokens is used as the objective function.
        In fine-tuning, the model can be trained on downstream tasks by replacing the decoders used for pretraining with task-specific heads.
        Moreover, it accepts both OCT and SLO images as input during inference.
    }
    \label{fig:approach}
\end{figure}

The proposed framework for training and testing our foundation model, based on a multimodal MAE approach~\cite{bachmann2022multimae}, is illustrated in Figure~\ref{fig:approach}.
The model architecture is based on a ViT~\cite{dosovitskiy2020image} encoder and modality-specific linear projection layers and Transformer decoders.
The ViT architecture was chosen because of its wide adoption in the literature (especially for foundation models~\cite{he2022mae,oquab2024dinov2,bachmann2022multimae,zhou2023retfound,ma2024segment}) and its demonstrated effectiveness in a variety of medical image classification~\cite{zhou2023retfound} and segmentation tasks~\cite{ma2024segment}.
The modality-specific linear projection layers consist of a flattening operation followed by a linear layer.
The decoders are shallow transformer networks consisting of a linear projection layer followed by a cross-attention layer, a multi-layer perceptron (MLP), two Transformer blocks~\cite{vaswani2017attention}, and a linear projection layer followed by a reshape operation to reconstruct the patches.
The cross-attention operation (represented by black arrows in Figure~\ref{fig:approach}) is used to allow the model to leverage information from the other modalities.

The pretraining pretext task consists of reconstructing the input OCT, SLO, and retinal layer segmentation images from masked or partial versions of themselves.
For this purpose, the images are first divided into modality-specific patches, and then the proportion of non-masked patches per modality is determined by sampling from a symmetric Dirichlet distribution~\cite{bachmann2022multimae} with a concentration parameter $\alpha=1$.
Then, non-masked tokens are sampled uniformly at random without replacement.
This sampling ensures that the model receives both unimodal inputs (from one modality) and multimodal inputs (from two or three modalities) during training.
Thus, it can handle any subset of the input modalities during downstream tasks.
In Supplementary Note~\ref{sup:sec:masking},
we provide more details on this masking strategy and the selection of $\alpha$.

After the masking process, the unmasked visible patches are projected onto tokens using the modality-specific linear projection layers, concatenated, and passed through the encoder to generate the latent features.
An additional \textit{global} token with a learned embedding is added to the input sequence to provide a global context for the model, similar to the CLS token in ViT~\cite{dosovitskiy2020image}.
Patch tokens are then fed into the modality-specific decoders to reconstruct the previously masked patches.

A reconstruction loss that measures the difference between the original and reconstructed masked patches was used as the objective function.
Following prior work~\cite{he2022mae,bachmann2022multimae}, the loss is defined as the L2 distance for image patches and as cross-entropy loss (CE) for pseudo-labels of retinal layers.
The total training loss $\mathcal{L}$ is the sum of the losses for each modality so that
\begin{equation}
    \mathcal{L} = \mathcal{L}_O + \mathcal{L}_N + \mathcal{L}_L = \|\hat{x}_O, x_O\|_2 + \|\hat{x}_N, x_N\|_2 + \text{CE}(\hat{x}_L, x_L)~,
\end{equation}
where the subscripts $O$, $N$, and $L$ denote OCT, SLO, and retinal layers pseudo-labels, and $\hat{x}$ and $x$ are the predicted and ground truth patches, respectively.

Retinal layer segmentations for the pretraining dataset were generated using a graph-theoretic approach and image processing techniques~{\cite{garvin2008intraretinal,antony2011automated}}.
This algorithm is OCT device-agnostic and guarantees the topological correctness of the segmentations.
In this way, the model can effectively learn, during pretraining, the specific features of the retinal layers as well as the general, physical structure of the retina and the arrangement of these layers within that structure.

To adapt the model to downstream tasks, the shallow transformer decoders are replaced with task-specific heads (see Figure~\ref{fig:approach}, bottom).
For classification, they are replaced by a single linear layer followed by a softmax activation function.
For segmentation, they are replaced by a simple convolutional segmentation head based on ConvNeXt~\cite{liu2022convnext}.

\subsection{Network architecture}\label{sec:architecture}

\begin{figure}[tbph]
    \centering
    \includegraphics[width=0.7\linewidth]{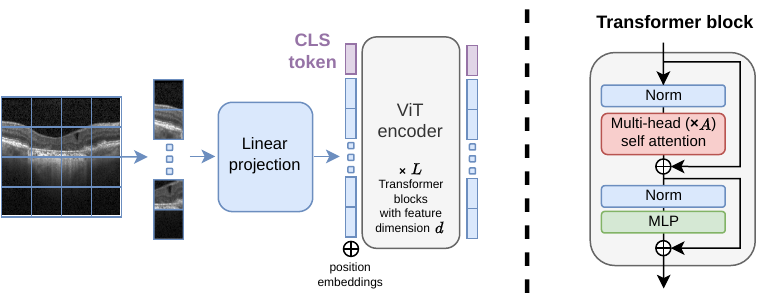}
    \caption{%
        \textbf{Overview of the ViT encoder.}
        The linear projection layer projects the input patches into embeddings of dimension $d$, which are then passed through a stack of $L$ Transformer blocks containing $A$ attention heads.
    }%
    \label{sup:fig:vit}
\end{figure}

The ViT backbone used in the experiments, shown in Figure~\ref{sup:fig:vit}, is the same as the one proposed by Dosovitskiy et al.~\cite{dosovitskiy2020image}.
The model consists of a linear projection layer followed by a positional embedding addition and a stack of $L$ Transformer blocks, each containing a multi-head self-attention mechanism and a feed-forward neural network, with a layer normalization and a residual connection around each sub-block.
The ViT-Base model has $L=12$ Transformer blocks, embedding dimension $d=768$, and $A=12$ attention heads ($\sim$86M parameters).
The ViT-Large model has $L=24$ Transformer blocks, embedding dimension $d=1024$, and $A=16$ attention heads ($\sim$307M parameters).

\begin{figure}[tbph]
    \centering
    \includegraphics[width=0.8\linewidth]{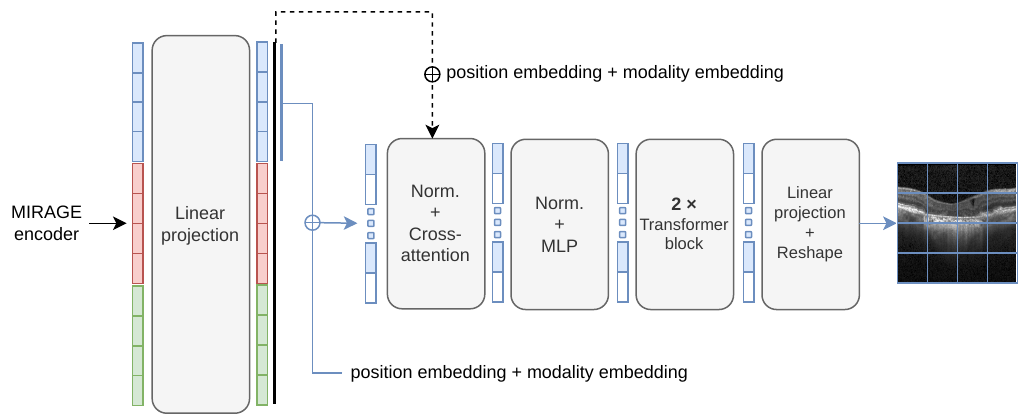}
    \caption{%
        \textbf{Overview of the modality-specific decoders.}
        Modality-specific features are fed into the transformer-based decoders along with the rest of the encoded tokens through cross-attention operations.
    }%
    \label{sup:fig:decoders}
\end{figure}

\begin{figure}[tbph]
    \centering
    \includegraphics[width=0.65\linewidth]{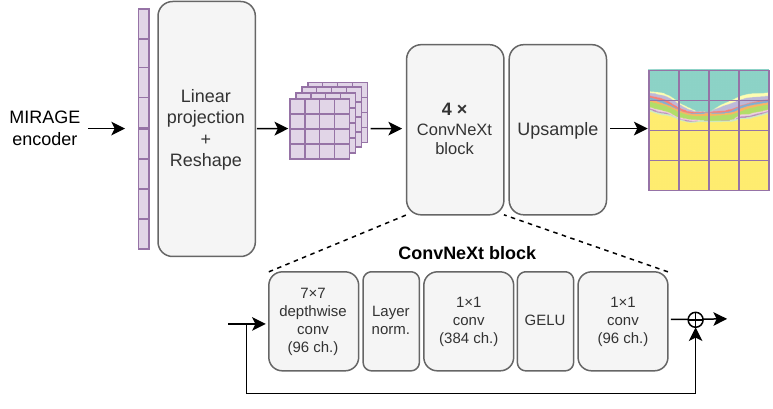}
    \caption{%
        \textbf{Overview of the ConvNeXt-based segmentation decoder.}
        The output tokens of the encoder are projected to a higher dimension and reshaped into a feature map, which is then processed by a series of ConvNeXt blocks and upsampled to full resolution.
    }%
    \label{sup:fig:convnext_decoder}
\end{figure}

For each modality, different linear projection layers are used to project the input patches into the embedding space of the ViT backbone.
The linear projection layers are implemented as a flatten operation followed by a fully-connected layer with $p^2$ input features and $d$ output features, where $p$ is the patch size and $d$ is the embedding dimension of the ViT backbone (768 and 1024 for ViT-Base and ViT-Large, respectively).
The rest of the linear projection layers in the model are implemented in the same way, only varying the number of input and output features when necessary.
Projected patches from the different modalities are concatenated into a sequence of tokens and given as input to the same Transformer encoder, as in the original ViT model described above.

The modality-specific decoders, depicted in Figure~\ref{sup:fig:decoders}, were implemented as shallow transformer networks~\cite{he2022mae}.
Each decoder consists of a linear projection layer followed by a cross-attention layer, a multi-layer perceptron (MLP), two Transformer blocks and a linear projection layer followed by a reshape operation to reconstruct the patches.
The cross-attention is performed between all encoded tokens (\textbf{K}, \textbf{V}) and the modality tokens (\textbf{Q}), allowing the decoder to integrate multimodal information.

For the classification experiments, we used a linear classifier head on top of the frozen ViT backbone to predict the class labels.
In particular, we used a global average pooling in the token dimension (excluding the global token) followed by a linear layer with $d$ input features and $C$ output classes, where $d$ is the embedding dimension of the ViT backbone and $C$ is the number of classes.
Finally, we applied a softmax activation function to obtain the class probabilities.

The segmentation head used in the segmentation experiments consists of the following operations (see Figure~\ref{sup:fig:convnext_decoder} for an overview).
First, a linear projection is applied to the output tokens of the encoder to increase their dimensionality to $d=6144$.
Then, the tokens are reshaped to form a feature map of size $H/4 \times W/4 \times D/8$.
Finally, 4 ConvNeXt blocks~\cite{liu2022convnext} are applied to this feature map before it is upsampled to full resolution ($H \times W$) using bilinear interpolation.

\subsection{Implementation details}

All experiments were implemented using PyTorch~\cite{paszke2019pytorch} (\rurl{pytorch.org}) and the \texttt{timm} library (\rurl{huggingface.co/docs/timm}).
For the evaluation, the \texttt{scikit-learn} (\rurl{scikit-learn.org}) library was used.
Following previous work~\cite{zhou2023retfound}, we used the ViT~\cite{dosovitskiy2020image} architecture pretrained on ImageNet using MAE~\cite{he2022mae} as the backbone of our model.
The main experiments were conducted using the ViT-Large architecture, while a version of MIRAGE based on the ViT-Base architecture was trained to study the impact of domain-specific multimodal pretraining and model capacity.

The pseudo-labels of retinal layers were generated using the Iowa Reference Algorithms 3.6 (\rurl{iibi.uiowa.edu/oct-reference}) (Retinal Image Analysis Lab, Iowa Institute for Biomedical Imaging, Iowa City, IA)~{\cite{garvin2008intraretinal,antony2011automated}}.

During model pretraining, the AdamW~\cite{loshchilov2018adamw} optimizer with a base learning rate of $10^{-4}$ and a weight decay of 0.05 was used.
The learning rate follows the linear scaling rule~\cite{goyal2017accurate}: $lr = base\_lr \times batchsize/256$.
Model weights were initialized from a ViT model pretrained with MAE~\cite{he2022mae} on ImageNet~\cite{deng2009imagenet}.
The pretraining lasted 1600 epochs, with a 40-epoch warm-up starting with a learning rate of $10^{-6}$, which was then decayed to 0 using cosine decay.
Training was performed with a batch size of 256 on a single A100 GPU (80 GB) with automatic mixed precision enabled.
The input images were resized offline to $512\times512$, and data augmentation was used to artificially increase the size of the dataset.
In particular, we applied random affine transformations, slight intensity shifts, and horizontal flipping.
The number of selected patches was fixed to 49 in the case of a single modality, and 98 for two or three modalities, with a patch size of $32\times32$ pixels.
The proportion of tokens per modality $\lambda^m$ was determined by sampling from a symmetric Dirichlet distribution with $\alpha = 1$, ensuring that the sum of $\lambda^m$ across all modalities equals 1.
In Supplementary Note~\ref{sup:sec:pre_results},
we provide our pre-experimental results confirming the effective functioning of the pretraining approach with the chosen hyperparameters.

For the classification tasks, modality-specific decoders were replaced with a single linear layer followed by a softmax activation function.
In the experiments, these are the only parameters of the model that are trained, while the encoder remains frozen.
We train the models for a maximum of 1000 epochs using the AdamW optimizer with a learning rate of $10^{-3}$ and a weight decay of $10^{-2}$.
Early stopping was implemented from epoch 20, with patience of 20 epochs, saving a model checkpoint if the balanced accuracy (BAcc) on the validation set exceeded the previous best or matched it with a lower loss.
The minimum improvement required was 0.1 percentage points.
The batch size is configured to be 25\% of the training set size, with a maximum of 64.
Similar to RETFound~\cite{zhou2023retfound}, we performed label smoothing with a smoothing factor of 0.1.
For each dataset, five random seeds were used to ensure the robustness of the results.
Random augmentation was applied to the input images during training, including random horizontal flipping, affine transformations, and slight intensity shifts.
For MIRAGE, min-max normalization was applied to the input images, while for the other models, they were standardized using ImageNet statistics.

For the segmentation tasks, the modality-specific decoders were replaced by different modules depending on the experiment.
For the state-of-the-art comparison, to maximize performance, we used a convolutional segmentation head based on ConvNeXt~{\cite{liu2022convnext}}, as proposed by Bachmann et al.~{\cite{bachmann2022multimae}}.
However, preliminary experiments showed that other segmentation heads such as DPT~\cite{ranftl2021vision} can be used with similar performance (Supplementary Table~\ref{sup:tab:ablation_decoder}).
In contrast, to analyze the impact of domain-specific multimodal pretraining, we used a linear probing strategy.
In particular, following DINOv2~{\cite{oquab2024dinov2}}, we used a single $1\times1$ convolutional layer followed by bicubic upsampling.
This tuning setting, while less effective than the ConvNeXt head, allows us to minimize the impact of the added modules and focus on the effect of the pretraining approach.
The best models were selected based on the validation set performance
and evaluated on the test set.
The models were then tuned for 200 epochs using the AdamW optimizer with a $10^{-4}$ learning rate and batch size of 4.
In the default tuning setup, the ViT encoder was frozen, and only the linear projection layers and the segmentation decoder were trained.
In the full fine-tuning setup, the encoder was also trained.
Random horizontal flipping and random cropping were applied to the input images during training.
The size of the cropped images was set to $1024\times1024$ pixels.
Similar to the classification tasks, images were normalized using min-max normalization for MIRAGE and ImageNet statistics for the other models.

For the specialist models, nnUNet~\cite{isensee2021nnunet,isensee2024nnunet2} was used as is, but fixing the training, validation, and test splits to ensure consistency with the other models.
On the other hand, SwinUNETR-V2~\cite{he2023swinunetrv2}, MedNeXt~\cite{roy2023mednext} and TransUNet~\cite{chen2024transunet} were trained using a similar setup to MIRAGE, with the same data augmentation, optimizer, hyperparameters, and model selection criteria.

In Supplementary Note~\ref{sup:sec:efficiency},
we provide a detailed analysis of the computational efficiency of MIRAGE and its different architectural variants (pretraining, classification, and segmentation) and a comparison with other FMs in terms of pretraining efficiency.
The results of this analysis are shown in Supplementary Tables~\ref{sup:tab:efficiency} and \ref{sup:tab:resources}.

\subsection{Classification evaluation procedure}

Model performance on the classification tasks was evaluated using the area under the receiver operating characteristic curve (AUROC), average precision (AP), and balanced accuracy (BAcc) metrics.
AUROC is calculated using the one-vs-all strategy, where the AUC of each class is computed against the rest~\cite{provost2000well,fawcett2006introduction}.
The overall AUROC is then calculated by averaging these AUCs using a weighted average, where the weight is the number of positive samples in each class.
The AP summarizes a precision-recall curve as the weighted mean of precisions achieved at each threshold, with the increase in recall from the previous threshold used as the weight.
Similar to the AUROC, the total AP is calculated by averaging the APs of each class, weighted by the number of positive samples for each label.
Finally, the BAcc is defined as the average of recall obtained on each class, as described by Mosley~\cite{mosley2013balanced}.
This combination of metrics allows us to obtain a comprehensive understanding of model performance, especially in the context of class imbalance.

For each task, the models were trained with five different random seeds to ensure robustness and reliability of the results.
This approach helps to capture the inherent variability in the training process, as different random seeds influence the initialization of the shuffling and augmentation of the training data.
The mean and standard deviation of the performance over the five replicas from the five seeds are then reported.

To assess the statistical significance of the performance difference between the best and second-best models in each task, we used the one-tailed Student's $t$ test~\cite{zhou2023retfound}.
On the other hand, to compare the performance of the models across different tasks, we used the Wilcoxon signed-rank test~\cite{demvsar2006statistical}.

\subsection{Segmentation evaluation procedure}

To evaluate the accuracy of the predicted segmentations, we followed standard practices in the literature~\cite{heimann2009comparison,chiu2015kernel,bogunovic2019retouch,fang2022dataset,ma2024segment} and calculated the volume Dice coefficient, volume intersection over union (IoU), and 95\textsuperscript{th} percentile of the Hausdorff distance (HD95).
Dice coefficient is a measure of the overlap between the predicted and manual segmentations, with a value of 1 indicating perfect overlap and 0 indicating no overlap.
It is calculated as the intersection of the predicted and manual segmentations divided by the sum of their volumes.
IoU is a similar metric, and it is calculated as the intersection of the segmentations divided by their union.
Finally, HD is a measure of the maximum distance between the predicted and manual segmentations.
The 95\textsuperscript{th} percentile of the distances, HD95, is used in order to reduce the impact of outliers.
This metric is measured in pixels.

Since the evaluation on RETOUCH is performed using the official evaluation server, only the average absolute volume difference (AVD) is available apart from the Dice coefficient.
Therefore, for RETOUCH, we report only these two metrics.
The AVD is calculated as the absolute difference between the predicted and ground truth lesion volumes, in cubic millimeters (mm\textsuperscript{3}), averaged across all lesions.

In the segmentation tasks, the models were trained to segment all classes simultaneously, including lesions and layers in datasets where both types of segmentation were available.
All models were trained with a single seed, as the training process was stable and little variability was observed between different runs.
In each dataset, we computed the metrics for every class for each B-scan; then, the B-scans were grouped by patient, and the averages for each lesion were computed, which were ultimately averaged to get the final performance for that patient.
The mean and standard deviation of the performance of the model were calculated at the patient level.
Similar to the classification tasks, the statistical significance of the performance difference between the best and second best models in each task was assessed using the one-tailed Student's $t$ test, while the Wilcoxon signed-rank test was used to compare the performance of the models across different tasks.

\section{Data availability}

The pretraining data used in this study come from an in-house dataset, while the data for the evaluation come from 14 publicly available datasets and two in-house datasets.
Publicly available data can be obtained from the original sources, with the exception of GOALS, which is no longer available but has been uploaded to our GitHub repository.
The links to the datasets are as follows:
Duke iAMD~\cite{farsiu2014quantitative} (\rurl{people.duke.edu/~sf59/RPEDC_Ophth_2013_dataset.htm});
Duke Srinivasan~\cite{srinivasan2014fully} (\rurl{people.duke.edu/~sf59/Srinivasan_BOE_2014_dataset.htm});
GAMMA~\cite{wu2023gamma} (\rurl{gamma.grand-challenge.org/});
Harvard Glaucoma~\cite{luo2023harvard} (\rurl{github.com/Harvard-Ophthalmology-AI-Lab/Harvard-GDP});
Kermany~\cite{kermany2018dataset,kermany2018identifying} (\rurl{data.mendeley.com/datasets/rscbjbr9sj/2});
Noor Eye Hospital~\cite{rasti2017macular} (\rurl{hrabbani.site123.me/available-datasets/dataset-for-oct-classification-50-normal-48-amd-50-dme});
OCTDL~\cite{kulyabin2024octdl} (\rurl{data.mendeley.com/datasets/sncdhf53xc/4});
OCTID~\cite{gholami2020octid} (\rurl{borealisdata.ca/dataverse/OCTID});
OLIVES~\cite{prabhushankar2022olives} (\rurl{zenodo.org/records/7105232});
UMN~\cite{rashno2018fully} (\rurl{people.ece.umn.edu/users/parhi/.DATA/});
AROI~\cite{melinscak2021aroi} (\rurl{ipg.fer.hr/ipg/resources/oct_image_database});
Duke DME~\cite{chiu2015kernel} (\rurl{people.duke.edu/~sf59/Chiu_BOE_2014_dataset.htm});
GOALS~\cite{fang2022dataset} (\rurl{github.com/j-morano/MIRAGE});
RETOUCH~\cite{bogunovic2019retouch} (\rurl{retouch.grand-challenge.org/}).
Notwithstanding, to facilitate reproducibility, we also provide these datasets with the same splits used in this study, except for Kermany~\cite{kermany2018dataset}, in our GitHub repository: \url{https://github.com/j-morano/MIRAGE}.
The Kermany dataset with the original split, which is the same we used in this work, can be accessed at the original URL provided by the authors and listed above.
Due to privacy concerns, the in-house VIBES~\cite{gerendas2022validation}, OPTIMA9C~\cite{oghbaie2024vlfatrollout}, and SGA~\cite{bui2022fundus} datasets cannot be made publicly available.

\section{Code availability}

The underlying code for this study, the weights of the pretrained models, and the full detailed results are available on GitHub and can be accessed via the following link: \url{https://github.com/j-morano/MIRAGE}.

\section{Acknowledgments}

This work was supported in part by Austrian Federal Ministry of Economy, Energy and Tourism, the National Foundation for Research, Technology and Development, the Christian Doppler Research Association, Heidelberg Engineering, and Austrian Science Fund (FWF) grant 10.55776/FG9.
For open access purposes, the author has applied a CC BY public copyright license to any author-accepted manuscript version arising from this submission.

\section{Author contributions}

J.M. conceived the methodology, designed and executed all experiments, conducted all subsequent statistical analyses, and drafted the manuscript.
B.F. helped design the experiments and contributed to the validation of the proposed method and the writing.
E.S. contributed to the preparation of the benchmark data, the implementation of the downstream experiments, and the writing.
R.F. contributed to the implementation of the segmentation baselines.
T.E. helped with the design of the experiments.
M.G. and G.F. contributed to the data preparation and analysis.
M.O. contributed to the analysis of the results and writing the supplementary material.
U.S. contributed to the collection and preparation of the pretraining dataset.
H.B. contributed to the design of the experiments, analysis of the results, and made substantial revisions and edits of the draft manuscript.
All authors contributed to the manuscript.

\section{Competing interests}

The authors declare no competing interests.




\clearpage

\appendix

\begin{center}
\textbf{\huge Supplementary Information}
\vspace{1cm}
\end{center}

\setcounter{figure}{0}
\setcounter{table}{0}


\renewcommand{\figurename}{Supplementary Figure}

\renewcommand{\tablename}{Supplementary Table}

\renewcommand{\thesection}{\arabic{section}}

\section{Supplementary Note 1: Extended results}
\label{sup:sec:results}

In the following, we provide the extended tables of the experimental results, showing not only the average performance across all datasets, but also the results for each individual dataset.
For the classification tasks, we report the area under the receiver operating characteristic curve (AUROC), average precision (AP), and balanced accuracy (BAcc).
For the segmentation tasks, we report the Dice score, intersection over union (IoU), and 95\textsuperscript{th} percentile of the Hausdorff distance (HD95) for all datasets except RETOUCH.
For the RETOUCH dataset, we used the official evaluation server of the challenge to compute the metrics, which include the Dice score and absolute volume difference (AVD).
To facilitate the comparison of the results, we highlight the best results in bold and underline the second best results.
In addition, we have color-coded the best results according to the model.
The color code is as follows (excluding models with no best results):
\textbfb{MIRAGE}[\textbfb{-Large}] (blue),
\textbfr{RETFound}~\cite{zhou2023retfound} (red),
\textbfg{DINOv2}~\cite{oquab2024dinov2} (green),
\textbf{SL-IN}~\cite{dosovitskiy2020image} (black),
\textbfm{nnUNet}~\cite{isensee2021nnunet} (magenta),
\textbfbg{MIRAGE--FFT} (cyan),
\textbft{MultiMAE}~\cite{bachmann2022multimae} (teal),
\textbfs{MAE-OCT} and \textbfs{MAE-SLO} (salmon),
\textbfdbg{MIRAGE-Base} (blue-gray),
\textbfli{OCT+Layers} (lime),
and \textbftu{TransUNet}~(dark gray).
In all cases, the one-tailed Student's $t$-test was used to assess the statistical significance of the performance difference between the best (in bold) and second best (underlined) models in each dataset.
The Wilcoxon signed-rank test was used to assess the statistical significance of the performance differences across all datasets.
Statistical significance is indicated by asterisks in the tables: *$p<0.05$, **$p<0.01$, ***$p<0.001$.
In those cases where in-house datasets were used for the evaluation, we provide the average performance across both all the datasets and only public datasets (excluding in-house datasets).


\begin{table}[tbph]
    \centering
    \caption{%
        \textbf{Performance of MIRAGE and state-of-the-art foundation models SL-IN~\cite{dosovitskiy2020image}, DINOv2~\cite{oquab2024dinov2}, and RETFound~\cite{zhou2023retfound} on the different datasets for the diagnosis and staging of ocular diseases on OCT.}
        All models are based on the ViT-Large architecture and were tuned using linear probing.
        Statistical significance between the best (bold) and second best (underlined) models in each and across all datasets was assessed using the one-tailed Student's $t$-test and Wilcoxon signed-rank test, respectively (*$p<0.05$, **$p<0.01$, ***$p<0.001$).
    }%
    \label{sup:tab:sota_class_oct}
    \resizebox{0.95\textwidth}{!}{%

        \begin{tabular}{lllll}
            \toprule

            \textbf{Dataset}
            & \textbf{Model}
            & \textbf{AUROC}
            & \textbf{AP}
            & \textbf{BAcc}
            \\

\midrule

\textbf{Duke iAMD}
& SL-IN
& 97.63 $\pm$ 0.23
& 97.56 $\pm$ 0.35
& 94.49 $\pm$ 0.45
\\

& DINOv2
& \underline{99.13 $\pm$ 0.34}
& \underline{99.23 $\pm$ 0.30}
& 95.53 $\pm$ 0.80
\\

& RETFound
& 98.24 $\pm$ 0.66
& 98.69 $\pm$ 0.50
& \underline{96.03 $\pm$ 1.72}
\\

& MIRAGE
& \textbfb{99.52 $\pm$ 0.23}**
& \textbfb{99.52 $\pm$ 0.23}**
& \textbfb{96.71 $\pm$ 1.42}
\\

\midrule

\textbf{GAMMA}
& SL-IN
& 74.83 $\pm$ 1.67
& 64.02 $\pm$ 1.89
& 54.72 $\pm$ 2.72
\\

& DINOv2
& 71.99 $\pm$ 2.62
& 60.87 $\pm$ 3.75
& 45.83 $\pm$ 4.48
\\

& RETFound
& \underline{82.30 $\pm$ 2.27}
& \underline{73.27 $\pm$ 3.65}
& \underline{59.17 $\pm$ 2.72}
\\

& MIRAGE
& \textbfb{87.50 $\pm$ 0.64}**
& \textbfb{81.00 $\pm$ 1.89}*
& \textbfb{63.61 $\pm$ 4.16}*
\\

\midrule

\textbf{Harvard Glaucoma}
& SL-IN
& 75.11 $\pm$ 0.90
& 74.19 $\pm$ 0.92
& 69.32 $\pm$ 1.02
\\

& DINOv2
& 72.29 $\pm$ 2.31
& 72.60 $\pm$ 2.48
& 64.80 $\pm$ 1.94
\\

& RETFound
& \underline{82.14 $\pm$ 0.68}
& \underline{81.53 $\pm$ 0.71}
& \underline{74.14 $\pm$ 1.28}
\\

& MIRAGE
& \textbfb{82.75 $\pm$ 0.65}
& \textbfb{82.54 $\pm$ 0.68}
& \textbfb{76.15 $\pm$ 1.25}
\\

\midrule

\textbf{Kermany}
& SL-IN
& 98.65 $\pm$ 0.04
& 95.71 $\pm$ 0.12
& 86.15 $\pm$ 0.68
\\

& DINOv2
& 98.67 $\pm$ 0.06
& 95.87 $\pm$ 0.11
& \underline{86.48 $\pm$ 0.39}
\\

& RETFound
& \underline{98.92 $\pm$ 0.07}
& \underline{96.50 $\pm$ 0.14}
& 85.89 $\pm$ 0.51
\\

& MIRAGE
& \textbfb{99.53 $\pm$ 0.01}***
& \textbfb{98.39 $\pm$ 0.04}***
& \textbfb{91.40 $\pm$ 0.25}***
\\

\midrule

\textbf{Noor Eye Hospital}
& SL-IN
& 97.03 $\pm$ 0.78
& 95.06 $\pm$ 1.21
& 84.67 $\pm$ 2.67
\\

& DINOv2
& \underline{97.53 $\pm$ 0.67}
& \underline{96.19 $\pm$ 0.86}
& \underline{90.00 $\pm$ 2.11}
\\

& RETFound
& 95.77 $\pm$ 0.87
& 94.24 $\pm$ 0.96
& 86.00 $\pm$ 2.49
\\

& MIRAGE
& \textbfb{98.63 $\pm$ 0.40}**
& \textbfb{97.69 $\pm$ 0.68}**
& \textbfb{92.67 $\pm$ 3.27}
\\

\midrule

\textbf{OCTDL}
& SL-IN
& 96.86 $\pm$ 0.16
& 91.24 $\pm$ 0.21
& \textbf{79.69 $\pm$ 1.39}
\\

& DINOv2
& 95.22 $\pm$ 0.42
& 88.26 $\pm$ 1.06
& \underline{77.59 $\pm$ 1.95}
\\

& RETFound
& \underline{96.96 $\pm$ 0.28}
& \underline{91.72 $\pm$ 0.66}
& 74.30 $\pm$ 1.39
\\

& MIRAGE
& \textbfb{98.27 $\pm$ 0.15}***
& \textbfb{93.32 $\pm$ 0.27}**
& 76.93 $\pm$ 2.94
\\

\midrule

\textbf{OCTID}
& SL-IN
& 98.72 $\pm$ 0.11
& 95.71 $\pm$ 0.44
& \textbf{89.54 $\pm$ 0.81}
\\

& DINOv2
& 98.81 $\pm$ 0.20
& 95.96 $\pm$ 0.83
& 85.04 $\pm$ 2.49
\\

& RETFound
& \underline{98.98 $\pm$ 0.28}
& \textbfr{96.81 $\pm$ 0.65}
& \underline{88.16 $\pm$ 2.04}
\\

& MIRAGE
& \textbfb{99.07 $\pm$ 0.28}
& \underline{96.54 $\pm$ 0.84}
& 87.53 $\pm$ 1.58
\\

\midrule

\textbf{OLIVES}
& SL-IN
& \underline{97.89 $\pm$ 0.09}
& \underline{97.48 $\pm$ 0.11}
& \underline{95.79 $\pm$ 0.54}
\\

& DINOv2
& 94.13 $\pm$ 0.23
& 93.46 $\pm$ 0.22
& 88.64 $\pm$ 0.39
\\

& RETFound
& \textbfr{98.05 $\pm$ 0.15}*
& \textbfr{97.68 $\pm$ 0.25}
& \textbfr{95.81 $\pm$ 0.27}
\\

& MIRAGE
& 96.06 $\pm$ 0.09
& 94.36 $\pm$ 0.24
& 93.14 $\pm$ 0.27
\\

\midrule

\textbf{OPTIMA9C}
& SL-IN
& 97.33 $\pm$ 0.07
& 86.65 $\pm$ 0.32
& 69.63 $\pm$ 0.83
\\

{\small (in-house)}
& DINOv2
& 97.44 $\pm$ 0.08
& 86.81 $\pm$ 0.14
& 67.27 $\pm$ 1.22
\\

& RETFound
& \underline{98.62 $\pm$ 0.08}
& \underline{91.76 $\pm$ 0.32}
& \underline{74.02 $\pm$ 0.64}
\\

& MIRAGE
& \textbfb{99.01 $\pm$ 0.02}***
& \textbfb{93.52 $\pm$ 0.16}***
& \textbfb{77.95 $\pm$ 0.45}***
\\

\midrule

\textbf{\textit{Average}}
& SL-IN
& 92.67 $\pm$ 9.51
& 88.63 $\pm$ 11.21
& 80.44 $\pm$ 12.86
\\

& DINOv2
& 91.69 $\pm$ 10.64
& 87.69 $\pm$ 12.23
& 77.91 $\pm$ 15.10
\\

& RETFound
& \underline{94.44 $\pm$ 6.66}
& \underline{91.35 $\pm$ 8.15}
& \underline{81.50 $\pm$ 11.48}
\\

& MIRAGE
& \textbfb{95.59 $\pm$ 5.80}***
& \textbfb{92.99 $\pm$ 6.39}***
& \textbfb{84.01 $\pm$ 10.51}***
\\

\midrule

\textbf{\textit{Average}}
& SL-IN
& 92.09 $\pm$ 9.93
& 88.87 $\pm$ 11.87
& 81.80 $\pm$ 13.02
\\

{\small (public only, excluding}
& DINOv2
& 90.97 $\pm$ 11.07
& 87.80 $\pm$ 12.96
& 79.24 $\pm$ 15.50
\\

{\small in-house)}
& RETFound
& \underline{93.92 $\pm$ 6.89}
& \underline{91.30 $\pm$ 8.65}
& \underline{82.44 $\pm$ 11.84}
\\

& MIRAGE
& \textbfb{95.17 $\pm$ 6.02}***
& \textbfb{92.92 $\pm$ 6.77}***
& \textbfb{84.77 $\pm$ 10.91}***
\\

            \bottomrule
        \end{tabular}

    }
\end{table}

\begin{table}[tbhp]
    \centering
    \caption{
        \textbf{Performance of MIRAGE and state-of-the-art foundation models SL-IN~\cite{dosovitskiy2020image}, DINOv2~\cite{oquab2024dinov2}, and RETFound~\cite{zhou2023retfound} on the different datasets for the diagnosis of ocular diseases on SLO.}
        All models are based on the ViT-Large architecture and were tuned using linear probing.
        Statistical significance between the best (bold) and second best (underlined) models in each and across all datasets was assessed using the one-tailed Student's $t$-test and Wilcoxon signed-rank test, respectively (*$p<0.05$, **$p<0.01$, ***$p<0.001$).
    }%
    \label{sup:tab:sota_class_slo}
    \resizebox{\textwidth}{!}{%
    
        \begin{tabular}{lllll}
            \toprule

            \textbf{Dataset}
            & \textbf{Model}
            & \textbf{AUROC}
            & \textbf{AP}
            & \textbf{BAcc}
            \\

\midrule

\textbf{OPTIMA9C}
& SL-IN
& \underline{91.96 $\pm$ 0.15}
& \underline{75.99 $\pm$ 0.45}
& \underline{52.96 $\pm$ 1.42}
\\

& DINOv2
& 91.89 $\pm$ 0.18
& 73.22 $\pm$ 0.44
& 52.67 $\pm$ 1.45
\\

& MIRAGE
& \textbfb{93.40 $\pm$ 0.14}***
& \textbfb{77.38 $\pm$ 0.47}*
& \textbfb{55.55 $\pm$ 0.45}**
\\

\midrule

\textbf{OLIVES}
& SL-IN
& \underline{73.22 $\pm$ 2.34}
& \textbf{73.71 $\pm$ 2.76}
& \underline{64.61 $\pm$ 2.70}
\\

& DINOv2
& 73.14 $\pm$ 3.38
& 71.55 $\pm$ 3.61
& 58.61 $\pm$ 2.12
\\

& MIRAGE
& \textbfb{74.08 $\pm$ 0.52}
& \underline{73.40 $\pm$ 0.52}
& \textbfb{66.72 $\pm$ 1.14}
\\

\midrule

\textbf{\textit{Average}}
& SL-IN
& \underline{82.59 $\pm$ 9.52}
& \underline{74.85 $\pm$ 2.28}
& \underline{58.79 $\pm$ 6.21}
\\

& DINOv2
& 82.52 $\pm$ 9.68
& 72.39 $\pm$ 2.70
& 55.64 $\pm$ 3.48
\\

& MIRAGE
& \textbfb{83.74 $\pm$ 9.67}*
& \textbfb{75.39 $\pm$ 2.05}
& \textbfb{61.13 $\pm$ 5.65}**
\\

            \bottomrule
        \end{tabular}

    }
\end{table}


\begin{table}[tbhp]
    \centering
    \caption{\textbf{Performance of MIRAGE and RETFound~\cite{zhou2023retfound} models on the first and last scans of the OLIVES dataset (DME/DR discrimination) under linear probing.}
        OLIVES data comes from two clinical studies where patients are treated throughout the process.
        As shown in the original paper~\cite{prabhushankar2022olives}, treatment produces meaningful changes in the retina, and causes a domain shift in the data for the same patient, which is reflected in the performance of the models on the last scans.
        The lower robustness of MIRAGE to this domain shift can be explained by the much lower number of samples from diabetic patients used in the pretraining of the model.
        In particular, less than 2\% of the samples from our VIBES dataset are from diabetic patients, while 85\% of the samples from the dataset used to train RETFound are from the Moorfields diabetic image dataset (MEH-MIDAS)~\cite{zhou2023retfound}.
    }%
    \label{sup:tab:olives_first_last}
    
        \begin{tabular}{lllllll}
            \toprule

            \textbf{Dataset}
            & \textbf{Model}
            & \textbf{AUROC}
            & \textbf{AP}
            & \textbf{BAcc}
            \\

\midrule

\textbf{OLIVES (first scans)}
& RETFound
& 98.80 $\pm$ 0.75
& 98.92 $\pm$ 0.67
& \textbfr{95.00 $\pm$ 0.00}
\\

& MIRAGE
& \textbfb{100.00 $\pm$ 0.00}
& \textbfb{100.00 $\pm$ 0.00}
& \textbfb{95.00 $\pm$ 0.00}
\\

\midrule

\textbf{OLIVES (last scans)}
& RETFound
& \textbfr{94.00 $\pm$ 3.63}
& \textbfr{94.71 $\pm$ 3.25}
& \textbfr{84.00 $\pm$ 5.83}
\\

& MIRAGE
& 80.20 $\pm$ 3.54
& 82.76 $\pm$ 3.43
& 70.00 $\pm$ 4.47
\\

            \bottomrule
        \end{tabular}

\end{table}

\begin{table}[tbhp]
    \centering
    \caption{%
        \textbf{Performance of MIRAGE and state-of-the-art foundation models SL-IN~\cite{dosovitskiy2020image}, DINOv2~\cite{oquab2024dinov2}, and RETFound~\cite{zhou2023retfound} on the cross-dataset evaluation setting for OCT diagnosis.}
        All models are based on the ViT-Large architecture and were tuned using linear probing.
        Statistical significance between the best (bold) and second best (underlined) models in each and across all datasets was assessed using the one-tailed Student's $t$-test and Wilcoxon signed-rank test, respectively (*$p<0.05$, **$p<0.01$, ***$p<0.001$).
    }%
    \label{sup:tab:sota_class_external}
    \resizebox{\textwidth}{!}{%
    
        \begin{tabular}{lllll}
            \toprule

            \textbf{Dataset}
            & \textbf{Model}
            & \textbf{AUROC}
            & \textbf{AP}
            & \textbf{BAcc}
            \\

\midrule

\textbf{Noor Eye Hospital}
& SL-IN
& \underline{90.05 $\pm$ 0.99}
& \underline{82.65 $\pm$ 1.82}
& \underline{66.72 $\pm$ 5.01}
\\

{\small trained on UMN + Duke}
& DINOv2
& 83.59 $\pm$ 1.23
& 73.49 $\pm$ 1.96
& \textbfg{66.77 $\pm$ 3.80}
\\

{\small Srinivasan}
& RETFound
& 81.63 $\pm$ 2.97
& 66.57 $\pm$ 4.34
& 47.39 $\pm$ 1.40
\\

& MIRAGE
& \textbfb{94.79 $\pm$ 0.74}**
& \textbfb{92.27 $\pm$ 0.74}***
& 64.98 $\pm$ 1.70
\\

\midrule

\textbf{UMN + Duke Srinivasan}
& SL-IN
& \underline{86.23 $\pm$ 0.72}
& \underline{80.74 $\pm$ 0.77}
& 65.01 $\pm$ 1.83
\\

{\small trained on Noor Eye Hospital}
& DINOv2
& 78.76 $\pm$ 1.01
& 72.73 $\pm$ 1.06
& 68.60 $\pm$ 1.82
\\

& RETFound
& 79.63 $\pm$ 1.06
& 72.81 $\pm$ 1.53
& \underline{68.74 $\pm$ 2.87}
\\

& MIRAGE
& \textbfb{88.59 $\pm$ 3.48}
& \textbfb{85.69 $\pm$ 4.78}
& \textbfb{79.81 $\pm$ 2.37}**
\\

            \bottomrule
        \end{tabular}

    }
\end{table}


\begin{table}[tbhp]
    \centering
    \caption{%
        \textbf{Performance of MIRAGE with different decoders on different downstream segmentation tasks:
        Segmenter~\cite{strudel2021segmenter}, DPT~\cite{ranftl2021vision}, and ConvNeXt~\cite{liu2022convnext,bachmann2022multimae}.}
        In all cases, the encoder is based on the ViT-Base architecture and is kept frozen during tuning.
        Statistical significance between the best (bold) and second best (underlined) models across all SLO and OCT datasets was assessed using the Wilcoxon signed-rank test.
        However, no statistical significance was found ($p>0.05$).
        MIRAGE performs particularly well with ConvNeXt and DPT decoders, but still offers adequate performance with the more lightweight Segmenter decoder.
    }%
    \label{sup:tab:ablation_decoder}
    \resizebox{\textwidth}{!}{%

\begin{tabular}{llllll}
    \toprule

    \textbf{Modality}
    & \textbf{Dataset}
    & \textbf{Model}
    & \textbf{Dice}
    & \textbf{IoU}
    & \textbf{HD95} $\downarrow$
    \\

    \midrule

\multirow{3}{*}{\textbf{SLO}}
& \multirow{3}{*}{\textbf{SGA (lesions)}}
    & Segmenter
    & 71.97 $\pm$ 22.70
    & 61.54 $\pm$ 23.88
    & 169.34 $\pm$ 78.47
    \\

    &
    & DPT
    & \textbf{77.07 $\pm$ 18.89}
    & \textbf{67.64 $\pm$ 21.61}
    & \textbf{153.31 $\pm$ 57.12}
    \\

    & & ConvNeXt
    & \underline{74.47 $\pm$ 22.11}
    & \underline{64.79 $\pm$ 24.23}
    & \underline{161.13 $\pm$ 88.94}
    \\

\midrule

\multirow{9}{*}{\textbf{OCT}}
& \multirow{3}{*}{\textbf{Duke DME (layers)}}
    & Segmenter
    & 72.39 $\pm$ 4.52
    & 57.53 $\pm$ 5.50
    & 19.45 $\pm$ 7.29
    \\

    & & DPT
    & \underline{82.68 $\pm$ 3.40}
    & \underline{70.96 $\pm$ 4.83}
    & \textbf{8.23 $\pm$ 4.23}
    \\

    &
    & ConvNeXt
    & \textbf{82.86 $\pm$ 3.60}
    & \textbf{71.23 $\pm$ 5.12}
    & \underline{8.57 $\pm$ 4.01}
    \\

\cmidrule(lr){2-6}

& \multirow{3}{*}{\textbf{Duke DME (lesions)}}
    & Segmenter
    & 62.38 $\pm$ 9.26
    & 45.87 $\pm$ 10.00
    & 50.32 $\pm$ 18.13
    \\

    &
    & DPT
    & \underline{63.92 $\pm$ 9.48}
    & \underline{47.54 $\pm$ 10.15}
    & \textbf{40.46 $\pm$ 11.58}
    \\

    & & ConvNeXt
    & \textbf{69.06 $\pm$ 9.15}
    & \textbf{53.36 $\pm$ 11.00}
    & \underline{49.32 $\pm$ 30.38}
    \\

\cmidrule(lr){2-6}

& \multirow{3}{*}{\textbf{GOALS (layers)}}
    & Segmenter
    & 82.87
    & 71.45
    & 14.20
    \\

    & & DPT
    & \underline{92.01}
    & \underline{85.33}
    & \underline{4.89}
    \\

    &
    & ConvNeXt
    & \textbf{92.87}
    & \textbf{86.79}
    & \textbf{4.87}
    \\

\midrule

\multirow{3}{*}{\textbf{OCT \& SLO}}
& \multirow{3}{*}{\textbf{\textit{Average}}}
   & Segmenter
    & 72.40 $\pm$ 7.25
    & 59.10 $\pm$ 9.17
    & 63.33 $\pm$ 62.75
    \\

   &
   & DPT
   & \underline{78.92 $\pm$ 10.17}
   & \underline{67.87 $\pm$ 13.49}
   & \textbf{51.72 $\pm$ 60.27}
   \\

   & & ConvNeXt
    & \textbf{79.82 $\pm$ 9.00}
    & \textbf{69.04 $\pm$ 12.08}
    & \underline{55.97 $\pm$ 63.17}
    \\

    \bottomrule
\end{tabular}

    }
\end{table}


\begin{table}[tbhp]
    \centering
    \caption{\textbf{Performance of MIRAGE and state-of-the-art foundation models DINOv2~\cite{oquab2024dinov2}, RETFound~\cite{zhou2023retfound}, and MedSAM~\cite{ma2024segment} on the different datasets for the segmentation of retinal lesions and layers.}
    Mean and standard deviation values are calculated across the patients in each dataset, except for the GOALS and RETOUCH datasets, where patient information is not available.
    All models are based on the ViT-Large architecture, except for MedSAM, which is only available in the ViT-Base version.
    The models were fine-tuned using both the decoder-only fine-tuning strategy.
    Statistical significance between the best (bold) and second best (underlined) models in each and across all datasets was assessed using the one-tailed Student's $t$-test and Wilcoxon signed-rank test, respectively (*$p<0.05$, **$p<0.01$, ***$p<0.001$).
    \textdagger~RETOUCH results were obtained using the official evaluation server of the challenge, which only includes the Dice score and absolute volume difference (AVD), which is reported in the HD95 column.
    Only the Dice score is included for the calculation of the average performance across all datasets.
    }%
    \label{sup:tab:sota_seg}
    \resizebox{\textwidth}{!}{%

\begin{tabular}{llllll}
    \toprule

    \textbf{Modality}
    & \textbf{Dataset}
    & \textbf{Model}
    & \textbf{Dice}
    & \textbf{IoU}
    & \textbf{HD95 / AVD\textsuperscript{\textdagger}} $\downarrow$
    \\

    \midrule

\multirow{28}{*}{\textbf{OCT}}
& \multirow{4}{*}{\textbf{AROI (layers)}}
    & DINOv2
    & 85.60 $\pm$ 2.08
    & 75.90 $\pm$ 2.99
    & 13.97 $\pm$ 9.07
    \\

    & & RETFound
    & 85.23 $\pm$ 1.51
    & 75.46 $\pm$ 2.04
    & 8.21 $\pm$ 2.30
    \\

    & & MedSAM
    & \underline{90.85 $\pm$ 2.13}
    & \underline{84.01 $\pm$ 2.95}
    & \underline{7.80 $\pm$ 4.99}
    \\

    & & MIRAGE
    & \textbfb{93.79 $\pm$ 0.85}***
    & \textbfb{88.67 $\pm$ 1.42}***
    & \textbfb{3.25 $\pm$ 1.43}**
    \\

\cmidrule(lr){2-6}

& \multirow{4}{*}{\textbf{AROI (lesions)}}
    & DINOv2
    & 28.82 $\pm$ 8.88
    & 19.64 $\pm$ 6.66
    & 69.69 $\pm$ 32.14
    \\

    & & RETFound
    & 28.09 $\pm$ 8.39
    & 19.42 $\pm$ 6.32
    & 67.91 $\pm$ 34.61
    \\

    & & MedSAM
    & \underline{38.26 $\pm$ 8.86}
    & \underline{27.96 $\pm$ 7.08}
    & \underline{46.65 $\pm$ 15.64}
    \\

    & & MIRAGE
    & \textbfb{52.18 $\pm$ 16.63}**
    & \textbfb{40.89 $\pm$ 13.70}**
    & \textbfb{37.56 $\pm$ 12.17}*
    \\

\cmidrule(lr){2-6}

& \multirow{4}{*}{\textbf{Duke DME (layers)}}
    & DINOv2
    & 76.44 $\pm$ 5.90
    & 63.03 $\pm$ 7.64
    & 21.38 $\pm$ 11.53
    \\

    & & RETFound
    & 75.32 $\pm$ 4.80
    & 61.47 $\pm$ 6.00
    & 19.60 $\pm$ 12.12
    \\

    & & MedSAM
    & \underline{79.06 $\pm$ 4.47}
    & \underline{66.14 $\pm$ 6.05}
    & \underline{12.01 $\pm$ 5.16}
    \\

    & & MIRAGE
    & \textbfb{83.02 $\pm$ 2.93}**
    & \textbfb{71.41 $\pm$ 4.18}**
    & \textbfb{9.81 $\pm$ 4.77}*
    \\

\cmidrule(lr){2-6}

& \multirow{4}{*}{\textbf{Duke DME (lesions)}}
    & DINOv2
    & 52.27 $\pm$ 12.63
    & 36.15 $\pm$ 11.34
    & 66.96 $\pm$ 43.42
    \\

    & & RETFound
    & 51.53 $\pm$ 9.53
    & 35.16 $\pm$ 8.91
    & 68.51 $\pm$ 25.25
    \\

    & & MedSAM
    & \underline{60.21 $\pm$ 9.74}
    & \underline{43.60 $\pm$ 9.58}
    & \underline{55.52 $\pm$ 27.47}
    \\

    & & MIRAGE
    & \textbfb{69.72 $\pm$ 9.16}*
    & \textbfb{54.14 $\pm$ 11.14}*
    & \textbfb{42.45 $\pm$ 22.88}
    \\

\cmidrule(lr){2-6}

& \multirow{4}{*}{\textbf{GOALS (layers)}}
    & DINOv2
    & 81.28
    & 68.68
    & 41.34
    \\

    & & RETFound
    & 62.79
    & 47.16
    & 121.79
    \\

    & & MedSAM
    & \underline{90.36}
    & \underline{82.57}
    & \underline{10.38}
    \\

    & & MIRAGE
    & \textbfb{92.46}
    & \textbfb{86.08}
    & \textbfb{4.96}
    \\

\cmidrule(lr){2-6}

& \multirow{4}{*}{\textbf{RETOUCH (lesions)}}
    & DINOv2
    & 53.60
    & -
    & 0.13\textsuperscript{\textdagger}
    \\

    & & RETFound
    & 57.32
    & -
    & 0.11\textsuperscript{\textdagger}
    \\

    & & MedSAM
    & \underline{69.40}
    & -
    & 0.06\textsuperscript{\textdagger}
    \\

    & & MIRAGE
    & \textbfb{79.60}
    & -
    & \textbfb{0.03}\textsuperscript{\textdagger}
    \\

\cmidrule(lr){2-6}

& \multirow{4}{*}{\textbf{\textit{Average}}}
    & DINOv2
    & 63.00 $\pm$ 19.99
    & 52.68 $\pm$ 21.29
    & 42.67 $\pm$ 22.80
    \\

    & & RETFound
    & 60.05 $\pm$ 18.15
    & 47.73 $\pm$ 19.58
    & 57.20 $\pm$ 40.57
    \\

    & & MedSAM
    & \underline{71.36 $\pm$ 18.37}
    & \underline{60.86 $\pm$ 21.98}
    & \underline{26.47 $\pm$ 20.34}
    \\

    & & MIRAGE
    & \textbfb{78.46 $\pm$ 14.26}*
    & \textbfb{68.24 $\pm$ 18.40}*
    & \textbfb{19.61 $\pm$ 16.87}*
    \\

\midrule

& 
    & DINOv2
    & 77.51 $\pm$ 8.36
    & 66.17 $\pm$ 10.26
    & 60.93 $\pm$ 59.86
    \\

    \multirow{1}{*}{\textbf{OCT}}
    & \multirow{1}{*}{\textbf{Duke iAMD (layers)}}
    & RETFound
    & \underline{78.84 $\pm$ 7.59}
    & \underline{67.56 $\pm$ 9.49}
    & 53.87 $\pm$ 53.22
    \\

    {\small cross-dataset}
    & {\small trained on AROI}
    & MedSAM
    & 75.77 $\pm$ 11.54
    & 66.55 $\pm$ 12.22
    & \underline{39.98 $\pm$ 31.38}
    \\

    & & MIRAGE
    & \textbfb{91.06 $\pm$ 2.43}***
    & \textbfb{84.62 $\pm$ 3.38}***
    & \textbfb{4.62 $\pm$ 4.52}***
    \\

\midrule

\multirow{3}{*}{\textbf{SLO}}
& \multirow{3}{*}{\textbf{SGA (lesions)}}
    & DINOv2
    & 61.34 $\pm$ 24.56
    & 49.13 $\pm$ 23.88
    & 210.02 $\pm$ 85.59
    \\

    &
    & MedSAM
    & \underline{72.20 $\pm$ 24.05}
    & \underline{62.22 $\pm$ 25.58}
    & \underline{182.22 $\pm$ 86.21}
    \\

    &
    & MIRAGE
    & \textbfb{75.31 $\pm$ 22.16}***
    & \textbfb{65.81 $\pm$ 24.14}***
    & \textbfb{164.42 $\pm$ 95.35}**
    \\

    \bottomrule
\end{tabular}

    }
\end{table}


\begin{table}[tbhp]
    \centering
    \caption{\textbf{Performance of MIRAGE and state-of-the-art \textit{specialist} segmentation models SwinUNETR-V2~\cite{he2023swinunetrv2}, MedNeXt~\cite{roy2023mednext}, TransUNet~\cite{chen2024transunet}, and nnUNet~\cite{isensee2021nnunet} on the different datasets for the segmentation of retinal lesions and layers.}
    Mean and standard deviation values are calculated across the patients in each dataset, except for the GOALS and RETOUCH datasets, where patient information is not available.
    MIRAGE was trained using both the decoder-only fine-tuning and full fine-tuning (FFT) strategies.
    Statistical significance between the best (bold) and second best (underlined) models in each and across all datasets was assessed using the one-tailed Student's $t$-test and Wilcoxon signed-rank test, respectively (*$p<0.05$, **$p<0.01$, ***$p<0.001$).
    }%
    \label{sup:tab:specialist_seg}
    \resizebox{0.9\textwidth}{!}{%

\begin{tabular}{llllll}
    \toprule

    \textbf{Modality}
    & \textbf{Dataset}
    & \textbf{Model}
    & \textbf{Dice}
    & \textbf{IoU}
    & \textbf{HD95 / AVD\textsuperscript{\textdagger}} $\downarrow$
    \\

    \midrule

\multirow{42}{*}{\textbf{OCT}}
& \multirow{6}{*}{\textbf{AROI (layers)}}
    & SwinUNETR-V2
    & 94.00 $\pm$ 1.42
    & 89.06 $\pm$ 2.34
    & 3.66 $\pm$ 1.82
    \\

    & & MedNeXt
    & 94.07 $\pm$ 1.45
    & 89.19 $\pm$ 2.29
    & 3.60 $\pm$ 1.68
    \\

    & & TransUNet
    & 94.63 $\pm$ 1.27
    & 90.10 $\pm$ 2.11
    & 2.74 $\pm$ 1.18
    \\

    & & nnUNet
    & \textbfm{95.05 $\pm$ 0.79}
    & \textbfm{90.81 $\pm$ 1.34}
    & \underline{2.33 $\pm$ 0.79}
    \\

    & & MIRAGE
    & 93.79 $\pm$ 0.85
    & 88.67 $\pm$ 1.42
    & 3.25 $\pm$ 1.43
    \\

    & & MIRAGE--FFT
    & \underline{95.02 $\pm$ 0.76}
    & \underline{90.73 $\pm$ 1.30}
    & \textbfbg{2.18 $\pm$ 0.47}
    \\

\cmidrule(lr){2-6}

& \multirow{6}{*}{\textbf{AROI (lesions)}}
    & SwinUNETR-V2
    & 48.98 $\pm$ 13.64
    & 38.20 $\pm$ 10.40
    & 41.27 $\pm$ 13.17
    \\

    & & MedNeXt
    & 49.70 $\pm$ 14.62
    & 39.22 $\pm$ 11.92
    & 37.85 $\pm$ 16.77
    \\

    & & TransUNet
    & 54.30 $\pm$ 16.23
    & 43.50 $\pm$ 12.94
    & \textbf{35.82 $\pm$ 13.57}
    \\

    & & nnUNet
    & \textbfm{62.99 $\pm$ 21.28}
    & \textbfm{51.87 $\pm$ 19.16}
    & 49.65 $\pm$ 30.74
    \\

    & & MIRAGE
    & 52.18 $\pm$ 16.63
    & 40.89 $\pm$ 13.70
    & \underline{37.56 $\pm$ 12.17}
    \\

    & & MIRAGE--FFT
    & \underline{60.51 $\pm$ 18.79}
    & \underline{48.74 $\pm$ 15.82}
    & 42.09 $\pm$ 14.84
    \\

\cmidrule(lr){2-6}

& \multirow{6}{*}{\textbf{Duke DME (layers)}}
    & SwinUNETR-V2
    & 81.90 $\pm$ 3.10
    & 69.88 $\pm$ 4.26
    & 15.14 $\pm$ 5.79
    \\

    & & MedNeXt
    & 80.87 $\pm$ 3.48
    & 68.44 $\pm$ 4.76
    & 12.51 $\pm$ 5.73
    \\

    & & TransUNet
    & 81.52 $\pm$ 3.25
    & 69.36 $\pm$ 4.42
    & 9.81 $\pm$ 5.44
    \\

    & & nnUNet
    & \textbfm{83.49 $\pm$ 2.55}
    & \textbfm{72.15 $\pm$ 3.53}
    & \underline{9.31 $\pm$ 4.06}
    \\

    & & MIRAGE
    & 83.02 $\pm$ 2.93
    & 71.41 $\pm$ 4.18
    & 9.81 $\pm$ 4.77
    \\

    & & MIRAGE--FFT
    & \underline{83.44 $\pm$ 3.00}
    & \underline{72.08 $\pm$ 4.20}
    & \textbfbg{7.85 $\pm$ 3.54}
    \\

\cmidrule(lr){2-6}

& \multirow{6}{*}{\textbf{Duke DME (lesions)}}
    & SwinUNETR-V2
    & 54.44 $\pm$ 11.16
    & 38.04 $\pm$ 10.56
    & 111.72 $\pm$ 98.36
    \\

    & & MedNeXt
    & 63.69 $\pm$ 8.96
    & 47.24 $\pm$ 9.85
    & 77.86 $\pm$ 53.93
    \\

    & & TransUNet
    & 65.12 $\pm$ 10.24
    & 48.97 $\pm$ 11.28
    & 51.15 $\pm$ 18.93
    \\

    & & nnUNet
    & \underline{66.05 $\pm$ 10.70}
    & \underline{50.05 $\pm$ 11.55}
    & \textbfm{38.11 $\pm$ 12.22}
    \\

    & & MIRAGE
    & 69.72 $\pm$ 9.16
    & 54.14 $\pm$ 11.14
    & 42.45 $\pm$ 22.88
    \\

    & & MIRAGE--FFT
    & \textbfbg{70.04 $\pm$ 7.99}
    & \textbfbg{54.36 $\pm$ 9.66}
    & \underline{38.86 $\pm$ 22.73}
    \\

\cmidrule(lr){2-6}

& \multirow{6}{*}{\textbf{GOALS (layers)}}
    & SwinUNETR-V2
    & 91.77
    & 84.92
    & 8.41
    \\

    & & MedNeXt
    & 91.55
    & 84.54
    & 6.86
    \\

    & & TransUNet
    & 92.38
    & 85.97
    & 7.04
    \\

    & & nnUNet
    & \textbfm{93.04}
    & \textbfm{87.08}
    & \textbfm{4.13}
    \\

    & & MIRAGE
    & \underline{92.46}
    & \underline{86.08}
    & \underline{4.96}
    \\

    & & MIRAGE--FFT
    & 92.01
    & 85.30
    & 5.32
    \\

\cmidrule(lr){2-6}

& \multirow{6}{*}{\textbf{RETOUCH (lesions)}}
    & SwinUNETR-V2
    & 74.29
    & -
    & 0.06\textsuperscript{\textdagger}
    \\

    & & MedNeXt
    & \underline{77.43}
    & -
    & 0.04\textsuperscript{\textdagger}
    \\

    & & TransUNet
    & 76.87
    & -
    & 0.05\textsuperscript{\textdagger}
    \\

    & & nnUNet
    & 72.93
    & -
    & 0.06\textsuperscript{\textdagger}
    \\

    & & MIRAGE
    & 79.60
    & -
    & \textbfb{0.03}\textsuperscript{\textdagger}
    \\

    & & MIRAGE--FFT
    & \textbfbg{79.61}
    & -
    & 0.04\textsuperscript{\textdagger}
    \\

\cmidrule(lr){2-6}

& \multirow{6}{*}{\textbf{\textit{Average}}}
    & SwinUNETR-V2
    & 74.23 $\pm$ 17.26
    & 64.02 $\pm$ 22.09
    & 36.04 $\pm$ 40.01
    \\

    & & MedNeXt
    & 76.22 $\pm$ 15.48
    & 65.73 $\pm$ 19.78
    & 27.73 $\pm$ 27.80
    \\

    & & TransUNet
    & 77.47 $\pm$ 14.28
    & 67.58 $\pm$ 18.84
    & 21.31 $\pm$ 18.88
    \\

    & & nnUNet
    & \underline{78.92 $\pm$ 12.49}
    & \textbfm{70.39 $\pm$ 17.06}
    & \underline{20.71 $\pm$ 19.41}
    \\

    & & MIRAGE
    & 78.46 $\pm$ 14.26
    & 68.24 $\pm$ 18.40
    & 19.61 $\pm$ 16.87
    \\

    & & MIRAGE--FFT
    & \textbfbg{80.10 $\pm$ 11.98}
    & \underline{70.24 $\pm$ 16.52}
    & \textbfbg{19.26 $\pm$ 17.45}
    \\

\midrule

& & SwinUNETR-V2
    & 46.88 $\pm$ 1.72
    & 43.90 $\pm$ 2.70
    & 201.81 $\pm$ 24.71
    \\

    &
    & MedNeXt
    & 47.05 $\pm$ 1.47
    & 44.05 $\pm$ 2.34
    & 203.56 $\pm$ 29.73
    \\

    \multirow{1}{*}{\textbf{OCT}}
    & \multirow{1}{*}{\textbf{Duke iAMD (layers)}}
    & TransUNet
    & 46.88 $\pm$ 1.59
    & 43.76 $\pm$ 2.61
    & 194.12 $\pm$ 36.45
    \\

    {\small cross-dataset}
    & {\small trained on AROI}
    & nnUNet
    & \underline{61.21 $\pm$ 15.33}
    & \underline{55.35 $\pm$ 12.98}
    & \underline{97.95 $\pm$ 60.82}
    \\

    & & MIRAGE
    & 91.06 $\pm$ 2.43
    & 84.62 $\pm$ 3.38
    & 4.62 $\pm$ 4.52
    \\

    & & MIRAGE--FFT
    & \textbfbg{91.29 $\pm$ 2.14}***
    & \textbfbg{84.92 $\pm$ 3.15}***
    & \textbfbg{3.94 $\pm$ 2.27}***
    \\

\midrule

\multirow{6}{*}{\textbf{SLO}}
& \multirow{6}{*}{\textbf{SGA (lesions)}}
    & SwinUNETR-V2
    & 74.51 $\pm$ 23.51
    & 65.09 $\pm$ 24.88
    & 169.03 $\pm$ 94.67
    \\

    & & MedNeXt
    & 77.22 $\pm$ 19.62
    & 67.87 $\pm$ 22.19
    & 157.15 $\pm$ 84.21
    \\

    & & TransUNet
    & \textbf{82.12 $\pm$ 15.92}*
    & \textbf{73.69 $\pm$ 18.65}*
    & \textbf{127.19 $\pm$ 57.98}
    \\

    & & nnUNet
    & \underline{79.41 $\pm$ 19.26}
    & \underline{71.33 $\pm$ 21.25}
    & 136.56 $\pm$ 74.10
    \\

    & & MIRAGE
    & 75.31 $\pm$ 22.16
    & 65.81 $\pm$ 24.14
    & 164.42 $\pm$ 95.35
    \\

    & & MIRAGE--FFT
    & 79.36 $\pm$ 17.07
    & 70.03 $\pm$ 19.92
    & \underline{136.16 $\pm$ 43.48}
    \\

    \bottomrule
\end{tabular}

    }
\end{table}


\begin{table}[tbhp]
    \centering
    \caption{\textbf{Performance of a ViT-Base model on the downstream classification tasks with linear probing pretrained with and without retinal layer pseudo-labels on our VIBES~\cite{gerendas2022validation} dataset.}
        Statistical significance between the best (bold) and second best models in each and across all datasets was assessed using the one-tailed Student's $t$-test and Wilcoxon signed-rank test, respectively (*$p<0.05$, **$p<0.01$, ***$p<0.001$).
    }%
    \label{sup:tab:pseudo_labeling_class}
    \resizebox{\textwidth}{!}{%

\begin{tabular}{lllll}
    \toprule

    \textbf{Dataset}
    & \textbf{Modalities}
    & \textbf{AUROC}
    & \textbf{AP}
    & \textbf{BAcc}
    \\

\midrule

\multirow{2}{*}{\textbf{Duke iAMD}}
& OCT
& 98.89 $\pm$ 0.06
& 99.01 $\pm$ 0.05
& 93.08 $\pm$ 0.80
\\

& OCT+Layers
& \textbfli{99.58 $\pm$ 0.03}***
& \textbfli{99.60 $\pm$ 0.03}***
& \textbfli{95.72 $\pm$ 1.03}**
\\

\midrule

\multirow{2}{*}{\textbf{GAMMA}}
& OCT
& 81.51 $\pm$ 0.18
& \textbfs{73.62 $\pm$ 2.34}
& 52.22 $\pm$ 1.67
\\

& OCT+Layers
& \textbfli{84.86 $\pm$ 0.36}***
& 72.98 $\pm$ 0.73
& \textbfli{52.50 $\pm$ 2.22}
\\

\midrule

\multirow{2}{*}{\textbf{Harvard Glaucoma}}
& OCT
& 76.61 $\pm$ 0.95
& 77.02 $\pm$ 0.69
& 69.67 $\pm$ 0.81
\\

& OCT+Layers
& \textbfli{83.92 $\pm$ 0.98}***
& \textbfli{83.87 $\pm$ 0.87}***
& \textbfli{77.17 $\pm$ 1.89}**
\\

\midrule

\multirow{2}{*}{\textbf{Kermany}}
& OCT
& 98.64 $\pm$ 0.07
& 95.84 $\pm$ 0.22
& 84.20 $\pm$ 0.39
\\

& OCT+Layers
& \textbfli{98.93 $\pm$ 0.01}***
& \textbfli{96.66 $\pm$ 0.07}***
& \textbfli{87.60 $\pm$ 0.42}***
\\

\midrule

\multirow{2}{*}{\textbf{Noor Eye Hospital}}
& OCT
& 97.50 $\pm$ 0.75
& 96.02 $\pm$ 1.08
& 86.00 $\pm$ 3.89
\\

& OCT+Layers
& \textbfli{99.60 $\pm$ 0.50}**
& \textbfli{99.27 $\pm$ 0.91}**
& \textbfli{94.67 $\pm$ 1.63}**
\\

\midrule

\multirow{2}{*}{\textbf{OCTDL}}
& OCT
& 96.65 $\pm$ 0.63
& 90.76 $\pm$ 1.32
& 73.96 $\pm$ 2.09
\\

& OCT+Layers
& \textbfli{98.41 $\pm$ 0.10}**
& \textbfli{94.12 $\pm$ 0.16}**
& \textbfli{80.92 $\pm$ 0.78}**
\\

\midrule

\multirow{2}{*}{\textbf{OCTID}}
& OCT
& 98.10 $\pm$ 0.07
& 94.95 $\pm$ 0.22
& 84.41 $\pm$ 0.32
\\

& OCT+Layers
& \textbfli{98.60 $\pm$ 0.04}***
& \textbfli{95.10 $\pm$ 0.08}
& \textbfli{85.78 $\pm$ 0.46}***
\\

\midrule

\multirow{2}{*}{\textbf{OLIVES}}
& OCT
& \textbfs{96.85 $\pm$ 0.03}**
& \textbfs{95.84 $\pm$ 0.09}*
& \textbfs{96.67 $\pm$ 0.41}*
\\

& OCT+Layers
& 96.28 $\pm$ 0.21
& 94.66 $\pm$ 1.08
& 92.81 $\pm$ 2.14
\\

\midrule

\multirow{2}{*}{\textbf{OPTIMA9C}}
& OCT
& \textbfs{99.02 $\pm$ 0.01}***
& \textbfs{93.51 $\pm$ 0.10}**
& 77.68 $\pm$ 0.55
\\

& OCT+Layers
& 98.82 $\pm$ 0.04
& 93.26 $\pm$ 0.10
& \textbfli{78.15 $\pm$ 0.73}
\\

\midrule

\multirow{2}{*}{\textbf{\textbf{\textit{Average}}}}
& OCT
& 93.75 $\pm$ 7.99
& 90.73 $\pm$ 8.59
& 79.77 $\pm$ 12.76
\\

& OCT+Layers
& \textbfli{95.44 $\pm$ 6.00}***
& \textbfli{92.17 $\pm$ 8.07}***
& \textbfli{82.81 $\pm$ 12.62}***
\\

    \bottomrule
\end{tabular}

    }
\end{table}

\begin{table}[tbhp]
    \centering
    \caption{\textbf{Performance of a ViT-Base model on the downstream segmentation tasks with linear probing pretrained with and without retinal layer pseudo-labels on our VIBES~\cite{gerendas2022validation} dataset.}
    Statistical significance between the best (bold) and second best models in each and across all datasets was assessed using the one-tailed Student's $t$-test and Wilcoxon signed-rank test, respectively (*$p<0.05$, **$p<0.01$, ***$p<0.001$).
    }%
    \label{sup:tab:pseudo_labeling_seg}
    \resizebox{\textwidth}{!}{%

\begin{tabular}{llllll}
    \toprule

    \textbf{Modality}
    & \textbf{Dataset}
    & \textbf{Model}
    & \textbf{Dice}
    & \textbf{IoU}
    & \textbf{HD95 / AVD\textsuperscript{\textdagger}} $\downarrow$
    \\

    \midrule

\multirow{14}{*}{\textbf{OCT}}
& \multirow{2}{*}{\textbf{AROI (layers)}}
    & OCT                    
    & 84.12 $\pm$ 2.89 
    & 73.58 $\pm$ 4.03 
    & 13.52 $\pm$ 4.93
    \\

    & & OCT+Layers           
    & \textbfli{89.47 $\pm$ 1.76}*** 
    & \textbfli{81.65 $\pm$ 2.55}*** 
    & \textbfli{9.09 $\pm$ 4.25}*
    \\

\cmidrule(lr){2-6}

& \multirow{2}{*}{\textbf{AROI (lesions)}}
    & OCT                    
    & 42.54 $\pm$ 17.83 
    & 30.29 $\pm$ 14.28 
    & 72.68 $\pm$ 40.85
    \\

    & & OCT+Layers           
    & \textbfli{47.27 $\pm$ 18.17} 
    & \textbfli{34.84 $\pm$ 14.65} 
    & \textbfli{56.10 $\pm$ 28.64}*
    \\

\cmidrule(lr){2-6}

& \multirow{2}{*}{\textbf{Duke DME (layers)}}
    & OCT                    
    & 72.23 $\pm$ 3.81 
    & 57.38 $\pm$ 4.62 
    & \textbfs{23.47 $\pm$ 11.61}
    \\

    & & OCT+Layers           
    & \textbfli{72.99 $\pm$ 4.15} 
    & \textbfli{58.37 $\pm$ 5.02}* 
    & 27.88 $\pm$ 14.52
    \\

\cmidrule(lr){2-6}

& \multirow{2}{*}{\textbf{Duke DME (lesions)}}
    & OCT                    
    & \textbfs{57.42 $\pm$ 16.29} 
    & \textbfs{41.63 $\pm$ 15.00} 
    & 65.07 $\pm$ 41.03
    \\

    & & OCT+Layers           
    & 56.73 $\pm$ 11.92 
    & 40.34 $\pm$ 11.12 
    & \textbfli{52.89 $\pm$ 28.38}
    \\

\cmidrule(lr){2-6}

& \multirow{2}{*}{\textbf{GOALS (layers)}}
    & OCT                    
    & 77.45    
    & 63.74    
    & 29.41   
    \\

    & & OCT+Layers           
    & \textbfli{81.20} 
    & \textbfli{68.80} 
    & \textbfli{25.70}
    \\

\cmidrule(lr){2-6}

& \multirow{2}{*}{\textbf{RETOUCH (lesions)}}
    & OCT                    
    & 64.49    
    & -                    
    & 0.09\textsuperscript{\textdagger} 
    \\

    & & OCT+Layers           
    & \textbfli{66.78} 
    & - 
    & \textbfli{0.06}\textsuperscript{\textdagger} 
    \\

\cmidrule(lr){2-6}

& \multirow{2}{*}{\textbf{\textit{Average}}}
    & OCT                    
    & 66.37 $\pm$ 13.68 
    & 53.32 $\pm$ 15.51 
    & 40.83 $\pm$ 23.58
    \\

    & & OCT+Layers           
    & \textbfli{69.07 $\pm$ 14.21}* 
    & \textbfli{56.80 $\pm$ 17.42} 
    & \textbfli{34.33 $\pm$ 17.73}
    \\

\midrule

\multirow{1}{*}{\textbf{OCT}}
& \multirow{1}{*}{\textbf{Duke iAMD (layers)}}
    & OCT                    
    & 55.40 $\pm$ 6.90 
    & 39.18 $\pm$ 6.09 
    & \textbfs{214.22 $\pm$ 19.67}***
    \\

    {\small cross-dataset}
    & {\small trained on AROI}
    & OCT+Layers           
    & \textbfli{65.91 $\pm$ 3.91}*** 
    & \textbfli{49.65 $\pm$ 3.86}*** 
    & 222.92 $\pm$ 15.67
    \\

    \bottomrule
\end{tabular}

    }
\end{table}


\begin{table}[tbhp]
    \centering
    \caption{\textbf{Performance of a ViT-Base model on the downstream classification tasks with linear probing for different pretraining strategies.}
    The strategies include MultiMAE~\cite{bachmann2022multimae}, pretrained on ImageNet using multimodal data, MAE-OCT and MAE-SLO, trained using MAE~\cite{he2022mae} on the OCT or SLO images of our VIBES~\cite{gerendas2022validation} dataset, respectively, and MIRAGE, our proposed model based on multimodal MAE trained on our multimodal VIBES dataset.
    Statistical significance between the best (bold) and second best (underlined) models in each and across all datasets was assessed using the one-tailed Student's $t$-test and Wilcoxon signed-rank test, respectively (*$p<0.05$, **$p<0.01$, ***$p<0.001$).
    }%
    \label{sup:tab:impact_pretraining}
    \resizebox{\textwidth}{!}{%
    
        \begin{tabular}{llllll}
            \toprule

            \textbf{Tuning}
            & \textbf{Dataset}
            & \textbf{Model}
            & \textbf{AUROC}
            & \textbf{AP}
            & \textbf{BAcc}
            \\

            \textbf{modality} \\

\midrule

\multirow{30}{*}{\textbf{OCT}}
& \multirow{3}{*}{\textbf{Duke iAMD}}
& MultiMAE
& 94.86 $\pm$ 2.41
& 95.91 $\pm$ 1.85
& 88.08 $\pm$ 1.87
\\

&
& MAE-OCT
& \underline{98.89 $\pm$ 0.06}
& \underline{99.01 $\pm$ 0.05}
& \underline{93.08 $\pm$ 0.80}
\\

&
& MIRAGE
& \textbfdbg{99.05 $\pm$ 0.09}*
& \textbfdbg{99.18 $\pm$ 0.07}*
& \textbfdbg{94.68 $\pm$ 1.26}*
\\

\cmidrule{2-6}

& \multirow{3}{*}{\textbf{GAMMA}}
& MultiMAE
& 74.64 $\pm$ 1.40
& 61.20 $\pm$ 1.60
& \underline{55.83 $\pm$ 2.22}
\\

&
& MAE-OCT
& \underline{81.51 $\pm$ 0.18}
& \underline{73.62 $\pm$ 2.34}
& 52.22 $\pm$ 1.67
\\

&
& MIRAGE
& \textbfdbg{85.52 $\pm$ 0.23}***
& \textbfdbg{74.26 $\pm$ 0.57}
& \textbfdbg{58.89 $\pm$ 4.27}
\\

\cmidrule{2-6}

& \multirow{3}{*}{\textbf{Harvard Glaucoma}}
& MultiMAE
& 72.06 $\pm$ 1.43
& 70.90 $\pm$ 1.57
& 65.63 $\pm$ 2.21
\\

&
& MAE-OCT
& \underline{76.61 $\pm$ 0.95}
& \underline{77.02 $\pm$ 0.69}
& \underline{69.67 $\pm$ 0.81}
\\

&
& MIRAGE
& \textbfdbg{78.57 $\pm$ 0.18}**
& \textbfdbg{78.95 $\pm$ 0.19}**
& \textbfdbg{70.48 $\pm$ 0.40}*
\\

\cmidrule{2-6}

& \multirow{3}{*}{\textbf{Kermany}}
& MultiMAE
& 97.66 $\pm$ 0.02
& 93.22 $\pm$ 0.06
& 79.44 $\pm$ 0.93
\\

&
& MAE-OCT
& \underline{98.64 $\pm$ 0.07}
& \underline{95.84 $\pm$ 0.22}
& \underline{84.20 $\pm$ 0.39}
\\

&
& MIRAGE
& \textbfdbg{98.77 $\pm$ 0.03}*
& \textbfdbg{96.23 $\pm$ 0.09}*
& \textbfdbg{85.84 $\pm$ 0.32}***
\\

\cmidrule{2-6}

& \multirow{3}{*}{\textbf{Noor Eye Hospital}}
& MultiMAE
& 96.73 $\pm$ 0.54
& 95.81 $\pm$ 0.46
& \underline{89.33 $\pm$ 1.33}
\\

&
& MAE-OCT
& \underline{97.50 $\pm$ 0.75}
& \underline{96.02 $\pm$ 1.08}
& 86.00 $\pm$ 3.89
\\

&
& MIRAGE
& \textbfdbg{98.53 $\pm$ 0.50}
& \textbfdbg{97.52 $\pm$ 0.95}
& \textbfdbg{93.33 $\pm$ 4.22}
\\

\cmidrule{2-6}

& \multirow{3}{*}{\textbf{OCTDL}}
& MultiMAE
& 93.28 $\pm$ 0.54
& 85.56 $\pm$ 0.68
& 67.28 $\pm$ 2.10
\\

&
& MAE-OCT
& \underline{96.65 $\pm$ 0.63}
& \underline{90.76 $\pm$ 1.32}
& \underline{73.96 $\pm$ 2.09}
\\

&
& MIRAGE
& \textbfdbg{97.51 $\pm$ 0.14}*
& \textbfdbg{92.61 $\pm$ 0.32}*
& \textbfdbg{77.83 $\pm$ 0.89}*
\\

\cmidrule{2-6}

& \multirow{3}{*}{\textbf{OCTID}}
& MultiMAE
& 97.15 $\pm$ 0.17
& 89.43 $\pm$ 0.50
& 75.51 $\pm$ 0.85
\\

&
& MAE-OCT
& \underline{98.10 $\pm$ 0.07}
& \underline{94.95 $\pm$ 0.22}
& \underline{84.41 $\pm$ 0.32}
\\

&
& MIRAGE
& \textbfdbg{98.81 $\pm$ 0.24}**
& \textbfdbg{95.90 $\pm$ 0.82}*
& \textbfdbg{85.83 $\pm$ 1.41}
\\

\cmidrule{2-6}

& \multirow{3}{*}{\textbf{OLIVES}}
& MultiMAE
& \textbft{97.03 $\pm$ 0.31}
& \underline{95.10 $\pm$ 1.04}
& \textbft{96.70 $\pm$ 0.39}
\\

&
& MAE-OCT
& \underline{96.85 $\pm$ 0.03}
& \textbfs{95.84 $\pm$ 0.09}
& \underline{96.67 $\pm$ 0.41}
\\

&
& MIRAGE
& 95.00 $\pm$ 0.13
& 91.83 $\pm$ 0.12
& 91.11 $\pm$ 0.07
\\

\cmidrule{2-6}

& \multirow{3}{*}{\textbf{OPTIMA9C}}
& MultiMAE
& 95.64 $\pm$ 0.25
& 81.93 $\pm$ 0.68
& 63.71 $\pm$ 2.96
\\

&
& MAE-OCT
& \textbfs{99.02 $\pm$ 0.01}**
& \underline{93.51 $\pm$ 0.10}
& \underline{77.68 $\pm$ 0.55}
\\

&
& MIRAGE
& \underline{98.91 $\pm$ 0.05}
& \textbfdbg{93.58 $\pm$ 0.05}
& \textbfdbg{78.54 $\pm$ 0.10}*
\\

\cmidrule{2-6}


& \multirow{3}{*}{\textbf{\textit{Average}}}
& MultiMAE
& 91.01 $\pm$ 9.60
& 85.45 $\pm$ 11.59
& 75.73 $\pm$ 13.06
\\

&
& MAE-OCT
& \underline{93.75 $\pm$ 7.99}
& \underline{90.73 $\pm$ 8.59}
& \underline{79.77 $\pm$ 12.76}
\\

&
& MIRAGE
& \textbfdbg{94.52 $\pm$ 6.97}***
& \textbfdbg{91.12 $\pm$ 8.15}**
& \textbfdbg{81.84 $\pm$ 11.24}***
\\

\midrule

\multirow{9}{*}{\textbf{SLO}}
& \multirow{3}{*}{\textbf{OPTIMA9C}}
& MultiMAE
& \underline{89.88 $\pm$ 0.95}
& \underline{70.19 $\pm$ 2.22}
& \underline{46.60 $\pm$ 2.30}
\\

&
& MAE-SLO
& 84.77 $\pm$ 0.48
& 59.67 $\pm$ 1.02
& 36.23 $\pm$ 1.46
\\

&
& MIRAGE
& \textbfdbg{92.83 $\pm$ 0.25}**
& \textbfdbg{75.60 $\pm$ 0.50}**
& \textbfdbg{51.30 $\pm$ 0.69}*
\\

\cmidrule{2-6}

& \multirow{3}{*}{\textbf{OLIVES}}
& MultiMAE
& \underline{77.17 $\pm$ 1.14}
& \underline{76.03 $\pm$ 1.21}
& 65.65 $\pm$ 1.51
\\

&
& MAE-SLO
& 64.57 $\pm$ 0.38
& 62.19 $\pm$ 0.49
& \underline{66.98 $\pm$ 0.54}
\\

&
& MIRAGE
& \textbfdbg{78.49 $\pm$ 0.74}
& \textbfdbg{77.49 $\pm$ 0.78}
& \textbfdbg{71.79 $\pm$ 0.98}***
\\

\cmidrule{2-6}


& \multirow{3}{*}{\textbf{\textit{Average}}}
& MultiMAE
& \underline{83.52 $\pm$ 6.44}
& \underline{73.11 $\pm$ 3.42}
& \underline{56.13 $\pm$ 9.72}
\\

&
& MAE-SLO
& 74.67 $\pm$ 10.11
& 60.93 $\pm$ 1.49
& 51.61 $\pm$ 15.42
\\

&
& MIRAGE
& \textbfdbg{85.66 $\pm$ 7.19}**
& \textbfdbg{76.54 $\pm$ 1.15}**
& \textbfdbg{61.55 $\pm$ 10.28}**
\\

            \bottomrule
        \end{tabular}

    }
\end{table}

\begin{table}[tbhp]
    \centering
    \caption{\textbf{Performance of a ViT-Base model on the downstream segmentation tasks with linear probing for different pretraining strategies}
    The strategies include MultiMAE~\cite{bachmann2022multimae}, pretrained on ImageNet using multimodal data, MAE-OCT, and MAE-SLO, both trained using MAE~\cite{he2022mae} on the OCT or SLO images of our VIBES~\cite{gerendas2022validation} dataset, respectively, and MIRAGE, our proposed model based on multimodal MAE trained on our multimodal VIBES dataset.
    Statistical significance between the best (bold) and second best (underlined) models in each and across all datasets was assessed using the one-tailed Student's $t$-test and Wilcoxon signed-rank test, respectively (*$p<0.05$, **$p<0.01$, ***$p<0.001$).
    }%
    \label{sup:tab:impact_pretraining_seg}
    \resizebox{\textwidth}{!}{%

\begin{tabular}{llllll}
    \toprule

    \textbf{Modality}
    & \textbf{Dataset}
    & \textbf{Model}
    & \textbf{Dice}
    & \textbf{IoU}
    & \textbf{HD95 / AVD\textsuperscript{\textdagger}} $\downarrow$
    \\

    \midrule

\multirow{21}{*}{\textbf{OCT}}
& \multirow{3}{*}{\textbf{AROI (layers)}}
    & MultiMAE
    & 74.01 $\pm$ 3.16
    & 60.28 $\pm$ 3.93
    & 32.46 $\pm$ 13.69
    \\

    & & MAE-OCT
    & \underline{84.12 $\pm$ 2.89}
    & \underline{73.58 $\pm$ 4.03}
    & \underline{13.52 $\pm$ 4.93}
    \\

    & & MIRAGE
    & \textbfdbg{89.73 $\pm$ 1.48}***
    & \textbfdbg{81.93 $\pm$ 2.32}***
    & \textbfdbg{8.59 $\pm$ 3.52}**
    \\

\cmidrule(lr){2-6}

& \multirow{3}{*}{\textbf{AROI (lesions)}}
    &  MultiMAE
    &  17.72 $\pm$ 7.38
    &  11.23 $\pm$ 5.02
    &  80.69 $\pm$ 28.04
    \\

    &
    & MAE-OCT
    & \underline{42.54 $\pm$ 17.83}
    & \underline{30.29 $\pm$ 14.28}
    & \underline{72.68 $\pm$ 40.85}
    \\

    & & MIRAGE
    & \textbfdbg{43.84 $\pm$ 14.84}
    & \textbfdbg{31.92 $\pm$ 11.52}
    & \textbfdbg{69.78 $\pm$ 69.78}
    \\

\cmidrule(lr){2-6}

& \multirow{3}{*}{\textbf{Duke DME (layers)}}
    & MultiMAE
    & 55.88 $\pm$ 8.49
    & 40.97 $\pm$ 8.14
    & 38.03 $\pm$ 23.22
    \\

    & & MAE-OCT
    & \underline{72.23 $\pm$ 3.81}
    & \underline{57.38 $\pm$ 4.62}
    & \underline{23.47 $\pm$ 11.61}
    \\

    &
    & MIRAGE
    & \textbfdbg{75.03 $\pm$ 3.85}***
    & \textbfdbg{60.80 $\pm$ 4.91}***
    & \textbfdbg{17.44 $\pm$ 9.93}**
    \\

\cmidrule(lr){2-6}

& \multirow{3}{*}{\textbf{Duke DME (lesions)}}
    & MultiMAE
    & \underline{1.29 $\pm$ 1.78}
    & \underline{0.65 $\pm$ 0.91}
    & \underline{165.52 $\pm$ 77.51}
    \\

    &
    & MAE-OCT
    & 57.42 $\pm$ 16.29
    & 41.63 $\pm$ 15.00
    & 65.07 $\pm$ 41.03
    \\

    & & MIRAGE
    & \textbfdbg{62.68 $\pm$ 7.72}
    & \textbfdbg{46.01 $\pm$ 8.23}
    & \textbfdbg{43.54 $\pm$ 24.19}
    \\

\cmidrule(lr){2-6}

& \multirow{3}{*}{\textbf{GOALS (layers)}}
    & MultiMAE
    & 65.30
    & 48.97
    & 85.58
    \\

    & & MAE-OCT
    & \underline{77.45}
    & \underline{63.74}
    & \textbfs{29.41}
    \\

    &
    & MIRAGE
    & \textbfdbg{80.52}
    & \textbfdbg{68.07}
    & \underline{32.27}
    \\

\cmidrule(lr){2-6}

& \multirow{3}{*}{\textbf{RETOUCH (lesions)}}
    & MultiMAE
    & \underline{32.34}
    & -
    & 0.16\textsuperscript{\textdagger}
    \\

    &
    & MAE-OCT
    & 64.49
    & -
    & \underline{0.09}\textsuperscript{\textdagger}
    \\

    & & MIRAGE
    & \textbfdbg{65.99}
    & -
    & \textbfdbg{0.07}\textsuperscript{\textdagger}
    \\

\cmidrule(lr){2-6}

& \multirow{3}{*}{\textbf{\textit{Average}}}
    & MultiMAE
    & 41.09 $\pm$ 26.13
    & 32.42 $\pm$ 22.72
    & 80.46 $\pm$ 47.68
    \\

    & & MAE-OCT
    & \underline{66.37 $\pm$ 13.68}
    & \underline{53.32 $\pm$ 15.51}
    & \underline{40.83 $\pm$ 23.58}
    \\

    &
    & MIRAGE
    & \textbfdbg{69.63 $\pm$ 14.60}*
    & \textbfdbg{57.75 $\pm$ 17.35}*
    & \textbfdbg{34.33 $\pm$ 21.42}
    \\

\midrule

\multirow{1}{*}{\textbf{OCT}}
& \multirow{1}{*}{\textbf{Duke iAMD (layers)}}
    & MultiMAE
    & 33.97 $\pm$ 11.20
    & 21.32 $\pm$ 8.22
    & 226.25 $\pm$ 27.87
    \\

    {\small cross-dataset}
    & {\small trained on AROI}
    & MAE-OCT
    & \underline{55.40 $\pm$ 6.90}
    & \underline{39.18 $\pm$ 6.09}
    & \textbfs{214.22 $\pm$ 19.67}***
    \\

    & & MIRAGE
    & \textbfdbg{63.32 $\pm$ 4.19}***
    & \textbfdbg{46.76 $\pm$ 4.03}***
    & \underline{217.92 $\pm$ 16.78}
    \\

\midrule

\multirow{3}{*}{\textbf{SLO}}
& \multirow{3}{*}{\textbf{SGA (lesions)}}
    & MultiMAE
    & 68.69 $\pm$ 23.43
    & 57.76 $\pm$ 24.21
    & 187.07 $\pm$ 88.52
    \\

    & & MAE-SLO
    & \underline{70.63 $\pm$ 22.54}
    & \underline{59.79 $\pm$ 23.36}
    & \underline{174.96 $\pm$ 78.10}
    \\

    & & MIRAGE
    & \textbfdbg{72.24 $\pm$ 21.68}
    & \textbfdbg{61.79 $\pm$ 23.55}
    & \textbfdbg{166.25 $\pm$ 73.90}
    \\

    \bottomrule
\end{tabular}

    }
\end{table}


\begin{table}[tbhp]
    \centering
    \caption{\textbf{Performance of the ViT-Large and ViT-Base versions of MIRAGE on the downstream classification tasks using linear probing.}
    Statistical significance between the best (bold) and second best models in each and across all datasets was assessed using the one-tailed Student's $t$-test and Wilcoxon signed-rank test, respectively (*$p<0.05$, **$p<0.01$, ***$p<0.001$).
    }%
    \label{sup:tab:large_vs_base_class}
    \resizebox{\textwidth}{!}{%
    
        \begin{tabular}{llllll}
            \toprule

            \textbf{Modality}
            & \textbf{Dataset}
            & \textbf{Model}
            & \textbf{AUROC}
            & \textbf{AP}
            & \textbf{BAcc}
            \\

\midrule

\multirow{24}{*}{\textbf{OCT}}
& \multirow{2}{*}{\textbf{Duke iAMD}}
& MIRAGE-B
& 99.05 $\pm$ 0.09
& 99.18 $\pm$ 0.07
& 94.68 $\pm$ 1.26
\\

& &  MIRAGE-L
& \textbfb{99.52 $\pm$ 0.23}*
& \textbfb{99.52 $\pm$ 0.23}*
& \textbfb{96.71 $\pm$ 1.42}
\\

\cmidrule{2-6}

& \multirow{2}{*}{\textbf{GAMMA}}
& MIRAGE-B
& 85.52 $\pm$ 0.23
& 74.26 $\pm$ 0.57
& 58.89 $\pm$ 4.27
\\

            & &  MIRAGE-L
& \textbfb{87.50 $\pm$ 0.64}**
& \textbfb{81.00 $\pm$ 1.89}***
& \textbfb{63.61 $\pm$ 4.16}
\\

\cmidrule{2-6}

& \multirow{2}{*}{\textbf{Harvard Glaucoma}}
& MIRAGE-B
& 78.57 $\pm$ 0.18
& 78.95 $\pm$ 0.19
& 70.48 $\pm$ 0.40
\\

& &  MIRAGE-L
& \textbfb{82.75 $\pm$ 0.65}***
& \textbfb{82.54 $\pm$ 0.68}***
& \textbfb{76.15 $\pm$ 1.25}***
\\

\cmidrule{2-6}

& \multirow{2}{*}{\textbf{Kermany}}
& MIRAGE-B
& 98.77 $\pm$ 0.03
& 96.23 $\pm$ 0.09
& 85.84 $\pm$ 0.32
\\

& &  MIRAGE-L
& \textbfb{99.53 $\pm$ 0.01}***
& \textbfb{98.39 $\pm$ 0.04}***
& \textbfb{91.40 $\pm$ 0.25}***
\\

\cmidrule{2-6}

& \multirow{2}{*}{\textbf{Noor Eye Hospital}}
& MIRAGE-B
& 98.53 $\pm$ 0.50
& 97.52 $\pm$ 0.95
& \textbfdbg{93.33 $\pm$ 4.22}
\\

& &  MIRAGE-L
& \textbfb{98.63 $\pm$ 0.40}
& \textbfb{97.69 $\pm$ 0.68}
& 92.67 $\pm$ 3.27
\\

\cmidrule{2-6}

& \multirow{2}{*}{\textbf{OCTDL}}
& MIRAGE-B
& 97.51 $\pm$ 0.14
& 92.61 $\pm$ 0.32
& \textbfdbg{77.83 $\pm$ 0.89}
\\

& &  MIRAGE-L
&  \textbfb{98.27 $\pm$ 0.15}***
&  \textbfb{93.32 $\pm$ 0.27}**
&  76.93 $\pm$ 2.94
\\

\cmidrule{2-6}

& \multirow{2}{*}{\textbf{OCTID}}
& MIRAGE-B
& 98.81 $\pm$ 0.24
& 95.90 $\pm$ 0.82
& 85.83 $\pm$ 1.41
\\

& &  MIRAGE-L
&  \textbfb{99.07 $\pm$ 0.28}
&  \textbfb{96.54 $\pm$ 0.84}
&  \textbfb{87.53 $\pm$ 1.58}
\\

\cmidrule{2-6}

& \multirow{2}{*}{\textbf{OLIVES}}
& MIRAGE-B
& 95.00 $\pm$ 0.13
& 91.83 $\pm$ 0.12
& 91.11 $\pm$ 0.07
\\

& & MIRAGE-L
& \textbfb{96.06 $\pm$ 0.09}***
& \textbfb{94.36 $\pm$ 0.24}***
& \textbfb{93.14 $\pm$ 0.27}***
\\

\cmidrule{2-6}

& \textbf{OPTIMA9C}
& MIRAGE-B
& 98.91 $\pm$ 0.05
& \textbfdbg{93.58 $\pm$ 0.05}
& \textbfdbg{78.54 $\pm$ 0.10}*
\\

& (in-house)
& MIRAGE-L
& \textbfb{99.01 $\pm$ 0.02}*
& 93.52 $\pm$ 0.16
& 77.95 $\pm$ 0.45
\\

\cmidrule{2-6}

& \multirow{2}{*}{\textbf{\textit{Average}}}
& MIRAGE-B
& 94.52 $\pm$ 6.97
& 91.12 $\pm$ 8.15
& 81.84 $\pm$ 11.24
\\

& & MIRAGE-L
& \textbfb{95.59 $\pm$ 5.80}***
& \textbfb{92.99 $\pm$ 6.39}***
& \textbfb{84.01 $\pm$ 10.51}***
\\

\cmidrule{2-6}

& \textbf{\textit{Average}}
& MIRAGE-B
& 93.97 $\pm$ 7.21
& 90.81 $\pm$ 8.60
& 82.25 $\pm$ 11.86
\\

& (public only, no in-house)
& MIRAGE-L
& \textbfb{95.17 $\pm$ 6.02}***
& \textbfb{92.92 $\pm$ 6.77}***
& \textbfb{84.77 $\pm$ 10.91}***
\\

\midrule

\multirow{7}{*}{\textbf{SLO}}
& \textbf{OPTIMA9C}
& MIRAGE-B
& 92.83 $\pm$ 0.25
& 75.60 $\pm$ 0.50
& 51.30 $\pm$ 0.69
\\

& (in-house)
& MIRAGE-L
& \textbfb{93.40 $\pm$ 0.14}**
& \textbfb{77.38 $\pm$ 0.47}***
& \textbfb{55.55 $\pm$ 0.45}***
\\

\cmidrule{2-6}

& \multirow{2}{*}{\textbf{OLIVES}}
& MIRAGE-B
& \textbfdbg{78.49 $\pm$ 0.74}***
& \textbfdbg{77.49 $\pm$ 0.78}***
& \textbfdbg{71.79 $\pm$ 0.98}**
\\

& & MIRAGE-L
& 74.08 $\pm$ 0.52
& 73.40 $\pm$ 0.52
& 66.72 $\pm$ 1.14
\\

\cmidrule{2-6}

& \multirow{2}{*}{\textbf{\textit{Average}}}
& MIRAGE-B
& \textbfdbg{85.66 $\pm$ 7.19}
& \textbfdbg{76.54 $\pm$ 1.15}
& \textbfdbg{61.55 $\pm$ 10.28}
\\

& & MIRAGE-L
& 83.74 $\pm$ 9.67
& 75.39 $\pm$ 2.05
& 61.13 $\pm$ 5.65
\\

            \bottomrule
        \end{tabular}

    }
\end{table}

\begin{table}[tbhp]
    \centering
    \caption{\textbf{Performance of the ViT-Large and ViT-Base versions of MIRAGE on the downstream segmentation tasks for OCT and SLO datasets using decoder-only fine-tuning and full fine-tuning (FFT).}
    Mean and standard deviation values are calculated across the patients in each dataset, except for the GOALS and RETOUCH datasets, where patient information is not available.
        Statistical significance between the best (bold) and second best models in each and across all datasets was assessed using the one-tailed Student's $t$-test and Wilcoxon signed-rank test, respectively (*$p<0.05$, **$p<0.01$, ***$p<0.001$).
    }%
    \label{sup:tab:large_vs_base_seg}
    \resizebox{\textwidth}{!}{%

\begin{tabular}{llllll}
    \toprule

    \textbf{Modality}
    & \textbf{Dataset}
    & \textbf{Model}
    & \textbf{Dice}
    & \textbf{IoU}
    & \textbf{HD95 / AVD\textsuperscript{\textdagger}} $\downarrow$
    \\

    \midrule

\multirow{14}{*}{\textbf{OCT}}
& \multirow{2}{*}{\textbf{AROI (layers)}}
    & MIRAGE-B
    & \textbfdbg{94.04 $\pm$ 1.18}
    & \textbfdbg{89.12 $\pm$ 1.93}
    & 3.39 $\pm$ 1.12
    \\

    & & MIRAGE-L
    & 93.79 $\pm$ 0.85
    & 88.67 $\pm$ 1.42
    & \textbfb{3.25 $\pm$ 1.43}
    \\

\cmidrule(lr){2-6}

& \multirow{2}{*}{\textbf{AROI (lesions)}}
    & MIRAGE-B
    & 49.12 $\pm$ 16.58
    & 38.11 $\pm$ 13.67
    & 38.64 $\pm$ 14.99
    \\

    & & MIRAGE-L
    & \textbfb{52.18 $\pm$ 16.63}
    & \textbfb{40.89 $\pm$ 13.70}
    & \textbfb{37.56 $\pm$ 12.17}
    \\

\cmidrule(lr){2-6}

& \multirow{2}{*}{\textbf{Duke DME (layers)}}
    & MIRAGE-B
    & 82.86 $\pm$ 3.60
    & 71.23 $\pm$ 5.12
    & \textbfdbg{8.57 $\pm$ 4.01}
    \\

    & & MIRAGE-L
    & \textbfb{83.02 $\pm$ 2.93}
    & \textbfb{71.41 $\pm$ 4.18}
    & 9.81 $\pm$ 4.77
    \\

\cmidrule(lr){2-6}

& \multirow{2}{*}{\textbf{Duke DME (lesions)}}
    & MIRAGE-B
    & 69.06 $\pm$ 9.15
    & 53.36 $\pm$ 11.00
    & 49.32 $\pm$ 30.38
    \\

    & & MIRAGE-L
    & \textbfb{69.72 $\pm$ 9.16}
    & \textbfb{54.14 $\pm$ 11.14}
    & \textbfb{42.45 $\pm$ 22.88}
    \\

\cmidrule(lr){2-6}

& \multirow{2}{*}{\textbf{GOALS (layers)}}
    & MIRAGE-B
    & \textbfdbg{92.87}
    & \textbfdbg{86.79}
    & \textbfdbg{4.87}
    \\

    & & MIRAGE-L
    & 92.46
    & 86.08
    & 4.96
    \\

\cmidrule(lr){2-6}

& \multirow{2}{*}{\textbf{RETOUCH (lesions)}}
    & MIRAGE-B
    & 75.87
    & -
    & 0.06\textsuperscript{\textdagger}
    \\

    & & MIRAGE-L
    & \textbfb{79.60}
    & -
    & \textbfb{0.03}\textsuperscript{\textdagger}
    \\

\cmidrule(lr){2-6}

& \multirow{2}{*}{\textbf{\textit{Average}}}
    & MIRAGE-B
    & 77.30 $\pm$ 15.37
    & 67.72 $\pm$ 19.58
    & 20.96 $\pm$ 19.17
    \\

    & & MIRAGE-L
    & \textbfb{78.46 $\pm$ 14.26}
    & \textbfb{68.24 $\pm$ 18.40}
    & \textbfb{19.61 $\pm$ 16.87}
    \\

\midrule

\multirow{1}{*}{\textbf{OCT}}
& \multirow{1}{*}{\textbf{Duke iAMD (layers)}}
    & MIRAGE-B
    & 90.14 $\pm$ 3.33
    & 83.30 $\pm$ 4.27
    & 12.29 $\pm$ 10.04
    \\

    {\small cross-dataset}
    & {\small trained on AROI}
    & MIRAGE-L
    & \textbfb{91.06 $\pm$ 2.43}***
    & \textbfb{84.62 $\pm$ 3.38}***
    & \textbfb{4.62 $\pm$ 4.52}***
    \\

\midrule

\multirow{2}{*}{\textbf{SLO}}
& \multirow{2}{*}{\textbf{SGA (lesions)}}
    & MIRAGE-B
    & 74.47 $\pm$ 22.11
    & 64.79 $\pm$ 24.23
    & \textbfdbg{161.13 $\pm$ 88.94}
    \\

    & & MIRAGE-L
    & \textbfb{75.31 $\pm$ 22.16}
    & \textbfb{65.81 $\pm$ 24.14}
    & 164.42 $\pm$ 95.35
    \\

    \bottomrule
\end{tabular}

    }
\end{table}

\clearpage

\section{Supplementary Note 2: Benchmark datasets}\label{sup:sec:data}

This section provides a brief description of the 16 datasets used in the evaluation benchmark.
B-scan dimensions are always in the format [B-scan height (A-scan depth) $\times$ B-scan width (\# A-scans)] pixels.
The number of B-scans per volume is indicated in each case.
The field of view (FOV) always refers to the retinal \textit{en-face} view.
The datasets are publicly available unless otherwise noted, and links to the datasets are provided.

\subsection{Classification datasets}


In the following, we provide a brief description of the 11 classification datasets used in the evaluation benchmark.
For all datasets that provide full OCT volumes, we limited our analysis to the central B-scan, as done in previous work~\cite{zhou2023retfound}.
Examples of images from each dataset and the corresponding ground truth labels are shown in Supplementary Figure~\ref{sup:fig:datasets}.

\begin{figure}[tbph]
    \centering
    \includegraphics[width=\textwidth]{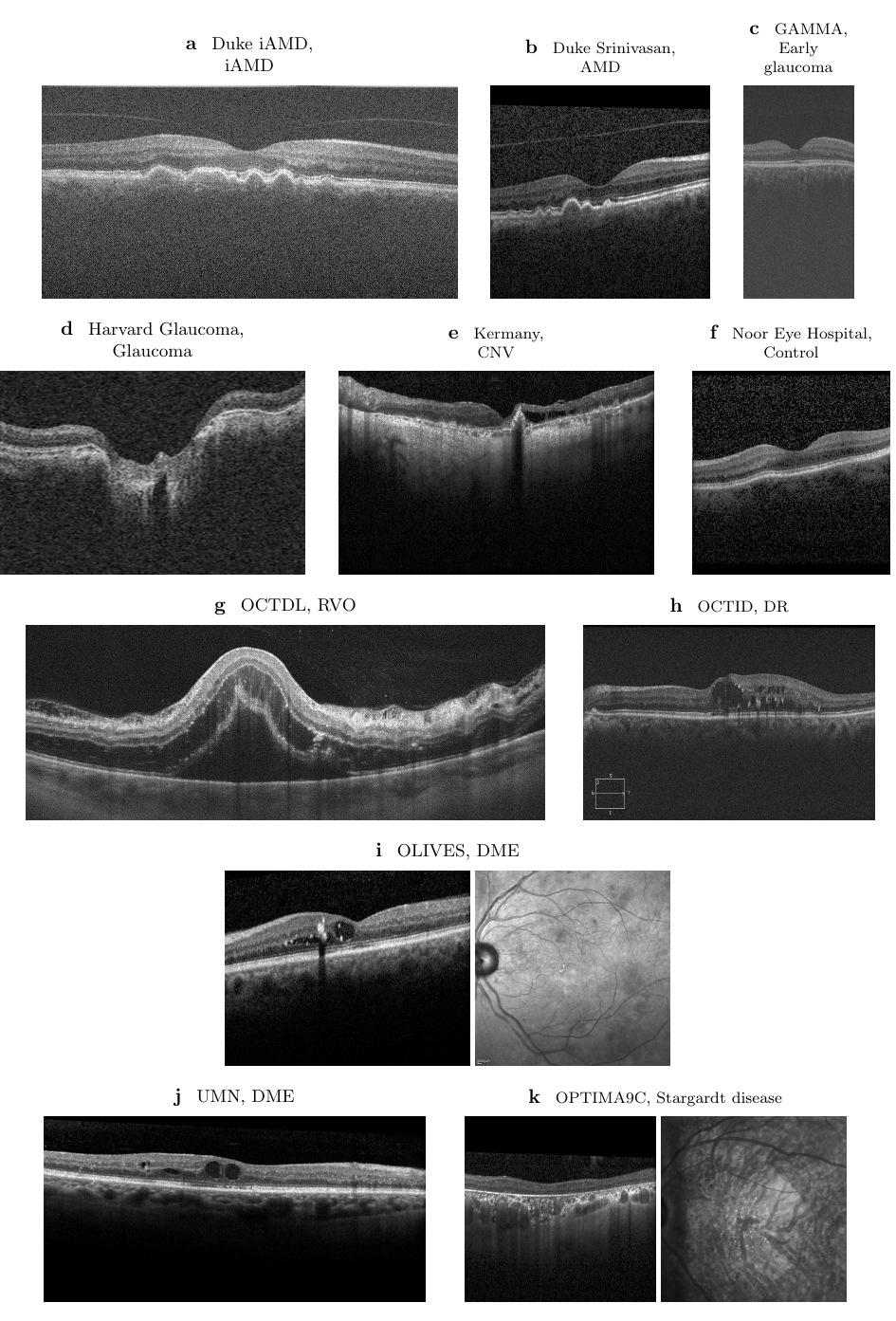}
    \caption{
        \textbf{Example images from classification datasets.}
        OCT and/or SLO images are shown for each dataset along with the corresponding ground truth labels.
    }%
    \label{sup:fig:datasets}
\end{figure}

\paragraph{Duke iAMD~\cite{farsiu2014quantitative}.}
The Duke intermediate age-related macular degeneration (Duke iAMD) dataset includes 38,400 SD-OCT B-scans from 269 iAMD patients and 115 healthy subjects centered on a 5 mm diameter at the fovea.
The images were acquired using a Bioptigen SD-OCT system (Research Triangle Park, NC) at four different clinics in the USA.
Each volume has a retinal FOV of $\sim 6.7 \times 6.7~\text{mm}^2$, and consists of 100 B-scans with dimensions of $512 \times 1000$ pixels.
In addition to the diagnosis, the dataset includes expert-annotated segmentation maps for three key retinal regions: the inner limiting membrane (ILM) to the inner retinal pigment epithelium (RPE) detachment complex (RPEDC), the inner RPEDC to the outer Bruch's membrane (BM), and below the BM.
For the classification evaluation, we focused on the detection of iAMD, while for the segmentation evaluation, we focused on the segmentation of all the aforementioned layers.
Link:~\url{https://people.duke.edu/~sf59/RPEDC_Ophth_2013_dataset.htm}

\paragraph{Duke Srinivasan~\cite{srinivasan2014fully}.}
This dataset presented by Srinivasan et al.~\cite{srinivasan2014fully} consists of OCT volumes acquired from 45 patients.
The scans were categorized according to the diagnosis of the patients: healthy (15 samples), dry AMD (15), and diabetic macular edema (DME) (15).
The scans were acquired using a Spectralis device (Heidelberg Engineering, Heidelberg, Germany) at Duke University, Harvard University, and the University of Michigan, USA.
The FOV and OCT dimensions are very heterogeneous;
for more information, see Table 1 from the original publication~\cite{srinivasan2014fully}.
For the purposes of this study, we focused on the detection of AMD and DME.
Link:~\url{https://people.duke.edu/~sf59/Srinivasan_BOE_2014_dataset.htm}

\paragraph{GAMMA~\cite{wu2023gamma}.}
This dataset originates from the Glaucoma grAding from Multi-Modality imAges (GAMMA) challenge~\cite{wu2023gamma}, held in conjunction with the International Conference on Medical Image Computing and Computer-Assisted Intervention (MICCAI) 2023.
It was provided by the Sun Yat-sen Ophthalmic Center, China, and contains pairs of color fundus and OCT images from randomly selected subjects with and without glaucoma.
In particular, the public dataset includes 100 samples from the same number of patients.
OCT volumes were acquired using a Topcon DRI OCT Triton device (Topcon, Tokyo, Japan).
All volumes are centered on the macula with a FOV of $3 \times 3$ mm.
Each volume has 25 B-scans with dimensions of $992 \times 512 \times$ pixels.
The dataset also includes glaucoma stages, foveal coordinates, and cup and optic disc segmentation masks.
There are 26 samples with early glaucoma, 24 with moderate or advanced glaucoma, and 50 normal samples.
In this study, we focus on the detection and staging of glaucoma.
Link: \url{https://gamma.grand-challenge.org/)}

\paragraph{Harvard Glaucoma~\cite{luo2023harvard}.}
This dataset from the Mass Eye and Ear of Harvard Medical School, USA, consists of 1\,000 samples from 1\,000 patients.
Each sample includes the diagnostic label indicating the presence or absence of glaucoma (557 and 443, respectively), an OCT volume centered on the optic nerve head, a retinal nerve fiber layer thickness (RNFLT) map, and other demographic and clinical information such as visual field test results and patient age.
The OCT volumes and RNFLT maps were acquired using a Cirrus (Carl Zeiss Meditec, Jena, Germany) device.
Each volume has a FOV of $6 \times 6 \text{mm}^2$ and 200 B-scans of $200 \times 300$ pixels.
Link: \url{https://github.com/Harvard-Ophthalmology-AI-Lab/Harvard-GDP}

\paragraph{Kermany~\cite{kermany2018dataset,kermany2018identifying}.}
This dataset, presented by Kermany et al.~\cite{kermany2018dataset,kermany2018identifying}, consists of 109\,309 OCT and 5\,862 chest X-ray images.
The OCT images, from 4\,686 patients, were acquired using an Spectralis device at different clinics in California, USA, Shanghai, China, and Beijing, China.
Each sample was then assigned to one of the following four classes: choroidal neovascularization (CNV) (37\,455 samples), DME (11\,598), drusen (8\,866), and control (51\,390).
The FOVs and dimensions of the OCT scans vary considerably across the dataset~\cite{kermany2018dataset}.
Link: \url{https://data.mendeley.com/datasets/rscbjbr9sj/2}

\paragraph{Noor Eye Hospital~\cite{rasti2017macular}.}
This dataset from the Noor Eye Hospital in Tehran, Iran, consists of 148 OCT volumes from the same number of patients.
Every volume is categorized into one of three classes, depending on the diagnosis: control (50 samples), dry AMD (48), and DME (50).
Scans are centered on the macula and were acquired using a Spectralis device with different FOVs, resolutions, and number of B-scans.
Link:~\url{https://hrabbani.site123.me/available-datasets/dataset-for-oct-classification-50-normal-48-amd-50-dme}

\paragraph{OCTDL~\cite{kulyabin2024octdl}.}
The Optical Coherence Tomography Dataset for Image-Based Deep Learning Methods (OCTDL) is a public dataset from the Ural Federal University Named afer the first president of Russia B. N. Yeltsin, Russia.
The dataset consists of 2\,064 B-scans from 821 patients labeled based on disease type and retinal pathology.
Scans were acquired using an Optovue Avanti RTVue XR (Fremont, CA, USA) device centered on the fovea with different field of views (FOVs) and image resolutions.
Specialists classified the images into one of the following classes: control (332 samples), AMD (1\,231), DME (147), epiretinal membrane (ERM) (155), retinal artery occlusion (RAO) (22), retinal vein occlusion (RVO) (101), and vitreomacular interface disease (VID) (76).
Link:~\url{https://data.mendeley.com/datasets/sncdhf53xc/4}

\paragraph{OCTID~\cite{gholami2020octid}.}
The Optical Coherence Tomography Image Database (OCTID) is a public dataset of OCT B-scans from the Sankara Nethralaya (SN) Eye Hospital,
India.
The images were acquired using a Cirrus HD-OCT machine
with a raster scan protocol with a 2 mm scan length and an image resolution of $512 \times 1024$ pixels.
Images were labeled based on the diagnosis of retinal clinical experts at the SN hospital.
Specifically, the selected 572 B-scans were categorized into the following classes: control (206 samples), macular hole (MH) (102), AMD (55), central serous retinopathy (CSR) (102), and diabetic retinopathy (DR) (107).
In each volumetric scan, a fovea-centered image was selected by an experienced clinical optometrist.
The images were then resized to 500x750 pixels.
Link:~\url{https://borealisdata.ca/dataverse/OCTID}

\paragraph{OLIVES~\cite{prabhushankar2022olives}.}
The Ophthalmic Labels for Investigating Visual Eye Semantics (OLIVES) is a longitudinal dataset consisting of 78\,189 OCT B-scans (1590 volumes) and scanning laser ophthalmoscopy (SLO) (referred to as near-infrared fundus images) from 96 patients with DR or DME.
The OLIVES dataset is derived from the PRIME~\cite{yu2021prime} and TREX-DME~\cite{payne2017trexdme} clinical studies
run at the Retina Consultants of Texas, USA.
Every volume has around 49 B-scans, with an image resolution of $496 \times 504$ pixels.
It includes 16 biomarker labels, 4 clinical labels, and a disease diagnosis for either DME (931 samples) or DR (659 samples).
The images were acquired using Spectralis devices.
Link:~\url{https://zenodo.org/records/7105232}

\paragraph{UMN~\cite{rashno2018fully}.}
The University of Minnesota (UMN) dataset, collected by the University of Minnesota Ophthalmology Clinic, USA, consists of 54 OCT volumes from 30 patients with DME and 24 patients with AMD.
All scans were acquired using a Spectralis device centered on the macula.
Each volume contains 25 B-scans with an image resolution of $496 \times 1\,024$ pixels.
Link:~\url{https://people.ece.umn.edu/users/parhi/.DATA/}

\paragraph{OPTIMA9C~\cite{oghbaie2024vlfatrollout}.}
This private dataset comes from the imaging data collection at OPTIMA Lab, Medical University of Vienna, Austria.
The dataset contains 4\,205 treatment-naive OCT volumes and their corresponding SLO images from 3\,652 patients.
The OCT volumes are baseline scans from randomized, multicenter clinical trials and include data from several manufacturers:
Spectralis, Cirrus, and Triton.
The dataset consists of nine classes based on clinical study data: iAMD (clinical studies NCT01790802 and NCT00891735, Observational study 1 \cite{schlanitz2017drusen}); three types of CNV (NCT02307682, NCT01780935, NCT01972789); geographic atrophy (GA) (NCT02503332, Observational study 2 \cite{bui2022fundus}); RVO (NCT01599650, NCT01535261); DME (NCT01331681, NCT01627249); Stargardt disease (Observational study 3 \cite{ritter2024deep}); and healthy samples
(NCT03465124 and the fellow eyes from the clinical studies NCT00891735, NCT01780935, NCT01948830, NCT01599650, and NCT01535261).
Thus, the classes and the number of samples per class are as follows:
control (183), RVO (763), Stargardt (130),
DME (1 091), iAMD (1 128), GA (452),
CNV1 (99), CNV2 (83), and CNV3 (276).
OPTIMA9C is notable for its variation in spatial resolution and disease severity, making it more challenging and unique than other datasets.
Image widths range from 200 to 1536 pixels, with heights varying from 480 to 1024 pixels, while the number of slices per volume ranges from 25 to 261 slices, with an average of 81.
For privacy reasons, the dataset cannot be made publicly available.

\subsection{Segmentation datasets}

The segmentation datasets used in the evaluation benchmark are briefly described below.
In contrast to the classification datasets, all the B-scans in the segmentation datasets are used for training and evaluating the models.
This was done following standard practices in the literature, where layer and lesion segmentation tasks are usually performed B-scan-wise, and not on the full volumes~\cite{roy2017relaynet,he2021structured,fazekas2023segmentation}.
Examples of images from each dataset and the corresponding segmentation labels are shown in Supplementary Figure~\ref{sup:fig:datasets_seg}.

\begin{figure}[tbph]
    \centering
    \includegraphics[width=\textwidth]{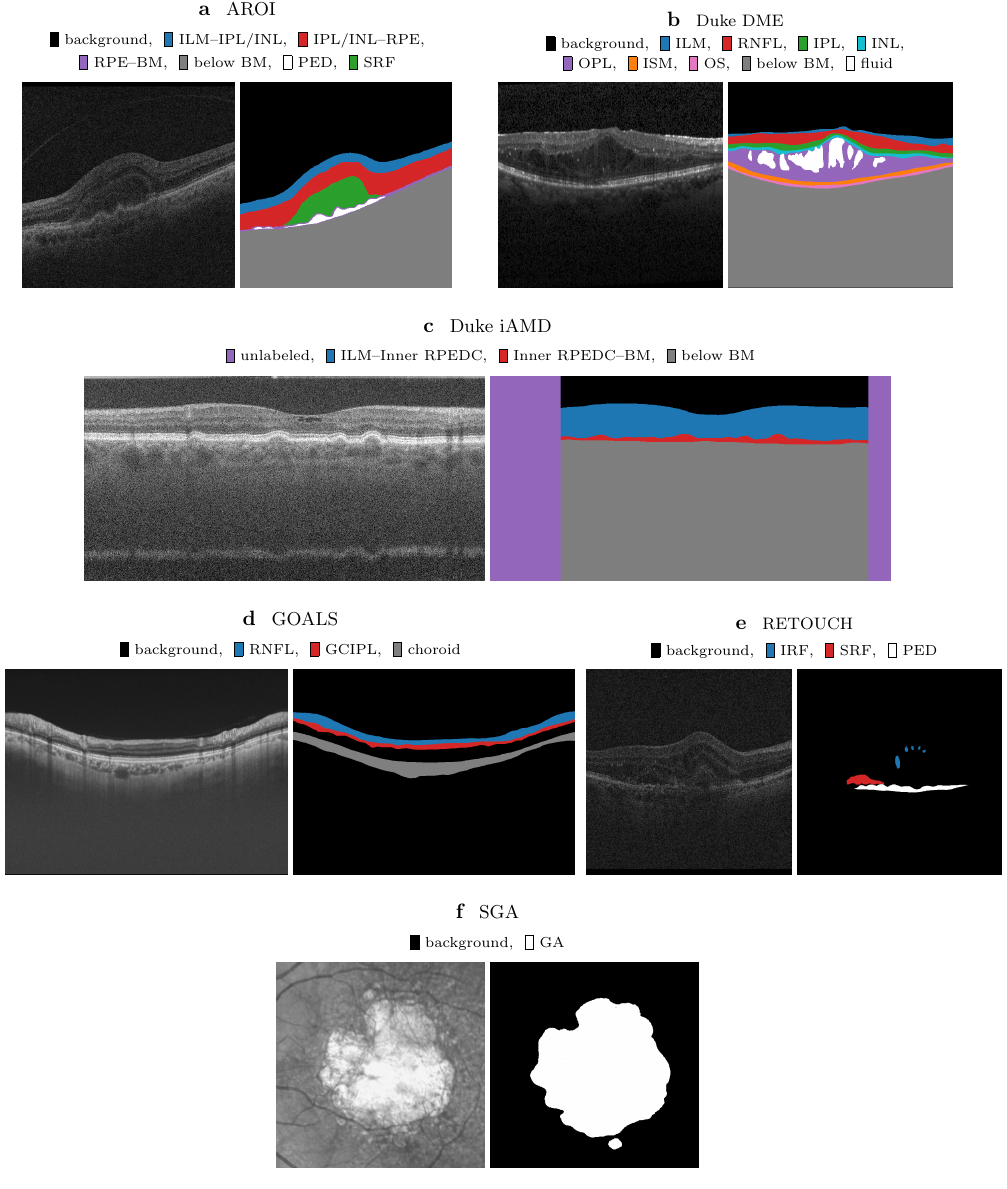}
    \caption{\textbf{Example images from segmentation datasets.}
    OCT and/or SLO images are shown for each dataset along with their corresponding segmentation masks.
    }%
    \label{sup:fig:datasets_seg}
\end{figure}

\paragraph{AROI~\cite{melinscak2021aroi}.}
The Annotated Retinal Optical coherence tomography Images (AROI) dataset comprises 3\,200 B-scans collected from 25 patients diagnosed with neovascular AMD.
Macular SD-OCT volumes were acquired with the Zeiss Cirrus HD OCT 4000 device at the Sestre milosrdnice University Hospital Center, Croatia.
Each OCT volume consisted of 128 B-scans with a resolution of $1024 \times 512$ pixels (pixel size $1.96 \times 11.74 \mu\text{m}$).
Retinal fluids and layers were annotated for {1\,136} B-scans.
In our study, one of the OCT volumes was discarded (\texttt{patient14}) due to inconsistencies in the annotations, resulting in 3\,072 B-scans from 24 patients.
Of the fluids, the following were annotated: pigment epithelium detachment (PED), subretinal fluid and subretinal hyperreflective material (both labeled as SRF), and intraretinal fluid (IRF) (named \textit{cyst} in the dataset).
Four layer boundaries were also annotated: internal limiting membrane (ILM), inner plexiform layer/inner nuclear layer (IPL/INL), retinal pigment epithelium (RPE), and Bruch's membrane (BM).
Translating the annotations from boundaries to layers results in the following labels: ILM--IPL/INL, IPL/INL--RPE, RPE--BM, below the BM.
Link:~\url{https://ipg.fer.hr/ipg/resources/oct_image_database}

\paragraph{Duke DME~\cite{chiu2015kernel}.}
The Duke Diabetic Macular Edema (Duke DME) dataset, provided by Duke University, USA, consists of 110 annotated OCT B-scan images from 10 patients with severe DME.
Each patient features 11 B-scans centered on the fovea with dimensions of $496 \times 536$ pixels.
Scans were acquired using a Spectralis device.
The dataset includes detailed annotations of eight retinal layer boundaries: ILM, retinal nerve fiber layer (RNFL), ganglion cell layer (GCL), IPL, INL, outer plexiform layer (OPL), outer nuclear layer (ONL), inner segments of photoreceptors (ISM), outer segments of photoreceptors (OS), and RPE.
It also includes annotations for fluid regions.
Link:~\url{https://people.duke.edu/~sf59/Chiu_BOE_2014_dataset.htm}

\paragraph{Duke iAMD~\cite{farsiu2014quantitative}.}
See the description in the Classification Datasets section above.

\paragraph{GOALS~\cite{fang2022dataset}.}
The Glaucoma OCT Analysis and Layer Segmentation (GOALS) dataset is part of the GOALS Challenge, held in conjunction with MICCAI 2022.
The full dataset comprises 300 circumpapillary OCT B-scans randomly selected from glaucoma study cohorts collected
at the Zhongshan Ophthalmic Center,
China.
However, only the 100 B-scans from the training subset were publicly available at the time of preparing the evaluation benchmark, so we used this subset in our study.
The number of patients is unknown, as no information is provided in the dataset description and no anonymized patient identifiers are included in the data.
FOV information is not provided either.
All scans were acquired using a Topcon DRI Swept Source OCT system.
All B-scans have a dimension of $800 \times 1100$ pixels.
Each B-scan is annotated with the following three layers: RNFL, GCL--IPL (GCIPL), and choroid.
Since the dataset is not publicly available at the time of writing, we will upload it to our repository: \url{https://github.com/j-morano/MIRAGE}.

\paragraph{RETOUCH~\cite{bogunovic2019retouch}.}
The retinal OCT fluid detection and segmentation benchmark and challenge (RETOUCH) dataset originates from the MICCAI 2017 retinal OCT fluid challenge.
Annotations and scans came from different clinical centers: Medical University of Vienna, Austria, Erasmus University Medical Center, Netherlands, and Radboud University Medical Center, Netherlands.
The dataset includes 70 OCT volumes, with half of the patients diagnosed with macular edema secondary to AMD and the other half with edema secondary to RVO.
Each B-scan is labeled with three retinal fluid types: IRF, subretinal fluid (SRF), and PED.
The training data consists of volumes from three OCT systems: 24 from Cirrus (Model 5000), 22 from Triton (Models T-1000/T-2000), and 24 from Spectralis.
Cirrus scans consist of 128 B-scans of $1024 \times 512$ pixels, Triton scans consist of 128 B-scans $512 \times 650$ or $512 \times 885$ pixels, and Spectralis scans consist of 49 B-scans $512 \times 496$ pixels.
There is at least one fluid lesion in each volume.
For this dataset, the results are obtained using the official evaluation script provided by the challenge organizers.
Link:~\url{https://retouch.grand-challenge.org/}

\paragraph{SGA~\cite{bui2022fundus}.}
This dataset for the segmentation of geographic atrophy (SGA) is a private dataset consisting of 965 samples consisting of OCT volumes as well as SLO and fundus autofluorescence (FAF) images.
All samples come from 100 patients (184 eyes) diagnosed with GA who were part of a clinical study on natural GA progression conducted at the Medical University of Vienna, Austria~\cite{bui2022fundus}.
The SLO and FAF images were acquired with a Spectralis device centered on the macula with a FOV of $6 \times 6~\text{mm}^2$ and a resolution of $1024 \times 1024$ pixels.
OCT and SLO images were co-registered by the device, while FAF and SLO images were registered with an in-house image registration pipeline based on aligning retinal vessel segmentation~\cite{arikan2019deep}.
All samples have GA \textit{en face} masks annotated by a retinal expert on FAF images.
In this work, we used only the SLO images and the corresponding GA masks.
For privacy reasons, the dataset cannot be made publicly available.

\clearpage
\section{Supplementary Note 3: Masking strategy}%
\label{sup:sec:masking}

During pretraining, following MultiMAE~\cite{bachmann2022multimae}, we sample the number of non-masked tokens for each modality from a symmetric Dirichlet distribution with a concentration parameter $\alpha=1$.
This value results in a diverse sampling across the different modalities.
Below, we provide an analysis of the effect of different values of $\alpha$ on the token distribution across three modalities: OCT, SLO, and Layers.

In Supplementary Figure~\ref{sup:fig:masking}, we show a 3D visualization of the number of non-masked tokens per modality for 2000 samples when using different values of $\alpha$.
In the plot, each point represents a sample, and its coordinates correspond to the number of non-masked tokens for each modality.
The number of non-masked tokens per modality out of the fixed total (98) was sampled from a Dirichlet distribution with the specified $\alpha$ value.

When $\alpha$ is set to 0.1, most samples are concentrated near the edges of the simplex, meaning that in most cases nearly all tokens are from a single modality.
This leads to frequent unimodal processing, potentially causing the model to ignore multimodal interactions.

For $\alpha$ = 1, samples are spread across the entire simplex, indicating that the model is more likely to receive tokens from multiple modalities.
However, as shown in the plot, it is still possible for the model to receive tokens from only one or two modalities.
In this way, the model can learn to process each modality independently while still leveraging multimodal information.

At $\alpha$ = 100, almost all samples are tightly clustered around the center of the simplex, indicating that the number of tokens per modality is roughly equal ($\sim$32 tokens per modality).
While this configuration ensures balanced multimodal learning, it may limit the ability of the model to process samples where only one modality is available.

To ensure that the model can effectively leverage the multimodal information while still being able to process each modality independently, we set $\alpha$ to 1, which provides a good balance between modality specialization and multimodal learning.

In our training approach, once the \textit{number} of non-masked tokens per modality has been sampled, the non-masked tokens are sampled uniformly at random without replacement and used as input to the model.

\begin{figure}[h]
    \centering
    \includegraphics[width=\textwidth]{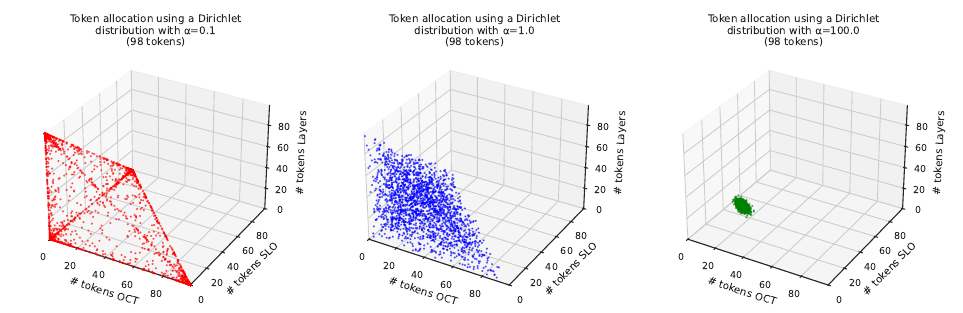}
    \caption{%
        \textbf{3D scatter plots of token allocations across OCT, SLO, and Layers for different Dirichlet concentration parameters ($\alpha$ = 0.1, 1, and 100).}
        Each point represents a sample, with the coordinates corresponding to the number of non-masked tokens for each modality.
        A total of 2000 samples are shown for each $\alpha$ value, with the number of total tokens fixed at 98.
        Lower $\alpha$ values (left) result in sparse, modality-dominant allocations, while higher $\alpha$ values (right) enforce more uniform token distributions.
    }%
    \label{sup:fig:masking}
\end{figure}

\clearpage
\section{Supplementary Note 4: Pre-experimental results}\label{sup:sec:pre_results}

To comprehensively assess the effectiveness of the proposed multimodal pretraining approach, we conducted a series of analyses evaluating the quality of the learned representations of MIRAGE without fine-tuning the model on downstream tasks.
In particular, we performed three different analyses: training loss, reconstruction visualization, and feature visualization.

\paragraph{Loss curves.}
To analyze the convergence behavior of the models during pretraining, we show in Supplementary Figure~\ref{sup:fig:loss_curves} the loss curves for the three training losses (one for each modality) of MIRAGE during the pretraining stage, as well as the total loss and the learning rate.
The curves show that the model converges well to values close to zero for all losses, indicating that the model is able to reconstruct the input patches effectively.

\begin{figure}[h]
    \centering
    \includegraphics[width=0.95\linewidth]{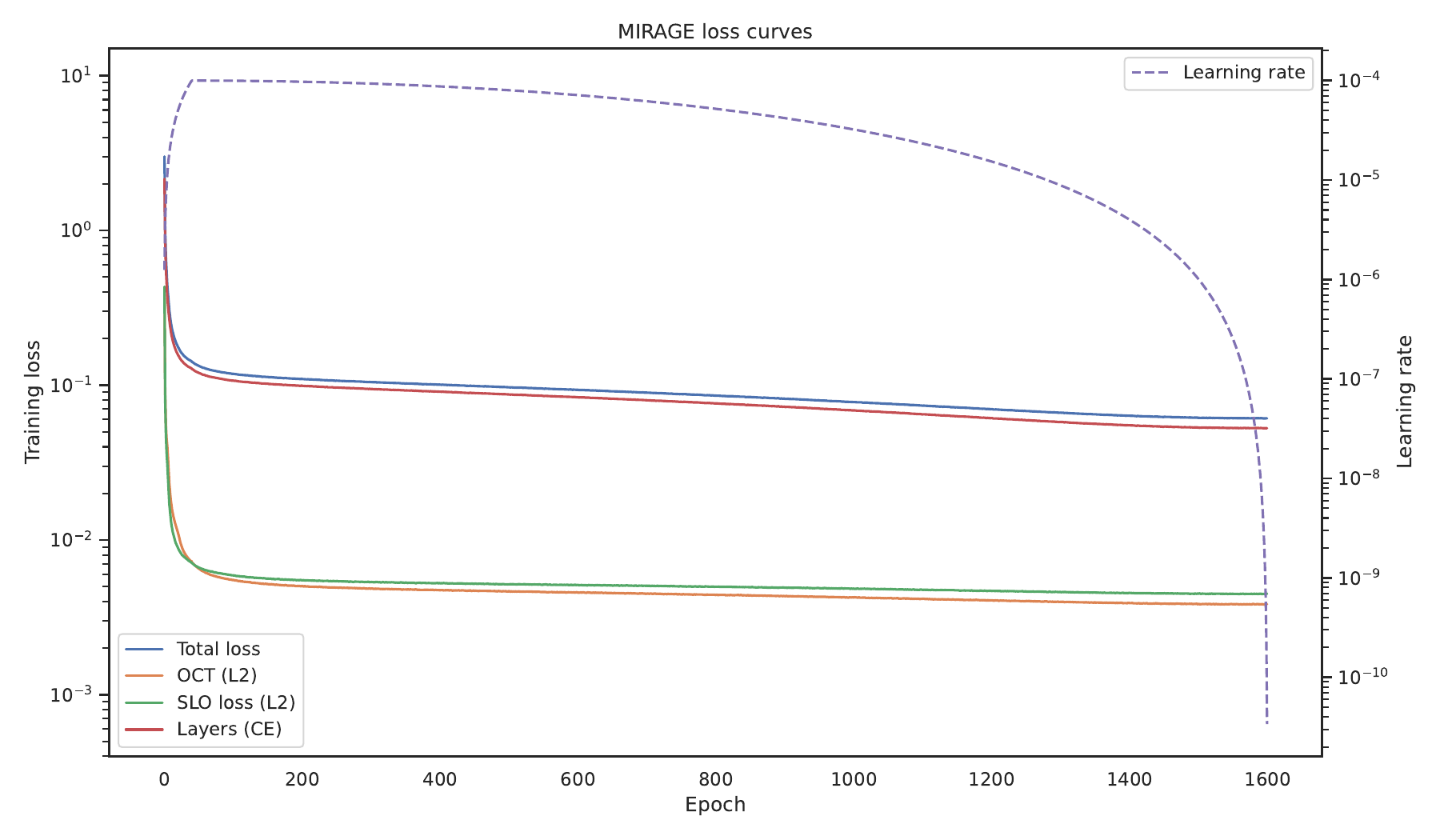}
    \caption{%
        \textbf{Loss curves of MIARGE during pretraining.}
        All training losses for the three modalities (OCT, SLO, and Layers) are significantly minimized during the pretraining stage.
    }%
    \label{sup:fig:loss_curves}
\end{figure}

\paragraph{Qualitative reconstruction results.}
To further validate the effectiveness of our paired multimodal pretraining strategy, we show in Supplementary Figure~\ref{sup:fig:reconstruction} the predictions of MIRAGE on our internal dataset for different masking ratios of the input modalities.
In the figure, it can be observed that the model is able to reconstruct the input patches from the different modalities with relatively high fidelity, even when a large portion of the input modality is masked.
For instance, we show that the model is able to predict the segmented layers from the OCT B-scans and the SLO, even when all patches from the input Layers modality are masked.
This is also true in the reverse direction, where the model is able to reconstruct the OCT B-scans from the segmented layers and SLO images.
Additionally, although the SLO reconstructions are of lower quality compared to OCT and Layers, the model is still able to reconstruct the coarse lesions from the OCT B-scans and the segmented layers.
These visualizations suggest that the model effectively learns the relationships between the different modalities during the pretraining phase.

\begin{figure}[tbph]
    \centering
    \includegraphics[width=\textwidth]{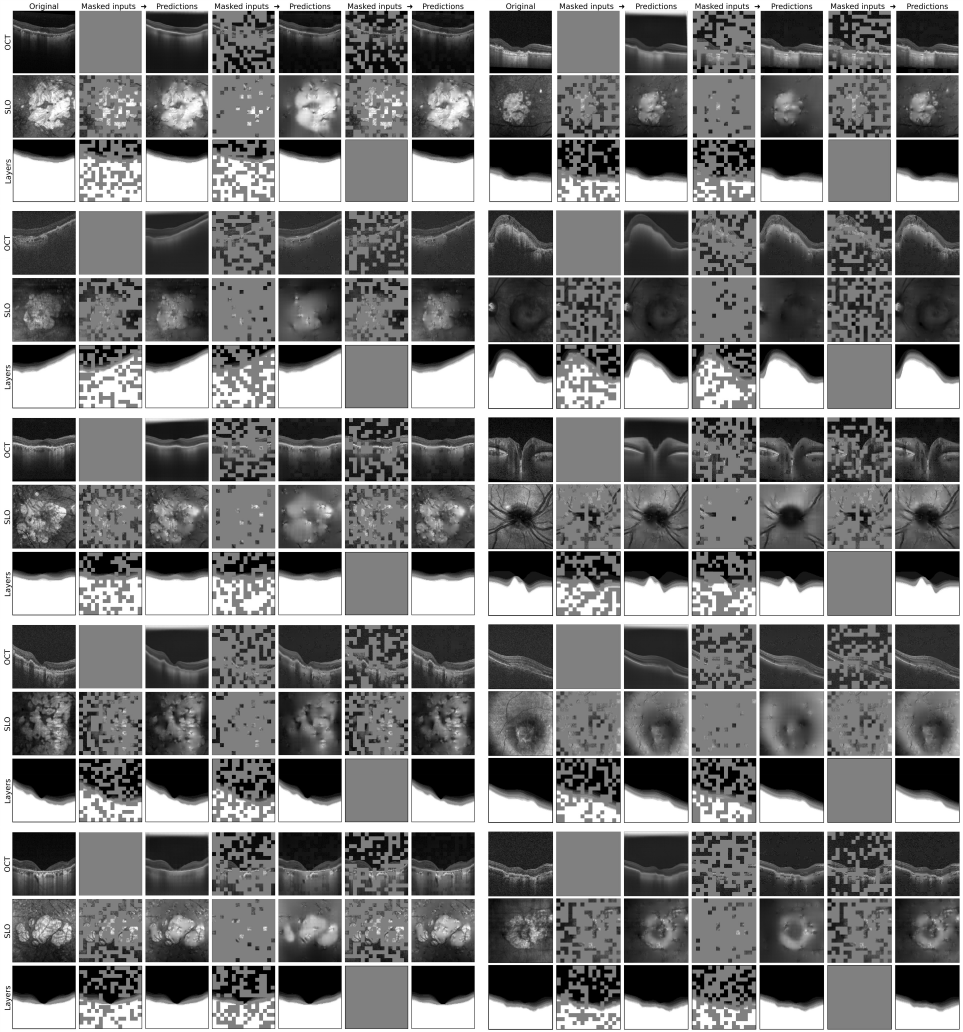}
    \caption{%
        \textbf{MIRAGE predicitons on our internal dataset.}
        For each sample (composed of an OCT B-scan, an SLO image, and a layer segmentation) we show the predictions with different masking proportions for each modality.
        The visualizations show that the model is able to reconstruct masked patches from the different modalities, suggesting that the learned representations encode meaningful multimodal information.
    }%
    \label{sup:fig:reconstruction}
\end{figure}

\paragraph{Feature visualization.}
To provide a qualitative understanding of the representations learned by MIRAGE, we visualize the embeddings produced by it, DINOv2, and RETFound for the OCTID and Kermany datasets.
Both models were used out-of-the-box without any fine-tuning.
To project the embeddings into a 2D space, we used the UMAP algorithm~\cite{mcinnes2018umap}.
The results are shown in Supplementary Figure~\ref{fig:sup:umap_vis}.
As shown in the figure, the embeddings produced by MIRAGE are well separated and roughly cluster the samples according to their classes.
While the embeddings produced by DINOv2 and RETFound also show good separation for OCTID, they are less distinct for the Kermany dataset.
This indicates that MIRAGE has learned meaningful data representations during the pretraining stage.

\begin{figure}[b!]
    \centering
    \includegraphics[width=0.95\textwidth]{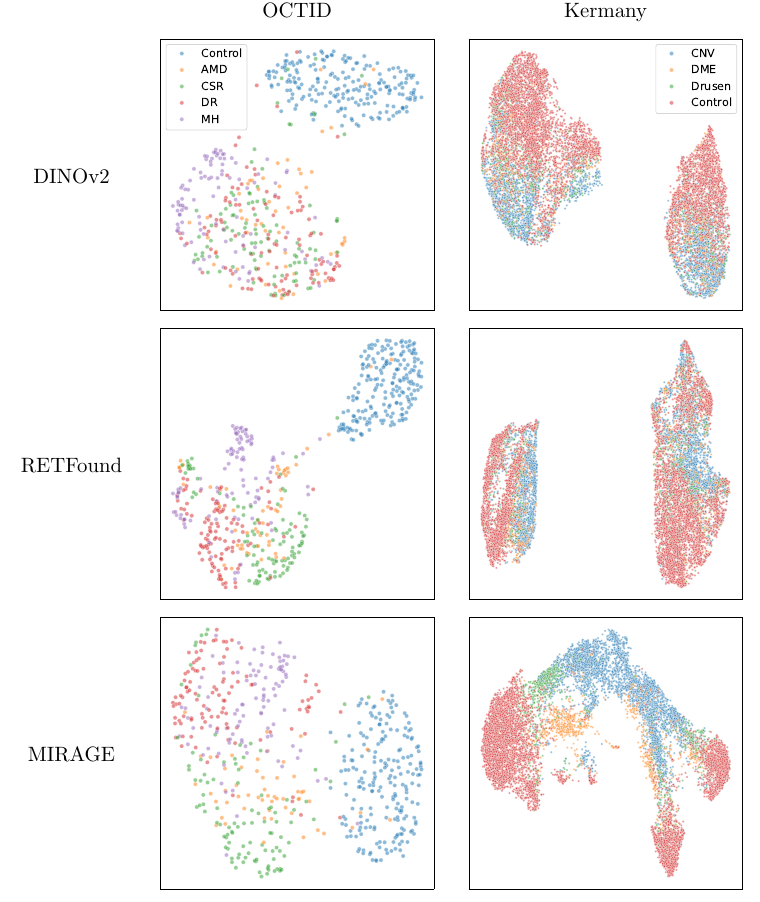}
    \caption{\textbf{Visualization of feature embeddings from different foundation models.}
    These plots show the embeddings produced by DINOv2, RETFound, and MIRAGE for the OCTID and Kermany datasets after projection to 2D using UMAP.
    MIRAGE embeddings show better separation and clustering of samples compared to DINOv2 and RETFound, suggesting better representation learning.
    }%
    \label{fig:sup:umap_vis}
\end{figure}

\clearpage

\section{Supplementary Note 5: Computational efficiency}\label{sup:sec:efficiency}

To evaluate the computational efficiency of the proposed MIRAGE model, we provide, in Supplementary Table~\ref{sup:tab:efficiency}, the number of parameters, FLOPs, and Mult-Adds operations for the different configurations of the model, i.e., for pretraining (with three modality-specific linear projection layers [LPL] and Transformer-based decoders), for classification (with only one LPL and a linear probing head) and for segmentation (with one LPL and a ConvNeXt-based segmentation head).
Both classification and segmentation configurations are common to all the adapted foundation models, which also share the same encoder architecture.
The only difference is the encoder size.
Thus, for RETFound and DINOv2, that use a ViT-Large encoder, the Large configuration is used, and for MedSAM, which uses ViT-Base, the Base.

In addition, we provide, in Supplementary Table~\ref{sup:tab:resources}, the computational resources and amount of data required to pretrain MIRAGE compared to other foundation models.
In particular, we show the training time, number of epochs, batch size, samples and learning approach used for pretraining along with the hardware.
All models except MedSAM and MultiMAE, which use ViT-Base, use ViT-Large encoders.
As shown in the table, MIRAGE requires fewer computational resources and less data compared to other medical foundation models such as RETFound and MedSAM.

\begin{table}[h]
    \centering
    \caption{%
        \textbf{Computational efficiency of models for pretraining, classification, and segmentation tasks.}
        The table shows the number of parameters, FLOPs and Mult-Adds operations of the models.
        While the values for pretraining are only for MIRAGE, the values for classification and segmentation are the same for all adapted foundation models, depending on the task and the size of the encoder.
        For example, RETFound and DINOv2 use a ViT-Large encoder, so the Large configuration is used.
        In contrast, MedSAM uses a ViT-Base encoder, so the Base configuration is used.
        Input size is fixed to $512 \times 512$ for the calculations.
    }%
    \label{sup:tab:efficiency}
    \begin{tabular}{llrrr}
        \toprule
        \textbf{Configuration} & \textbf{Size} & \textbf{\# Parameters} & \textbf{FLOPs} & \textbf{Mult-Adds} \\
        & & \textbf{(M)} & \textbf{(G)} & \textbf{(G)} \\
        \midrule
        Pretraining & Large & 318.23 & 236.85 & 1.92 \\
        & Base & 99.01 & 69.26 & 1.30 \\
        \midrule
        Classification & Large & 309.40 & 155.89 & 1.14 \\
        & Base & 90.37 & 44.11 & 0.57  \\
        \midrule
        Segmentation & Large & 315.52 & 396.08 & 2.71 \\
        & Base & 96.16 & 171.56 & 2.22 \\
        \bottomrule
    \end{tabular}%
\end{table}

\begin{table}[h]
    \centering
    \caption{%
        \textbf{Computational resources and data requirements for pretraining MIRAGE compared to other foundation models.}
        The table shows the training time, the number of epochs, the batch size, the number of samples, the data type and the learning method (fully supervised [FS] or self-supervised [SS]) used for pretraining along with the used hardware.
    }%
    \label{sup:tab:resources}
    \resizebox{\textwidth}{!}{%
    \begin{tabular}{lrrrrrrrl}
        \toprule
        \textbf{Method} & \textbf{Training time} & \textbf{\# Epochs} & \textbf{Batch size} & \textbf{\# Samples} & \textbf{Data} & \textbf{Method} & \textbf{Hardware} \\
        \midrule
        SL-IN & 23 days & 90 & 4096 & 14.2M & RGB natural images, Classes & FS & 10x TPUv3
        \\
        DINOv2 & $>$41 days & - & 2048 & 142M & RGB natural images & SS & 160x V100 (32GB)
        \\
        MultiMAE & - & 1600 & 2048 & 1.28M & RGB, Depth, Seg. masks & SS & 8x A100 (80GB)
        \\
        MedSAM & - & 150 & 160 & 1.57M & 10 modalities, Seg. masks & FS & 20x A100 (80GB)
        \\
        RETFound & 14 days & 800 & 1792 & 736k & OCT & SS & 8x A100 (40GB)
        \\
        MIRAGE & 25 days & 1600 & 256 & 261k & OCT, SLO, Layers & SS & 1x A100 (80GB)
        \\
        \bottomrule
    \end{tabular}%
    }
\end{table}

\clearpage
\bibliographystyle{naturemag}
\bibliography{references}

\begin{thebibliography}{100}
\expandafter\ifx\csname url\endcsname\relax
  \def\url#1{\texttt{#1}}\fi
\expandafter\ifx\csname urlprefix\endcsname\relax\def\urlprefix{URL }\fi
\providecommand{\bibinfo}[2]{#2}
\providecommand{\eprint}[2][]{\url{#2}}

\bibitem{burton2021lancet}
\bibinfo{author}{Burton, M.~J.} \emph{et~al.}
\newblock \bibinfo{title}{The lancet global health commission on global eye
  health: vision beyond 2020}.
\newblock \emph{\bibinfo{journal}{The Lancet Global Health}}
  \textbf{\bibinfo{volume}{9}}, \bibinfo{pages}{e489--e551}
  (\bibinfo{year}{2021}).

\bibitem{kanski2011clinical}
\bibinfo{author}{Kanski, J.~J.} \& \bibinfo{author}{Bowling, B.}
\newblock \emph{\bibinfo{title}{Clinical Ophthalmology: A Systematic Approach}}
  (\bibinfo{publisher}{Elsevier Health Sciences}, \bibinfo{year}{2011}),
  \bibinfo{edition}{seventh} edn.

\bibitem{mohammadpour2020diagnostics}
\bibinfo{author}{Mohammadpour, M.}
\newblock \emph{\bibinfo{title}{Diagnostics in Ocular Imaging: Cornea, Retina,
  Glaucoma and Orbit}} (\bibinfo{publisher}{Springer Nature},
  \bibinfo{year}{2020}).

\bibitem{SE2016prer}
\bibinfo{author}{Schmidt-Erfurth, U.} \& \bibinfo{author}{Waldstein, S.~M.}
\newblock \bibinfo{title}{A paradigm shift in imaging biomarkers in neovascular
  age-related macular degeneration}.
\newblock \emph{\bibinfo{journal}{Progress in Retinal and Eye Research}}
  \textbf{\bibinfo{volume}{50}}, \bibinfo{pages}{1--24} (\bibinfo{year}{2016}).

\bibitem{schuman2024optical}
\bibinfo{author}{Schuman, J.~S.}, \bibinfo{author}{Fujimoto, J.~G.},
  \bibinfo{author}{Duker, J.} \& \bibinfo{author}{Ishikawa, H.}
\newblock \emph{\bibinfo{title}{Optical coherence tomography of ocular
  diseases}} (\bibinfo{publisher}{CRC Press}, \bibinfo{year}{2024}).

\bibitem{bemme2021reliability}
\bibinfo{author}{Bemme, S.} \emph{et~al.}
\newblock \bibinfo{title}{Reliability of subjective assessment of
  spectral-domain {OCT} pathologic features by multiple raters in retinal vein
  occlusion}.
\newblock \emph{\bibinfo{journal}{Ophthalmology Science}}
  \textbf{\bibinfo{volume}{1}}, \bibinfo{pages}{100031} (\bibinfo{year}{2021}).

\bibitem{schmitz2023interreader}
\bibinfo{author}{Schmitz-Valckenberg, S.} \emph{et~al.}
\newblock \bibinfo{title}{Interreader agreement and longitudinal progression of
  incomplete/complete retinal pigment epithelium and outer retinal atrophy in
  age-related macular degeneration}.
\newblock \emph{\bibinfo{journal}{Ophthalmology Retina}}
  \textbf{\bibinfo{volume}{7}}, \bibinfo{pages}{1059--1068}
  (\bibinfo{year}{2023}).

\bibitem{defauw2018clinically}
\bibinfo{author}{De~Fauw, J.} \emph{et~al.}
\newblock \bibinfo{title}{Clinically applicable deep learning for diagnosis and
  referral in retinal disease}.
\newblock \emph{\bibinfo{journal}{Nature Medicine}}
  \textbf{\bibinfo{volume}{24}}, \bibinfo{pages}{1342–1350}
  (\bibinfo{year}{2018}).

\bibitem{schmidt2018artificial}
\bibinfo{author}{Schmidt-Erfurth, U.}, \bibinfo{author}{Sadeghipour, A.},
  \bibinfo{author}{Gerendas, B.~S.}, \bibinfo{author}{Waldstein, S.~M.} \&
  \bibinfo{author}{Bogunovi{\'c}, H.}
\newblock \bibinfo{title}{Artificial intelligence in retina}.
\newblock \emph{\bibinfo{journal}{Progress in retinal and eye research}}
  \textbf{\bibinfo{volume}{67}}, \bibinfo{pages}{1--29} (\bibinfo{year}{2018}).

\bibitem{Yim2020}
\bibinfo{author}{Yim, J.} \emph{et~al.}
\newblock \bibinfo{title}{Predicting conversion to wet age-related macular
  degeneration using deep learning}.
\newblock \emph{\bibinfo{journal}{Nature Medicine}} \bibinfo{pages}{1--8}
  (\bibinfo{year}{2020}).

\bibitem{Ting2019}
\bibinfo{author}{Ting, D. S.~W.} \emph{et~al.}
\newblock \bibinfo{title}{{Artificial intelligence and deep learning in
  ophthalmology}}.
\newblock \emph{\bibinfo{journal}{British Journal of Ophthalmology}}
  \textbf{\bibinfo{volume}{103}}, \bibinfo{pages}{167--175}
  (\bibinfo{year}{2019}).

\bibitem{schmidt2022aibased}
\bibinfo{author}{Schmidt-Erfurth, U.} \emph{et~al.}
\newblock \bibinfo{title}{{AI}-based monitoring of retinal fluid in disease
  activity and under therapy}.
\newblock \emph{\bibinfo{journal}{Progress in Retinal and Eye Research}}
  \textbf{\bibinfo{volume}{86}}, \bibinfo{pages}{100972}
  (\bibinfo{year}{2022}).

\bibitem{reiter2024ai}
\bibinfo{author}{Reiter, G.~S.} \emph{et~al.}
\newblock \bibinfo{title}{{AI} in the clinical management of {GA}: A novel
  therapeutic universe requires novel tools}.
\newblock \emph{\bibinfo{journal}{Progress in Retinal and Eye Research}}
  \textbf{\bibinfo{volume}{103}}, \bibinfo{pages}{101305}
  (\bibinfo{year}{2024}).

\bibitem{paschali2018generalizability}
\bibinfo{author}{Paschali, M.}, \bibinfo{author}{Conjeti, S.},
  \bibinfo{author}{Navarro, F.} \& \bibinfo{author}{Navab, N.}
\newblock \bibinfo{title}{Generalizability vs. robustness: investigating
  medical imaging networks using adversarial examples}.
\newblock In \emph{\bibinfo{booktitle}{Proceedings of the International
  Conference on Medical Image Computing and Computer-Assisted Intervention}},
  \bibinfo{pages}{493--501} (\bibinfo{organization}{Springer},
  \bibinfo{year}{2018}).

\bibitem{hemelings2023generalizable}
\bibinfo{author}{Hemelings, R.} \emph{et~al.}
\newblock \bibinfo{title}{A generalizable deep learning regression model for
  automated glaucoma screening from fundus images}.
\newblock \emph{\bibinfo{journal}{NPJ digital medicine}}
  \textbf{\bibinfo{volume}{6}}, \bibinfo{pages}{112} (\bibinfo{year}{2023}).

\bibitem{yang2024limits}
\bibinfo{author}{Yang, Y.}, \bibinfo{author}{Zhang, H.},
  \bibinfo{author}{Gichoya, J.~W.}, \bibinfo{author}{Katabi, D.} \&
  \bibinfo{author}{Ghassemi, M.}
\newblock \bibinfo{title}{The limits of fair medical imaging {AI} in real-world
  generalization}.
\newblock \emph{\bibinfo{journal}{Nature Medicine}} \bibinfo{pages}{1--11}
  (\bibinfo{year}{2024}).

\bibitem{ericsson2022self}
\bibinfo{author}{Ericsson, L.}, \bibinfo{author}{Gouk, H.},
  \bibinfo{author}{Loy, C.~C.} \& \bibinfo{author}{Hospedales, T.~M.}
\newblock \bibinfo{title}{Self-supervised representation learning:
  Introduction, advances, and challenges}.
\newblock \emph{\bibinfo{journal}{IEEE Signal Processing Magazine}}
  \textbf{\bibinfo{volume}{39}}, \bibinfo{pages}{42--62}
  (\bibinfo{year}{2022}).

\bibitem{he2022mae}
\bibinfo{author}{He, K.} \emph{et~al.}
\newblock \bibinfo{title}{Masked autoencoders are scalable vision learners}.
\newblock In \emph{\bibinfo{booktitle}{Proceedings of the IEEE/CVF Conference
  on Computer Vision and Pattern Recognition}}, \bibinfo{pages}{15979--15988}
  (\bibinfo{year}{2022}).

\bibitem{le2020contrastive}
\bibinfo{author}{Le-Khac, P.~H.}, \bibinfo{author}{Healy, G.} \&
  \bibinfo{author}{Smeaton, A.~F.}
\newblock \bibinfo{title}{Contrastive representation learning: A framework and
  review}.
\newblock \emph{\bibinfo{journal}{IEEE Access}} \textbf{\bibinfo{volume}{8}},
  \bibinfo{pages}{193907--193934} (\bibinfo{year}{2020}).

\bibitem{radford2021clip}
\bibinfo{author}{Radford, A.} \emph{et~al.}
\newblock \bibinfo{title}{Learning transferable visual models from natural
  language supervision}.
\newblock In \emph{\bibinfo{booktitle}{Proceedings of the International
  Conference on Machine Learning}}, \bibinfo{pages}{8748--8763}
  (\bibinfo{organization}{PMLR}, \bibinfo{year}{2021}).

\bibitem{caron2021emerging}
\bibinfo{author}{Caron, M.} \emph{et~al.}
\newblock \bibinfo{title}{Emerging properties in self-supervised vision
  transformers}.
\newblock In \emph{\bibinfo{booktitle}{Proceedings of the IEEE/CVF
  International Conference on Computer Vision}}, \bibinfo{pages}{9650--9660}
  (\bibinfo{year}{2021}).

\bibitem{girdhar2023imagebind}
\bibinfo{author}{Girdhar, R.} \emph{et~al.}
\newblock \bibinfo{title}{{ImageBind}: One embedding space to bind them all}.
\newblock In \emph{\bibinfo{booktitle}{Proceedings of the IEEE/CVF Conference
  on Computer Vision and Pattern Recognition}}, \bibinfo{pages}{15180--15190}
  (\bibinfo{year}{2023}).

\bibitem{bachmann2022multimae}
\bibinfo{author}{Bachmann, R.}, \bibinfo{author}{Mizrahi, D.},
  \bibinfo{author}{Atanov, A.} \& \bibinfo{author}{Zamir, A.}
\newblock \bibinfo{title}{{MultiMAE}: Multi-modal multi-task masked
  autoencoders}.
\newblock In \emph{\bibinfo{booktitle}{Proceedings of the European Conference
  on Computer Vision}}, \bibinfo{pages}{348--367}
  (\bibinfo{organization}{Springer}, \bibinfo{year}{2022}).

\bibitem{morano2020multimodal}
\bibinfo{author}{Morano, J.}, \bibinfo{author}{Hervella, {\'A}.~S.},
  \bibinfo{author}{Barreira, N.}, \bibinfo{author}{Novo, J.} \&
  \bibinfo{author}{Rouco, J.}
\newblock \bibinfo{title}{Multimodal transfer learning-based approaches for
  retinal vascular segmentation}.
\newblock In \emph{\bibinfo{booktitle}{ECAI 2020}}, \bibinfo{pages}{1866--1873}
  (\bibinfo{publisher}{IOS Press}, \bibinfo{year}{2020}).

\bibitem{li2020self}
\bibinfo{author}{Li, X.}, \bibinfo{author}{Jia, M.}, \bibinfo{author}{Islam,
  M.~T.}, \bibinfo{author}{Yu, L.} \& \bibinfo{author}{Xing, L.}
\newblock \bibinfo{title}{Self-supervised feature learning via exploiting
  multi-modal data for retinal disease diagnosis}.
\newblock \emph{\bibinfo{journal}{IEEE Transactions on Medical Imaging}}
  \textbf{\bibinfo{volume}{39}}, \bibinfo{pages}{4023--4033}
  (\bibinfo{year}{2020}).

\bibitem{morano2023self}
\bibinfo{author}{Morano, J.} \emph{et~al.}
\newblock \bibinfo{title}{Self-supervised learning via inter-modal
  reconstruction and feature projection networks for label-efficient
  {3D-to-2D} segmentation}.
\newblock In \bibinfo{editor}{Greenspan, H.} \emph{et~al.} (eds.)
  \emph{\bibinfo{booktitle}{Proceedings of the International Conference on
  Medical Image Computing and Computer-Assisted Intervention}},
  \bibinfo{pages}{589--599} (\bibinfo{publisher}{Springer Nature Switzerland},
  \bibinfo{address}{Cham}, \bibinfo{year}{2023}).

\bibitem{silva2023flair}
\bibinfo{author}{Silva-Rodríguez, J.}, \bibinfo{author}{Chakor, H.},
  \bibinfo{author}{Kobbi, R.}, \bibinfo{author}{Dolz, J.} \&
  \bibinfo{author}{{Ben Ayed}, I.}
\newblock \bibinfo{title}{A foundation language-image model of the retina
  ({FLAIR}): encoding expert knowledge in text supervision}.
\newblock \emph{\bibinfo{journal}{Medical Image Analysis}}
  \textbf{\bibinfo{volume}{99}}, \bibinfo{pages}{103357}
  (\bibinfo{year}{2025}).

\bibitem{sukei2024multimodal}
\bibinfo{author}{S{\"u}kei, E.} \emph{et~al.}
\newblock \bibinfo{title}{Multi-modal representation learning in retinal
  imaging using self-supervised learning for enhanced clinical predictions}.
\newblock \emph{\bibinfo{journal}{Scientific Reports}}
  \textbf{\bibinfo{volume}{14}}, \bibinfo{pages}{26802} (\bibinfo{year}{2024}).

\bibitem{dosovitskiy2020image}
\bibinfo{author}{Dosovitskiy, A.} \emph{et~al.}
\newblock \bibinfo{title}{An image is worth 16x16 words: Transformers for image
  recognition at scale}.
\newblock In \emph{\bibinfo{booktitle}{Proceedings of the International
  Conference on Learning Representations}} (\bibinfo{year}{2021}).

\bibitem{bommasani2021opportunities}
\bibinfo{author}{Bommasani, R.} \emph{et~al.}
\newblock \bibinfo{title}{On the opportunities and risks of foundation models}.
\newblock \emph{\bibinfo{journal}{arXiv preprint arXiv:2108.07258}}
  (\bibinfo{year}{2021}).

\bibitem{tu2023towards}
\bibinfo{author}{Tu, T.} \emph{et~al.}
\newblock \bibinfo{title}{Towards generalist biomedical {AI}}.
\newblock \emph{\bibinfo{journal}{NEJM AI}} \textbf{\bibinfo{volume}{1}},
  \bibinfo{pages}{AIoa2300138} (\bibinfo{year}{2024}).

\bibitem{oquab2024dinov2}
\bibinfo{author}{Oquab, M.} \emph{et~al.}
\newblock \bibinfo{title}{{DINO}v2: Learning robust visual features without
  supervision}.
\newblock \emph{\bibinfo{journal}{Transactions on Machine Learning Research}}
  (\bibinfo{year}{2024}).

\bibitem{zhou2023retfound}
\bibinfo{author}{Zhou, Y.} \emph{et~al.}
\newblock \bibinfo{title}{A foundation model for generalizable disease
  detection from retinal images}.
\newblock \emph{\bibinfo{journal}{Nature}} \textbf{\bibinfo{volume}{622}},
  \bibinfo{pages}{156--163} (\bibinfo{year}{2023}).

\bibitem{qiu2024visionfm}
\bibinfo{author}{Qiu, J.} \emph{et~al.}
\newblock \bibinfo{title}{Development and validation of a multimodal multitask
  vision foundation model for generalist ophthalmic artificial intelligence}.
\newblock \emph{\bibinfo{journal}{NEJM AI}} \textbf{\bibinfo{volume}{1}},
  \bibinfo{pages}{AIoa2300221} (\bibinfo{year}{2024}).

\bibitem{shi2024eyefound}
\bibinfo{author}{Shi, D.} \emph{et~al.}
\newblock \bibinfo{title}{{EyeFound}: A multimodal generalist foundation model
  for ophthalmic imaging}.
\newblock \emph{\bibinfo{journal}{arXiv preprint arXiv:2405.11338}}
  (\bibinfo{year}{2024}).

\bibitem{shi2024eyeclip}
\bibinfo{author}{Shi, D.} \emph{et~al.}
\newblock \bibinfo{title}{{EyeCLIP}: A visual-language foundation model for
  multi-modal ophthalmic image analysis}.
\newblock \emph{\bibinfo{journal}{arXiv preprint arXiv:2409.06644}}
  (\bibinfo{year}{2024}).

\bibitem{deng2009imagenet}
\bibinfo{author}{Deng, J.} \emph{et~al.}
\newblock \bibinfo{title}{{ImageNet}: A large-scale hierarchical image
  database}.
\newblock In \emph{\bibinfo{booktitle}{Proceedings of the IEEE/CVF Conference
  on Computer Vision and Pattern Recognition}}, \bibinfo{pages}{248--255}
  (\bibinfo{organization}{IEEE}, \bibinfo{year}{2009}).

\bibitem{cai2024uni4eye++}
\bibinfo{author}{Cai, Z.}, \bibinfo{author}{Lin, L.}, \bibinfo{author}{He, H.},
  \bibinfo{author}{Cheng, P.} \& \bibinfo{author}{Tang, X.}
\newblock \bibinfo{title}{{Uni4Eye++}: A general masked image modeling
  multi-modal pre-training framework for ophthalmic image classification and
  segmentation}.
\newblock \emph{\bibinfo{journal}{IEEE Transactions on Medical Imaging}}
  (\bibinfo{year}{2024}).

\bibitem{dong2023maskclip}
\bibinfo{author}{Dong, X.} \emph{et~al.}
\newblock \bibinfo{title}{{MaskCLIP}: Masked self-distillation advances
  contrastive language-image pretraining}.
\newblock In \emph{\bibinfo{booktitle}{Proceedings of the IEEE/CVF Conference
  on Computer Vision and Pattern Recognition}}, \bibinfo{pages}{10995--11005}
  (\bibinfo{year}{2023}).

\bibitem{mazurowski2023segment}
\bibinfo{author}{Mazurowski, M.~A.} \emph{et~al.}
\newblock \bibinfo{title}{Segment anything model for medical image analysis: an
  experimental study}.
\newblock \emph{\bibinfo{journal}{Medical Image Analysis}}
  \textbf{\bibinfo{volume}{89}}, \bibinfo{pages}{102918}
  (\bibinfo{year}{2023}).

\bibitem{huang2024segment}
\bibinfo{author}{Huang, Y.} \emph{et~al.}
\newblock \bibinfo{title}{Segment anything model for medical images?}
\newblock \emph{\bibinfo{journal}{Medical Image Analysis}}
  \textbf{\bibinfo{volume}{92}}, \bibinfo{pages}{103061}
  (\bibinfo{year}{2024}).

\bibitem{ma2024segment}
\bibinfo{author}{Ma, J.} \emph{et~al.}
\newblock \bibinfo{title}{Segment anything in medical images}.
\newblock \emph{\bibinfo{journal}{Nature Communications}}
  \textbf{\bibinfo{volume}{15}}, \bibinfo{pages}{654} (\bibinfo{year}{2024}).

\bibitem{zhu2024medsam2}
\bibinfo{author}{Zhu, J.}, \bibinfo{author}{Qi, Y.} \& \bibinfo{author}{Wu, J.}
\newblock \bibinfo{title}{{Medical SAM} 2: Segment medical images as video via
  {Segment Anything Model} 2}.
\newblock \emph{\bibinfo{journal}{arXiv preprint arXiv:2408.00874}}
  (\bibinfo{year}{2024}).

\bibitem{kirillov2023segment}
\bibinfo{author}{Kirillov, A.} \emph{et~al.}
\newblock \bibinfo{title}{Segment anything}.
\newblock In \emph{\bibinfo{booktitle}{Proceedings of the IEEE/CVF
  International Conference on Computer Vision}}, \bibinfo{pages}{4015--4026}
  (\bibinfo{year}{2023}).

\bibitem{fazekas2023adapting}
\bibinfo{author}{Fazekas, B.}, \bibinfo{author}{Morano, J.},
  \bibinfo{author}{Lachinov, D.}, \bibinfo{author}{Aresta, G.} \&
  \bibinfo{author}{Bogunovi{\'{c}}, H.}
\newblock \bibinfo{title}{Adapting segment anything model ({SAM}) for retinal
  {OCT}}.
\newblock In \bibinfo{editor}{Antony, B.} \emph{et~al.} (eds.)
  \emph{\bibinfo{booktitle}{Ophthalmic Medical Image Analysis}},
  \bibinfo{pages}{92--101} (\bibinfo{publisher}{Springer Nature Switzerland},
  \bibinfo{address}{Cham}, \bibinfo{year}{2023}).

\bibitem{gerendas2022validation}
\bibinfo{author}{Gerendas, B.~S.} \emph{et~al.}
\newblock \bibinfo{title}{Validation of an automated fluid algorithm on
  real-world data of neovascular age-related macular degeneration over five
  years}.
\newblock \emph{\bibinfo{journal}{RETINA}} \textbf{\bibinfo{volume}{42}}
  (\bibinfo{year}{2022}).

\bibitem{garvin2008intraretinal}
\bibinfo{author}{Garvin, M.~K.} \emph{et~al.}
\newblock \bibinfo{title}{Intraretinal layer segmentation of macular optical
  coherence tomography images using optimal 3-d graph search}.
\newblock \emph{\bibinfo{journal}{IEEE Transactions on Medical Imaging}}
  \textbf{\bibinfo{volume}{27}}, \bibinfo{pages}{1495--1505}
  (\bibinfo{year}{2008}).

\bibitem{antony2011automated}
\bibinfo{author}{Antony, B.} \emph{et~al.}
\newblock \bibinfo{title}{Automated 3-d method for the correction of axial
  artifacts in spectral-domain optical coherence tomography images}.
\newblock \emph{\bibinfo{journal}{Biomedical Optics Express}}
  \textbf{\bibinfo{volume}{2}}, \bibinfo{pages}{2403} (\bibinfo{year}{2011}).

\bibitem{farsiu2014quantitative}
\bibinfo{author}{Farsiu, S.} \emph{et~al.}
\newblock \bibinfo{title}{Quantitative classification of eyes with and without
  intermediate age-related macular degeneration using optical coherence
  tomography}.
\newblock \emph{\bibinfo{journal}{Ophthalmology}}
  \textbf{\bibinfo{volume}{121}}, \bibinfo{pages}{162--172}
  (\bibinfo{year}{2014}).
\newblock
  \urlprefix\url{https://people.duke.edu/~sf59/RPEDC_Ophth_2013_dataset.htm}.

\bibitem{wu2023gamma}
\bibinfo{author}{Wu, J.} \emph{et~al.}
\newblock \bibinfo{title}{{GAMMA} challenge: Glaucoma grading from
  multi-modality images}.
\newblock \emph{\bibinfo{journal}{Medical Image Analysis}}
  \textbf{\bibinfo{volume}{90}}, \bibinfo{pages}{102938}
  (\bibinfo{year}{2023}).
\newblock \urlprefix\url{https://gamma.grand-challenge.org/}.

\bibitem{luo2023harvard}
\bibinfo{author}{Luo, Y.}, \bibinfo{author}{Shi, M.}, \bibinfo{author}{Tian,
  Y.}, \bibinfo{author}{Elze, T.} \& \bibinfo{author}{Wang, M.}
\newblock \bibinfo{title}{Harvard glaucoma detection and progression: A
  multimodal multitask dataset and generalization-reinforced semi-supervised
  learning}.
\newblock In \emph{\bibinfo{booktitle}{Proceedings of the IEEE/CVF
  International Conference on Computer Vision}}, \bibinfo{pages}{20471--20482}
  (\bibinfo{year}{2023}).
\newblock
  \urlprefix\url{https://ophai.hms.harvard.edu/datasets/harvard-gdp1000/}.

\bibitem{kermany2018dataset}
\bibinfo{author}{Kermany, D.}
\newblock \bibinfo{title}{Labeled optical coherence tomography ({OCT}) and
  chest {X}-ray images for classification} (\bibinfo{year}{2018}).
\newblock \urlprefix\url{https://data.mendeley.com/datasets/rscbjbr9sj/2}.

\bibitem{kermany2018identifying}
\bibinfo{author}{Kermany, D.~S.} \emph{et~al.}
\newblock \bibinfo{title}{Identifying medical diagnoses and treatable diseases
  by image-based deep learning}.
\newblock \emph{\bibinfo{journal}{Cell}} \textbf{\bibinfo{volume}{172}},
  \bibinfo{pages}{1122--1131.e9} (\bibinfo{year}{2018}).

\bibitem{rasti2017macular}
\bibinfo{author}{Rasti, R.}, \bibinfo{author}{Rabbani, H.},
  \bibinfo{author}{Mehridehnavi, A.} \& \bibinfo{author}{Hajizadeh, F.}
\newblock \bibinfo{title}{Macular {OCT} classification using a multi-scale
  convolutional neural network ensemble}.
\newblock \emph{\bibinfo{journal}{IEEE Transactions on Medical Imaging}}
  \textbf{\bibinfo{volume}{37}}, \bibinfo{pages}{1024--1034}
  (\bibinfo{year}{2017}).
\newblock
  \urlprefix\url{https://hrabbani.site123.me/available-datasets/dataset-for-oct-classification-50-normal-48-amd-50-dme}.

\bibitem{kulyabin2024octdl}
\bibinfo{author}{Kulyabin, M.} \emph{et~al.}
\newblock \bibinfo{title}{{OCTDL}: Optical coherence tomography dataset for
  image-based deep learning methods}.
\newblock \emph{\bibinfo{journal}{Scientific Data}}
  \textbf{\bibinfo{volume}{11}}, \bibinfo{pages}{365} (\bibinfo{year}{2024}).
\newblock \urlprefix\url{https://dx.doi.org/10.21227/fpvs-8n55}.

\bibitem{gholami2020octid}
\bibinfo{author}{Gholami, P.}, \bibinfo{author}{Roy, P.},
  \bibinfo{author}{Parthasarathy, M.~K.} \& \bibinfo{author}{Lakshminarayanan,
  V.}
\newblock \bibinfo{title}{{OCTID}: Optical coherence tomography image
  database}.
\newblock \emph{\bibinfo{journal}{Computers \& Electrical Engineering}}
  \textbf{\bibinfo{volume}{81}}, \bibinfo{pages}{106532}
  (\bibinfo{year}{2020}).
\newblock \urlprefix\url{https://borealisdata.ca/dataverse/OCTID}.

\bibitem{prabhushankar2022olives}
\bibinfo{author}{Prabhushankar, M.} \emph{et~al.}
\newblock \bibinfo{title}{{OLIVES} dataset: Ophthalmic labels for investigating
  visual eye semantics}.
\newblock In \emph{\bibinfo{booktitle}{Advances in Neural Information
  Processing Systems}} (\bibinfo{year}{2022}).
\newblock \urlprefix\url{https://zenodo.org/records/7105232}.

\bibitem{oghbaie2024vlfatrollout}
\bibinfo{author}{Oghbaie, M.}, \bibinfo{author}{Araújo, T.},
  \bibinfo{author}{Schmidt-Erfurth, U.} \& \bibinfo{author}{Bogunović, H.}
\newblock \bibinfo{title}{{VLFATRollout}: Fully transformer-based classifier
  for retinal {OCT} volumes}.
\newblock \emph{\bibinfo{journal}{Computerized Medical Imaging and Graphics}}
  \textbf{\bibinfo{volume}{118}}, \bibinfo{pages}{102452}
  (\bibinfo{year}{2024}).

\bibitem{rashno2018fully}
\bibinfo{author}{Rashno, A.} \emph{et~al.}
\newblock \bibinfo{title}{Fully automated segmentation of fluid/cyst regions in
  optical coherence tomography images with diabetic macular edema using
  neutrosophic sets and graph algorithms}.
\newblock \emph{\bibinfo{journal}{IEEE Transactions on Biomedical Engineering}}
  \textbf{\bibinfo{volume}{65}}, \bibinfo{pages}{989--1001}
  (\bibinfo{year}{2018}).
\newblock \urlprefix\url{https://people.ece.umn.edu/users/parhi/.DATA/}.

\bibitem{srinivasan2014fully}
\bibinfo{author}{Srinivasan, P.~P.} \emph{et~al.}
\newblock \bibinfo{title}{Fully automated detection of diabetic macular edema
  and dry age-related macular degeneration from optical coherence tomography
  images}.
\newblock \emph{\bibinfo{journal}{Biomedical Optics Express}}
  \textbf{\bibinfo{volume}{5}}, \bibinfo{pages}{3568--3577}
  (\bibinfo{year}{2014}).
\newblock
  \urlprefix\url{https://people.duke.edu/~sf59/Srinivasan_BOE_2014_dataset.htm}.

\bibitem{chiu2015kernel}
\bibinfo{author}{Chiu, S.~J.} \emph{et~al.}
\newblock \bibinfo{title}{Kernel regression based segmentation of optical
  coherence tomography images with diabetic macular edema}.
\newblock \emph{\bibinfo{journal}{Biomedical Optics Express}}
  \textbf{\bibinfo{volume}{6}}, \bibinfo{pages}{1172--1194}
  (\bibinfo{year}{2015}).
\newblock
  \urlprefix\url{https://people.duke.edu/~sf59/Chiu_BOE_2014_dataset.htm}.

\bibitem{melinscak2021aroi}
\bibinfo{author}{Melinščak, M.}, \bibinfo{author}{Radmilovič, M.},
  \bibinfo{author}{Vatavuk, Z.} \& \bibinfo{author}{Lončarić, S.}
\newblock \bibinfo{title}{{AROI}: Annotated retinal {OCT} images database}.
\newblock In \emph{\bibinfo{booktitle}{Proceedings of the International
  Convention on Information, Communication and Electronic Technology}},
  \bibinfo{pages}{371--376} (\bibinfo{year}{2021}).
\newblock \urlprefix\url{https://ipg.fer.hr/ipg/resources/oct_image_database}.

\bibitem{bogunovic2019retouch}
\bibinfo{author}{Bogunovi{\'c}, H.} \emph{et~al.}
\newblock \bibinfo{title}{{RETOUCH}: The retinal {OCT} fluid detection and
  segmentation benchmark and challenge}.
\newblock \emph{\bibinfo{journal}{IEEE Transactions on Medical Imaging}}
  \textbf{\bibinfo{volume}{38}}, \bibinfo{pages}{1858--1874}
  (\bibinfo{year}{2019}).
\newblock \urlprefix\url{https://retouch.grand-challenge.org/}.

\bibitem{fang2022dataset}
\bibinfo{author}{Fang, H.} \emph{et~al.}
\newblock \bibinfo{title}{Dataset and evaluation algorithm design for {GOALS}
  challenge}.
\newblock In \bibinfo{editor}{Antony, B.} \emph{et~al.} (eds.)
  \emph{\bibinfo{booktitle}{Ophthalmic Medical Image Analysis}},
  \bibinfo{pages}{135--142} (\bibinfo{publisher}{Springer International
  Publishing}, \bibinfo{address}{Cham}, \bibinfo{year}{2022}).

\bibitem{bui2022fundus}
\bibinfo{author}{Bui, P. T.~A.} \emph{et~al.}
\newblock \bibinfo{title}{Fundus autofluorescence and optical coherence
  tomography biomarkers associated with the progression of geographic atrophy
  secondary to age-related macular degeneration}.
\newblock \emph{\bibinfo{journal}{Eye}} \textbf{\bibinfo{volume}{36}},
  \bibinfo{pages}{2013--2019} (\bibinfo{year}{2022}).

\bibitem{liu2022convnext}
\bibinfo{author}{Liu, Z.} \emph{et~al.}
\newblock \bibinfo{title}{A convnet for the 2020s}.
\newblock In \emph{\bibinfo{booktitle}{Proceedings of the IEEE/CVF Conference
  on Computer Vision and Pattern Recognition}}, \bibinfo{pages}{11976--11986}
  (\bibinfo{year}{2022}).

\bibitem{he2023swinunetrv2}
\bibinfo{author}{He, Y.} \emph{et~al.}
\newblock \bibinfo{title}{{SwinUNETR-V2}: Stronger swin transformers with
  stagewise convolutions for {3D} medical image segmentation}.
\newblock In \emph{\bibinfo{booktitle}{International Conference on Medical
  Image Computing and Computer-Assisted Intervention}},
  \bibinfo{pages}{416--426} (\bibinfo{organization}{Springer},
  \bibinfo{year}{2023}).

\bibitem{roy2023mednext}
\bibinfo{author}{Roy, S.} \emph{et~al.}
\newblock \bibinfo{title}{{MedNeXt}: Transformer-driven scaling of convnets for
  medical image segmentation}.
\newblock In \emph{\bibinfo{booktitle}{International Conference on Medical
  Image Computing and Computer-Assisted Intervention}},
  \bibinfo{pages}{405--415} (\bibinfo{organization}{Springer},
  \bibinfo{year}{2023}).

\bibitem{chen2024transunet}
\bibinfo{author}{Chen, J.} \emph{et~al.}
\newblock \bibinfo{title}{{TransUNet}: Rethinking the {U-Net} architecture
  design for medical image segmentation through the lens of transformers}.
\newblock \emph{\bibinfo{journal}{Medical Image Analysis}}
  \textbf{\bibinfo{volume}{97}}, \bibinfo{pages}{103280}
  (\bibinfo{year}{2024}).

\bibitem{isensee2021nnunet}
\bibinfo{author}{Isensee, F.}, \bibinfo{author}{Jaeger, P.~F.},
  \bibinfo{author}{Kohl, S. A.~A.}, \bibinfo{author}{Petersen, J.} \&
  \bibinfo{author}{Maier-Hein, K.~H.}
\newblock \bibinfo{title}{{nnU-Net}: A self-configuring method for deep
  learning-based biomedical image segmentation}.
\newblock \emph{\bibinfo{journal}{Nature Methods}}
  \textbf{\bibinfo{volume}{18}}, \bibinfo{pages}{203--211}
  (\bibinfo{year}{2021}).

\bibitem{isensee2024nnunet2}
\bibinfo{author}{Isensee, F.} \emph{et~al.}
\newblock \bibinfo{title}{{nnU-Net} revisited: A call for rigorous validation
  in {3D} medical image segmentation}.
\newblock In \emph{\bibinfo{booktitle}{International Conference on Medical
  Image Computing and Computer-Assisted Intervention}},
  \bibinfo{pages}{488--498} (\bibinfo{organization}{Springer},
  \bibinfo{year}{2024}).

\bibitem{liu2021swin}
\bibinfo{author}{Liu, Z.} \emph{et~al.}
\newblock \bibinfo{title}{Swin transformer: Hierarchical vision transformer
  using shifted windows}.
\newblock In \emph{\bibinfo{booktitle}{Proceedings of the IEEE/CVF
  international conference on computer vision}}, \bibinfo{pages}{10012--10022}
  (\bibinfo{year}{2021}).

\bibitem{vaswani2017attention}
\bibinfo{author}{Vaswani, A.} \emph{et~al.}
\newblock \bibinfo{title}{Attention is all you need}.
\newblock \emph{\bibinfo{journal}{Advances in Neural Information Processing
  Systems}} \textbf{\bibinfo{volume}{30}} (\bibinfo{year}{2017}).

\bibitem{ronneberger2015unet}
\bibinfo{author}{Ronneberger, O.}, \bibinfo{author}{Fischer, P.} \&
  \bibinfo{author}{Brox, T.}
\newblock \bibinfo{title}{U-net: Convolutional networks for biomedical image
  segmentation}.
\newblock In \emph{\bibinfo{booktitle}{Proceedings of the international
  conference on Medical Image Computing and Computer-Assisted Intervention}},
  \bibinfo{pages}{234--241} (\bibinfo{organization}{Springer},
  \bibinfo{year}{2015}).

\bibitem{he2021structured}
\bibinfo{author}{He, Y.} \emph{et~al.}
\newblock \bibinfo{title}{Structured layer surface segmentation for retina
  {OCT} using fully convolutional regression networks}.
\newblock \emph{\bibinfo{journal}{Medical Image Analysis}}
  \textbf{\bibinfo{volume}{68}}, \bibinfo{pages}{101856}
  (\bibinfo{year}{2021}).

\bibitem{fazekas2023segmentation}
\bibinfo{author}{Fazekas, B.} \emph{et~al.}
\newblock \bibinfo{title}{Segmentation of {B}ruch's membrane in retinal {OCT}
  with {AMD} using anatomical priors and uncertainty quantification}.
\newblock \emph{\bibinfo{journal}{IEEE Journal of Biomedical and Health
  Informatics}} \textbf{\bibinfo{volume}{27}}, \bibinfo{pages}{41--52}
  (\bibinfo{year}{2023}).

\bibitem{morano2024rrwnet}
\bibinfo{author}{Morano, J.}, \bibinfo{author}{Aresta, G.} \&
  \bibinfo{author}{Bogunović, H.}
\newblock \bibinfo{title}{{RRWNet}: Recursive refinement network for effective
  retinal artery/vein segmentation and classification}.
\newblock \emph{\bibinfo{journal}{Expert Systems with Applications}}
  \textbf{\bibinfo{volume}{256}}, \bibinfo{pages}{124970}
  (\bibinfo{year}{2024}).

\bibitem{azad2024medical}
\bibinfo{author}{Azad, R.} \emph{et~al.}
\newblock \bibinfo{title}{Medical image segmentation review: The success of
  {U-Net}}.
\newblock \emph{\bibinfo{journal}{IEEE Transactions on Pattern Analysis and
  Machine Intelligence}} \bibinfo{pages}{1--20} (\bibinfo{year}{2024}).

\bibitem{singh2022flava}
\bibinfo{author}{Singh, A.} \emph{et~al.}
\newblock \bibinfo{title}{{FLAVA}: A foundational language and vision alignment
  model}.
\newblock In \emph{\bibinfo{booktitle}{Proceedings of the IEEE/CVF Conference
  on Computer Vision and Pattern Recognition}}, \bibinfo{pages}{15638--15650}
  (\bibinfo{year}{2022}).

\bibitem{chen2024towards}
\bibinfo{author}{Chen, R.~J.} \emph{et~al.}
\newblock \bibinfo{title}{Towards a general-purpose foundation model for
  computational pathology}.
\newblock \emph{\bibinfo{journal}{Nature Medicine}}
  \textbf{\bibinfo{volume}{30}}, \bibinfo{pages}{850--862}
  (\bibinfo{year}{2024}).

\bibitem{guymer2023age}
\bibinfo{author}{Guymer, R.~H.} \& \bibinfo{author}{Campbell, T.~G.}
\newblock \bibinfo{title}{Age-related macular degeneration}.
\newblock \emph{\bibinfo{journal}{The Lancet}} \textbf{\bibinfo{volume}{401}},
  \bibinfo{pages}{1459--1472} (\bibinfo{year}{2023}).

\bibitem{devlin2018bert}
\bibinfo{author}{Devlin, J.}, \bibinfo{author}{Chang, M.-W.},
  \bibinfo{author}{Lee, K.} \& \bibinfo{author}{Toutanova, K.}
\newblock \bibinfo{title}{Bert: Pre-training of deep bidirectional transformers
  for language understanding}.
\newblock \emph{\bibinfo{journal}{arXiv preprint arXiv:1810.04805}}
  (\bibinfo{year}{2018}).

\bibitem{brown2020language}
\bibinfo{author}{Brown, T.} \emph{et~al.}
\newblock \bibinfo{title}{Language models are few-shot learners}.
\newblock In \bibinfo{editor}{Larochelle, H.}, \bibinfo{editor}{Ranzato, M.},
  \bibinfo{editor}{Hadsell, R.}, \bibinfo{editor}{Balcan, M.} \&
  \bibinfo{editor}{Lin, H.} (eds.) \emph{\bibinfo{booktitle}{Advances in Neural
  Information Processing Systems}}, vol.~\bibinfo{volume}{33},
  \bibinfo{pages}{1877--1901} (\bibinfo{publisher}{Curran Associates, Inc.},
  \bibinfo{year}{2020}).

\bibitem{moor2023foundation}
\bibinfo{author}{Moor, M.} \emph{et~al.}
\newblock \bibinfo{title}{Foundation models for generalist medical artificial
  intelligence}.
\newblock \emph{\bibinfo{journal}{Nature}} \textbf{\bibinfo{volume}{616}},
  \bibinfo{pages}{259--265} (\bibinfo{year}{2023}).

\bibitem{zhang2024challenges}
\bibinfo{author}{Zhang, S.} \& \bibinfo{author}{Metaxas, D.}
\newblock \bibinfo{title}{On the challenges and perspectives of foundation
  models for medical image analysis}.
\newblock \emph{\bibinfo{journal}{Medical Image Analysis}}
  \textbf{\bibinfo{volume}{91}}, \bibinfo{pages}{102996}
  (\bibinfo{year}{2024}).

\bibitem{george2020attention}
\bibinfo{author}{George, Y.} \emph{et~al.}
\newblock \bibinfo{title}{Attention-guided {3D-CNN} framework for glaucoma
  detection and structural-functional association using volumetric images}.
\newblock \emph{\bibinfo{journal}{IEEE journal of biomedical and health
  informatics}} \textbf{\bibinfo{volume}{24}}, \bibinfo{pages}{3421--3430}
  (\bibinfo{year}{2020}).

\bibitem{lin2021assessing}
\bibinfo{author}{Lin, A.~C.}, \bibinfo{author}{Lee, C.~S.},
  \bibinfo{author}{Blazes, M.}, \bibinfo{author}{Lee, A.~Y.} \&
  \bibinfo{author}{Gorin, M.~B.}
\newblock \bibinfo{title}{Assessing the clinical utility of expanded macular
  {OCTs} using machine learning}.
\newblock \emph{\bibinfo{journal}{Translational Vision Science and Technology}}
  \textbf{\bibinfo{volume}{10}}, \bibinfo{pages}{32--32}
  (\bibinfo{year}{2021}).

\bibitem{liu2024simultaneous}
\bibinfo{author}{Liu, H.} \emph{et~al.}
\newblock \bibinfo{title}{Simultaneous alignment and surface regression using
  hybrid {2D–3D} networks for {3D} coherent layer segmentation of retinal
  {OCT} images with full and sparse annotations}.
\newblock \emph{\bibinfo{journal}{Medical Image Analysis}}
  \textbf{\bibinfo{volume}{91}}, \bibinfo{pages}{103019}
  (\bibinfo{year}{2024}).

\bibitem{fazekas2025sdlayernet}
\bibinfo{author}{Fazekas, B.} \emph{et~al.}
\newblock \bibinfo{title}{{SD-LayerNet}: Robust and label-efficient retinal
  layer segmentation via anatomical priors}.
\newblock \emph{\bibinfo{journal}{Computer Methods and Programs in
  Biomedicine}} \textbf{\bibinfo{volume}{261}}, \bibinfo{pages}{108586}
  (\bibinfo{year}{2025}).

\bibitem{li2021multiscale}
\bibinfo{author}{Li, J.} \emph{et~al.}
\newblock \bibinfo{title}{Multi-scale {GCN}-assisted two-stage network for
  joint segmentation of retinal layers and discs in peripapillary {OCT}
  images}.
\newblock \emph{\bibinfo{journal}{Biomed. Opt. Express}}
  \textbf{\bibinfo{volume}{12}}, \bibinfo{pages}{2204--2220}
  (\bibinfo{year}{2021}).

\bibitem{hu2021lora}
\bibinfo{author}{Hu, E.~J.} \emph{et~al.}
\newblock \bibinfo{title}{{LoRA}: Low-rank adaptation of large language
  models}.
\newblock \emph{\bibinfo{journal}{arXiv preprint arXiv:2106.09685}}
  (\bibinfo{year}{2021}).

\bibitem{steiner2022howto}
\bibinfo{author}{Steiner, A.~P.} \emph{et~al.}
\newblock \bibinfo{title}{How to train your {ViT}? data, augmentation, and
  regularization in vision transformers}.
\newblock \emph{\bibinfo{journal}{Transactions on Machine Learning Research}}
  (\bibinfo{year}{2022}).

\bibitem{jeong2024medical}
\bibinfo{author}{Jeong, D.}, \bibinfo{author}{Garg, S.},
  \bibinfo{author}{Lipton, Z.~C.} \& \bibinfo{author}{Oberst, M.}
\newblock \bibinfo{title}{Medical adaptation of large language and
  vision-language models: Are we making progress?}
\newblock In \emph{\bibinfo{booktitle}{Proceedings of the 2024 Conference on
  Empirical Methods in Natural Language Processing}},
  \bibinfo{pages}{12143--12170} (\bibinfo{year}{2024}).

\bibitem{antaki2024vision}
\bibinfo{author}{Antaki, F.}, \bibinfo{author}{Chopra, R.} \&
  \bibinfo{author}{Keane, P.~A.}
\newblock \bibinfo{title}{{Vision-Language} models for feature detection of
  macular diseases on optical coherence tomography}.
\newblock \emph{\bibinfo{journal}{JAMA Ophthalmol}}
  \textbf{\bibinfo{volume}{142}}, \bibinfo{pages}{573--576}
  (\bibinfo{year}{2024}).

\bibitem{roy2017relaynet}
\bibinfo{author}{Roy, A.~G.} \emph{et~al.}
\newblock \bibinfo{title}{{ReLayNet}: Retinal layer and fluid segmentation of
  macular optical coherence tomography using fully convolutional networks}.
\newblock \emph{\bibinfo{journal}{Biomedical Optics Express}}
  \textbf{\bibinfo{volume}{8}}, \bibinfo{pages}{3627--3642}
  (\bibinfo{year}{2017}).

\bibitem{paszke2019pytorch}
\bibinfo{author}{Paszke, A.} \emph{et~al.}
\newblock \bibinfo{title}{Pytorch: An imperative style, high-performance deep
  learning library}.
\newblock \emph{\bibinfo{journal}{Advances in Neural Information Processing
  Systems}} \textbf{\bibinfo{volume}{32}} (\bibinfo{year}{2019}).

\bibitem{loshchilov2018adamw}
\bibinfo{author}{Loshchilov, I.} \& \bibinfo{author}{Hutter, F.}
\newblock \bibinfo{title}{Decoupled weight decay regularization}.
\newblock In \emph{\bibinfo{booktitle}{Proceedings of the International
  Conference on Learning Representations}} (\bibinfo{year}{2018}).

\bibitem{goyal2017accurate}
\bibinfo{author}{Goyal, P.} \emph{et~al.}
\newblock \bibinfo{title}{Accurate, large minibatch {SGD}: training {ImageNet}
  in 1 hour}.
\newblock \emph{\bibinfo{journal}{arXiv preprint arXiv:1706.02677}}
  (\bibinfo{year}{2017}).

\bibitem{ranftl2021vision}
\bibinfo{author}{Ranftl, R.}, \bibinfo{author}{Bochkovskiy, A.} \&
  \bibinfo{author}{Koltun, V.}
\newblock \bibinfo{title}{Vision transformers for dense prediction}.
\newblock In \emph{\bibinfo{booktitle}{Proceedings of the IEEE/CVF
  international conference on computer vision}}, \bibinfo{pages}{12179--12188}
  (\bibinfo{year}{2021}).

\bibitem{provost2000well}
\bibinfo{author}{Provost, F.} \& \bibinfo{author}{Domingos, P.}
\newblock \bibinfo{title}{Well-trained {PETs}: Improving probability estimation
  trees}.
\newblock \emph{\bibinfo{journal}{CeDER Working Paper IS-00-04, Stern School of
  Business, New York University}}  (\bibinfo{year}{2000}).

\bibitem{fawcett2006introduction}
\bibinfo{author}{Fawcett, T.}
\newblock \bibinfo{title}{An introduction to {ROC} analysis}.
\newblock \emph{\bibinfo{journal}{Pattern recognition letters}}
  \textbf{\bibinfo{volume}{27}}, \bibinfo{pages}{861--874}
  (\bibinfo{year}{2006}).

\bibitem{mosley2013balanced}
\bibinfo{author}{Mosley, L.}
\newblock \bibinfo{title}{A balanced approach to the multi-class imbalance
  problem}  (\bibinfo{year}{2013}).

\bibitem{demvsar2006statistical}
\bibinfo{author}{Dem{\v{s}}ar, J.}
\newblock \bibinfo{title}{Statistical comparisons of classifiers over multiple
  data sets}.
\newblock \emph{\bibinfo{journal}{The Journal of Machine learning research}}
  \textbf{\bibinfo{volume}{7}}, \bibinfo{pages}{1--30} (\bibinfo{year}{2006}).

\bibitem{heimann2009comparison}
\bibinfo{author}{Heimann, T.} \emph{et~al.}
\newblock \bibinfo{title}{Comparison and evaluation of methods for liver
  segmentation from {CT} datasets}.
\newblock \emph{\bibinfo{journal}{IEEE transactions on medical imaging}}
  \textbf{\bibinfo{volume}{28}}, \bibinfo{pages}{1251--1265}
  (\bibinfo{year}{2009}).

\bibitem{strudel2021segmenter}
\bibinfo{author}{Strudel, R.}, \bibinfo{author}{Garcia, R.},
  \bibinfo{author}{Laptev, I.} \& \bibinfo{author}{Schmid, C.}
\newblock \bibinfo{title}{Segmenter: Transformer for semantic segmentation}.
\newblock In \emph{\bibinfo{booktitle}{Proceedings of the IEEE/CVF
  international conference on computer vision}}, \bibinfo{pages}{7262--7272}
  (\bibinfo{year}{2021}).

\bibitem{yu2021prime}
\bibinfo{author}{Yu, H.~J.} \emph{et~al.}
\newblock \bibinfo{title}{Real-time photographic- and fluorescein
  angiographic-guided management of diabetic retinopathy: Randomized {PRIME}
  trial outcomes}.
\newblock \emph{\bibinfo{journal}{American Journal of Ophthalmology}}
  \textbf{\bibinfo{volume}{226}}, \bibinfo{pages}{126--136}
  (\bibinfo{year}{2021}).

\bibitem{payne2017trexdme}
\bibinfo{author}{Payne, J.~F.} \emph{et~al.}
\newblock \bibinfo{title}{Randomized trial of treat and extend {Ranibizumab}
  with and without navigated laser for diabetic macular edema: {TREX-DME} 1
  year outcomes}.
\newblock \emph{\bibinfo{journal}{Ophthalmology}}
  \textbf{\bibinfo{volume}{124}}, \bibinfo{pages}{74--81}
  (\bibinfo{year}{2017}).

\bibitem{schlanitz2017drusen}
\bibinfo{author}{Schlanitz, F.~G.} \emph{et~al.}
\newblock \bibinfo{title}{Drusen volume development over time and its relevance
  to the course of age-related macular degeneration}.
\newblock \emph{\bibinfo{journal}{British Journal of Ophthalmology}}
  \textbf{\bibinfo{volume}{101}}, \bibinfo{pages}{198--203}
  (\bibinfo{year}{2017}).

\bibitem{ritter2024deep}
\bibinfo{author}{Ritter, M.} \emph{et~al.}
\newblock \bibinfo{title}{Deep learning based quantification of photoreceptor
  and retinal pigment epithelium degeneration as predictive factors in
  stargardt disease}.
\newblock \emph{\bibinfo{journal}{Investigative Ophthalmology \& Visual
  Science}} \textbf{\bibinfo{volume}{65}}, \bibinfo{pages}{3777--3777}
  (\bibinfo{year}{2024}).

\bibitem{arikan2019deep}
\bibinfo{author}{Arikan, M.}, \bibinfo{author}{Sadeghipour, A.},
  \bibinfo{author}{Gerendas, B.}, \bibinfo{author}{Told, R.} \&
  \bibinfo{author}{Schmidt-Erfurt, U.}
\newblock \bibinfo{title}{Deep learning based multi-modal registration for
  retinal imaging}.
\newblock In \bibinfo{editor}{Suzuki, K.} \emph{et~al.} (eds.)
  \emph{\bibinfo{booktitle}{Interpretability of Machine Intelligence in Medical
  Image Computing and Multimodal Learning for Clinical Decision Support}},
  \bibinfo{pages}{75--82} (\bibinfo{publisher}{Springer International
  Publishing}, \bibinfo{address}{Cham}, \bibinfo{year}{2019}).

\bibitem{mcinnes2018umap}
\bibinfo{author}{McInnes, L.}, \bibinfo{author}{Healy, J.} \&
  \bibinfo{author}{Melville, J.}
\newblock \bibinfo{title}{{UMAP}: Uniform manifold approximation and projection
  for dimension reduction}.
\newblock \emph{\bibinfo{journal}{arXiv preprint arXiv:1802.03426}}
  (\bibinfo{year}{2018}).

\end{thebibliography}


\end{document}